\documentclass[11pt]{article}

\usepackage{acl}

\usepackage{times}
\usepackage{latexsym}
\usepackage[T1]{fontenc}
\usepackage[utf8]{inputenc}
\usepackage{enumitem}
\usepackage{amsmath}
\usepackage{amssymb}
\usepackage{multirow}
\usepackage{microtype}
\usepackage{inconsolata}
\usepackage{graphicx}
\usepackage{subcaption}
\usepackage{booktabs}

\usepackage{xcolor}
\definecolor{beigebackground}{RGB}{245,245,220}%

\title{Prompts Without Evidence: How Neuroimaging Mentions Shift Clinical Vision-Language Model Predictions}

\author{Doan Nam Long Vu,  Simone Balloccu \\
  Natural Language Processing for Expert Domains (ExpNLP),\\ Technical University of Darmstadt
}

\begin{document}
\maketitle

\begin{abstract}
Trustworthy clinical AI must use real evidence and avoid relying on surface-level artifacts. We evaluate 12 open-weight vision-language models (VLMs) on two clinical neuroimaging cohorts for binary classification of affective disorders and cognitive decline. Both cohorts include structural magnetic resonance imaging (MRI) acquired under their original research protocols. Prior work does not establish the included neuroimaging inputs as reliable stand-alone diagnostic evidence for the present tasks. Nevertheless, when neuroimaging context is introduced, smaller VLMs gain up to 0.66 F1 under the evaluated augmented conditions, becoming competitive with models an order of magnitude larger. Confidence estimation shows that most of the calibration improvement for the analyzed smaller models occurs after the MRI reference is added to the prompt, before any image is supplied. Our preliminary expert case study finds that faithfulness remains low in every condition examined, with the reviewed model introducing unverified clinical details. Finally, in our single-model intervention, preference alignment suppresses MRI-referencing behavior but reduces the augmented-condition advantage, leaving the underlying issue unresolved. These results caution against reading surface metric gains as evidence of true multimodal integration, with direct implications for clinical VLM deployment\footnote{\raggedright\url{https://github.com/long21wt/neuro-prompt-no-evidence}\par}.
\end{abstract}

\section{Introduction}
The application of vision-language models (VLMs) to clinical decision-making has attracted growing interest, with recent work exploring their use for diagnostic classification from multimodal patient data~\cite{moor2023foundation, Singhal2023, li2023llavamed}. A natural expectation is that richer inputs should improve performance only when supplying \emph{meaningful} diagnostic evidence. While spurious correlations and hallucinations in VLMs have received growing attention~\cite{zhong-etal-2024-investigating, howard-etal-2025-uncovering}, less attention has been paid to how models respond when a clinically related context is introduced without established stand-alone diagnostic value. The distinction matters: a change in model behavior after such context is added does not by itself demonstrate meaningful integration of clinical evidence. We study this phenomenon in two clinical cohorts containing structured textual data and neuroimaging acquired under their original study protocols. Prior work does not establish these neuroimaging inputs as reliable stand-alone diagnostic evidence for the tasks considered here~\cite{winter2022,winter2024,Lombardi2020-oj,Ansart2020-mk}. In consultation with domain experts, we exclude obvious classification shortcuts from the inputs used for the affective-disorder and cognitive-decline tasks.

Across 12 open-weight VLMs, we show that mentioning neuroimaging in the prompt shifts predictions, with the largest gains in small models whose text-only behavior is near-degenerate. On \textsc{FOR2107} \cite{kircher2019-db}, \texttt{Qwen2.5-\allowbreak{}VL-\allowbreak{}3B} \cite{xu2025qwen25vltechnicalreport} improves by up to 0.58 F1, which lifts it from a near-degenerate baseline to the range of models an order of magnitude larger (e.g., \texttt{Qwen2.5-\allowbreak{}VL-\allowbreak{}72B}). This gain is not fully explained by added diagnostic evidence: most of the observed gain occurs after adding the textual MRI-availability preamble, before any image is supplied. Where adding an image yields further gains, replacing the subject-specific MRI with an unrelated out-of-distribution control image does not consistently reduce performance, a pattern consistent with sensitivity to image presence and limited use of diagnostic image content. We explore this preamble-associated component with a phrase-level probe and a false-modality ablation on smaller models. We relate the resulting pattern to the broader modality-collapse phenomenon~\cite{parcalabescu-frank-2023-mm, sim-etal-2025-vlms} in a clinical setting, as well as prior work on prompt sensitivity~\cite{lu-etal-2024-prompts, ismithdeen-etal-2025-promptception} and priming effects~\cite{jones2022capturing, yoshida-etal-2025-visual}. We run a preliminary case study with a domain expert, analyzing reasoning traces produced by \texttt{Qwen2.5-\allowbreak{}VL-\allowbreak{}3B}, the model that exhibits the largest preamble-only shift. In the reviewed traces, the shift is accompanied by justifications containing unverified clinical details. Finally, we test preference alignment on \texttt{Qwen2.5-\allowbreak{}VL-\allowbreak{}3B} and find that it suppresses MRI-referencing behavior and removes most of the preamble-associated advantage, while performance in the augmented conditions also declines. Our contributions are:

\begin{itemize}[leftmargin=*]
  \item We evaluate 12 open-weight VLMs on \textsc{FOR2107} \cite{kircher2019-db} and \textsc{OASIS-3} \cite{LaMontagne2019}, two cohorts for which prior work does not establish the included neuroimaging inputs as reliable stand-alone diagnostic evidence for the present tasks.
  \item Confidence estimation shows that most of the calibration improvement in the analyzed smaller models occurs after the textual preamble is added; a phrase-level probe and false-modality ablation further examine the accompanying prediction shift.
  \item A preliminary expert case study with a clinical psychologist finds higher mean ratings for traces generated with multimodal context than for text-only traces, while mean faithfulness remains below 3.0 in every group examined.
  \item A preference-alignment intervention suppresses MRI-referencing behavior, while performance declines in the augmented conditions.
\end{itemize}

We analyze this phenomenon at three levels: \emph{predictive} (F1 across conditions, Section~\ref{subsec:results}), \emph{behavioral} (confidence shifts and reasoning traces, Sections~\ref{sec:confidence}--\ref{sec:multimodal_inference}), and \emph{representational} (the preamble-associated hidden-state shift and its response to alignment, Sections~\ref{sec:preamble}--\ref{sec:dpo}). Together, these analyses provide complementary views of a large performance gain that is not fully explained by the added diagnostic evidence under our controls. Surface evaluations can therefore overstate VLM diagnostic ability in clinical settings and overlook problems that matter for deployment.

\begin{figure*}[tp]
  \centering
  \includegraphics[width=\linewidth]{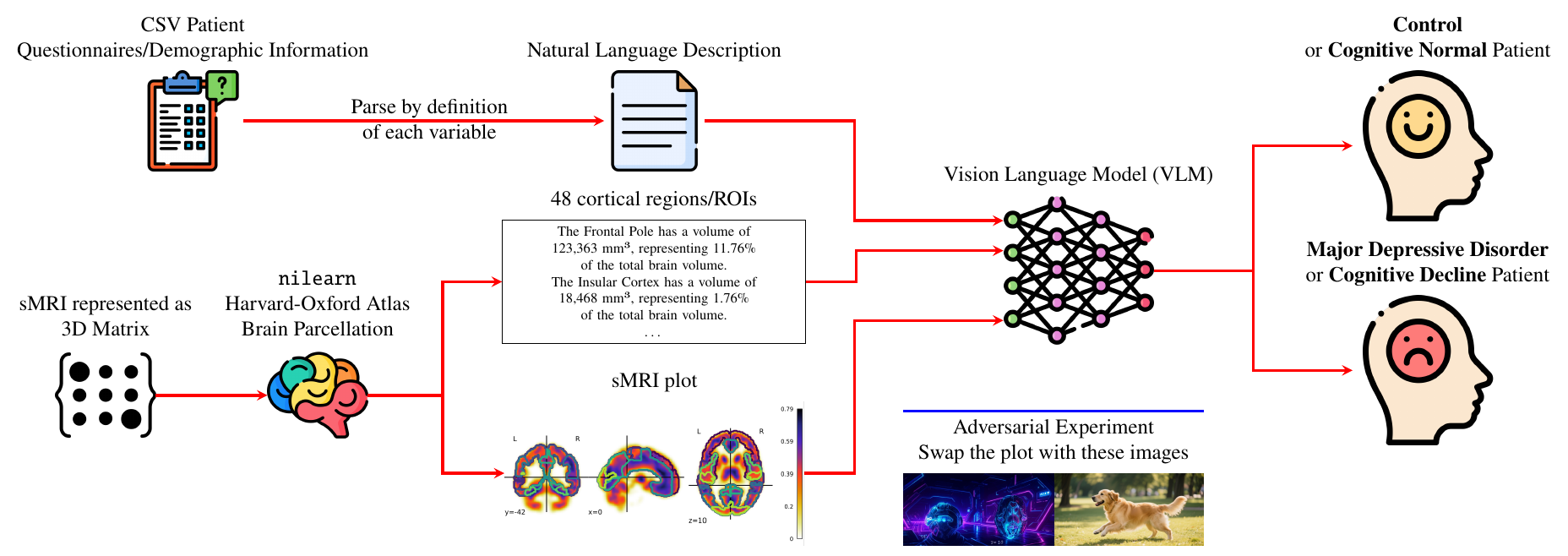}
  \caption{Overview of the VLM evaluation pipeline.}
  \label{fig:vlm-pipeline}
\end{figure*}

\section{Related Work}
We review related work on prompt sensitivity, priming effects, and modality collapse in vision-language models (VLMs).

\subsection{Prompt Sensitivity in VLMs}
While prompt sensitivity has been extensively studied in large language models (LLMs)~\cite{ishibashi-etal-2023-evaluating, lu-etal-2024-prompts}, it remains underexplored in VLMs. Among the few studies on this, \citet{ismithdeen-etal-2025-promptception} demonstrates that VLMs suffer from severe prompt sensitivity that leads to inconsistent classification performance and a concerning reliance on language priors.

In healthcare, prompt sensitivity to instruction phrasing can severely impact reliability: prompt variations cause F1 fluctuations of up to 0.25 in LLMs~\cite{ceballos-arroyo-etal-2024-open} and accuracy variations of up to 6\% in VLMs on health benchmarks~\cite{ismithdeen-etal-2025-promptception}.

\subsection{Priming Effect in VLMs}
The priming effect is a cognitive phenomenon in which prior exposure to a stimulus
influences subsequent judgments or behaviors, often without conscious awareness~\cite{Meyer1971FacilitationIR}. This effect can arise through the activation of related concepts, associations, or response tendencies that shape how later inputs are interpreted, even when no explicit alternative description is presented~\cite{koo-etal-2024-benchmarking}.

In VLMs, priming refers to how textual and visual cues shape model outputs, mirroring the priming concept from cognitive science. \citet{yoshida-etal-2025-visual} directly investigated this phenomenon in large-scale VLMs and found that accompanying images systematically shift model responses in the intended direction, suggesting that visual information can play an active role in language processing. Recent research has further revealed both the power and limitations of priming in adapting VLMs to new tasks and domains~\cite{jones2022capturing, koo-etal-2024-benchmarking}. As these models increasingly incorporate multimodal inputs, such priming effects can be further intensified by interactions between textual and visual signals~\cite{Gulati__Sebe_Lepri_Oliver_2025, zhang-etal-2026-images}. For instance, \citet{zhang-etal-2026-images} showed that in fact-checking settings, VLMs may favor previously introduced textual context over salient visual evidence, and related behaviors have been observed in general visual question answering (VQA) tasks~\cite{shu2025when}.

\subsection{Modality Collapse in VLMs}
Modality collapse refers to the phenomenon where a unimodal model achieves comparable accuracy to a multimodal model on a vision-language task, revealing that one modality is not meaningfully utilized~\cite{pmlr-v162-javaloy22a, parcalabescu-frank-2023-mm}. A recent survey by \citet{sim-etal-2025-vlms} provides a systematic taxonomy of contributing factors, including dataset bias, model behavior, lack of fine-grained supervisory signals, and task setup, and reviews methods for quantifying modality contribution and cross-modal interaction. A consistent finding across this literature is that the text modality dominates, with visual input failing to influence predictions even when it is ostensibly required by the task~\cite{zhu-etal-2022-diagnosing}. In the clinical domain, a related failure mode is known as ``\emph{shortcut learning}'': models often rely on imaging artifacts or dataset features rather than actual pathology~\cite{degrave-etal-2021-shortcut, geirhos-etal-2020-shortcut}.

Previous work on modality collapse shows that text often dominates. We instead characterize which textual cues are associated with this collapse-like pattern in a clinical setting and show that the tested preference-alignment intervention suppresses MRI references without preserving augmented-condition performance, leaving the underlying mechanism unresolved.

\section{Methodology}
\label{sec:methodology}
Our pipeline (see Figure~\ref{fig:vlm-pipeline}) integrates two primary modalities: \textbf{tabular clinical data stored as comma-separated values} (CSV) and \textbf{structural brain MRI (sMRI)} scans. It has three sequential phases:
\begin{enumerate}[topsep=0pt, partopsep=0pt, parsep=0pt, itemsep=0pt, leftmargin=*]
    \item \textbf{Clinical Tabular Data Serialization:} We serialize participant-level CSV data into natural language by mapping each variable's value to its clinical description. This transformation has been shown to substantially improve comprehension and reasoning in language models~\cite{hegselmann2023tabllm, vu2025roleplaying}.
    \item \textbf{sMRI Information Extraction:}\footnote{T1-weighted sMRI scans in our cohorts are stored as 3D volumes in \texttt{.nii} or \texttt{.nii.gz} format, with common dimensions of $256 \times 256 \times 176$-$208$ voxels.} Following previous work~\citep{kapoor2025computer, Galteau2025, 10.1162/IMAG.a.157}, we use \texttt{nilearn}\footnote{\url{https://github.com/nilearn/nilearn}} and the Harvard-Oxford probabilistic cortical atlas~\cite{havard-oxford} to visualize sMRI volumes and extract regional anatomical measurements via brain parcellation. Per-region volumetric measurements are serialized into descriptive text following the same approach as the tabular data. For visualization, we render three orthogonal slices (sagittal, coronal, axial) annotated with Montreal Neurological Institute (MNI) coordinates, hemisphere labels, and voxel contour overlays.
    \item \textbf{Multimodal Prediction:} The serialized text, parcellation descriptions, and MRI visualizations are combined in a structured chat template and provided to a VLM for binary classification. Input components can be included or removed at inference time, which defines the five conditions we evaluate (Section~\ref{sec:conditions}). The full prompt template is shown in Appendix~\ref{app:prompt}.
\end{enumerate}

\section{Experiment}
\begin{table*}[tp]
    \centering
    \small
    \resizebox{\linewidth}{!}{%
    \begin{tabular}{clccccl}
    \toprule
    \textbf{Cond.} & \textbf{Input} & \textbf{CSV} & \textbf{Prompt(MRI)} & \textbf{Parcel.} & \textbf{Image} & \textbf{Image Type} \\
    \midrule
    C1
      & \textsc{text(csv)}
      & \checkmark & & & &  \\
    C2
      & \textsc{text(csv)} $+$ \textsc{prompt(mri)}
      & \checkmark & \checkmark & & &  \\
    C3
      & \textsc{text(csv)} $+$ \textsc{prompt(mri)} $+$ \textsc{plot(MRI)}
      & \checkmark & \checkmark & & \checkmark & \texttt{nilearn} brain plot \\
    C4
      & \textsc{text(csv, parcel)} $+$ \textsc{prompt(mri)} $+$ \textsc{plot(MRI)}
      & \checkmark & \checkmark & \checkmark & \checkmark & \texttt{nilearn} brain plot \\
    C5
      & \textsc{text(csv, parcel)} $+$ \textsc{prompt(mri)} $+$ \textsc{plot(swap)}
      & \checkmark & \checkmark & \checkmark & \checkmark & OOD image (ablation) \\
    \bottomrule
    \end{tabular}}
    \caption{Experimental conditions and their input components. Conditions are designed to vary the supplied input components: C1 uses tabular text only, C2 adds the MRI preamble without imaging data, C3--C4 progressively add MRI content, and C5 replaces the subject-specific MRI plot with an out-of-distribution control image to test whether performance depends on image content or merely image presence.}\label{tab:conditions}
\end{table*}

\paragraph{Datasets}
We evaluate our pipeline on two clinical neuroimaging datasets:

\textsc{\textbf{FOR2107}} is a German multicenter cohort study on the neurobiology of affective disorders~\cite{kircher2019-db}. It includes patients with Major Depressive Disorder (MDD) and matched controls, with deep phenotyping spanning structural MRI, clinical assessments, neuropsychological testing, and demographic information. Prior work on this cohort reports classification accuracies of only $\approx$56\% with univariate neuroimaging markers~\cite{winter2022} and a maximum reported accuracy of 62\% with multivariate machine learning (ML) over a 4M-model search across structural and functional MRI, diffusion tensor imaging (DTI), a polygenic risk score, and environmental risk factors~\cite{winter2024}. \textsc{FOR2107} therefore provides a challenging classification setting in which performance gains warrant scrutiny. Our binary task distinguishes \textit{Active MDD} from \textit{Control}.

\begin{table}[htbp]
    \centering
    \small
    \begin{tabular}{llr}
        \toprule
        \textbf{Dataset} & \textbf{Group / Status} & \textbf{\# Samples} \\
        \midrule
        \multirow{2}{*}{FOR2107} & Active MDD    & 701   \\
                                 & Control       & 1,071 \\
        \cmidrule{1-3}
        \multirow{2}{*}{OASIS-3} & Cognitive Decline  & 487 \\
                                 & Cognitive Normal   & 849 \\
        \bottomrule
    \end{tabular}
    \caption{Class distribution of the clinical subsets used in this work.}
    \label{tab:dataset_summary}
\end{table}

\textsc{\textbf{OASIS-3}} is a longitudinal dataset from the Washington University Knight Alzheimer Disease Research Center, available through a controlled-access application process and data-use agreement~\cite{LaMontagne2019}. It includes participants ranging from cognitively normal adults to individuals at various stages of cognitive decline, accompanied by multimodal MRI sessions and clinical assessments.
Prior work does not support structural MRI as a stand-alone diagnostic test for this task~\citep{Lombardi2020-oj, Ansart2020-mk}; accordingly, we do not treat it as such on \textsc{OASIS-3}.
To match the classification setup with \textsc{FOR2107}, we focus on the target labels \textit{Cognitive Decline} and \textit{Cognitive Normal}.

Table~\ref{tab:dataset_summary} summarizes the class distributions of the two datasets. Both datasets are governed by strict data-use agreements that prohibit redistribution and require formal application for access. 
To our knowledge, neither has appeared in any natural language processing (NLP) or VLM publication. This may reduce, but does not eliminate, the risk of direct benchmark contamination \citet{balloccu-etal-2024-leak} relative to public benchmarks such as the subsets of 86--160 samples used by \citet{ceballos-arroyo-etal-2024-open} or the 197 Health and Medicine examples in \citet{ismithdeen-etal-2025-promptception}. Both \textsc{FOR2107} and \textsc{OASIS-3} contain many clinical variables drawn from multiple CSV files. Some of these variables provide obvious classification shortcuts (e.g., suicidal thoughts). In consultation with domain experts in clinical psychology (Appendix~\ref{app:experts}), we excluded these features so that the tested VLMs could not use them to complete the task trivially (see Appendix~\ref{app:for2107-variables} and Appendix~\ref{app:oasis-3-variables} for the full list of variables used in this work).

\begin{figure*}[tp]
    \centering
    \includegraphics[width=\linewidth]{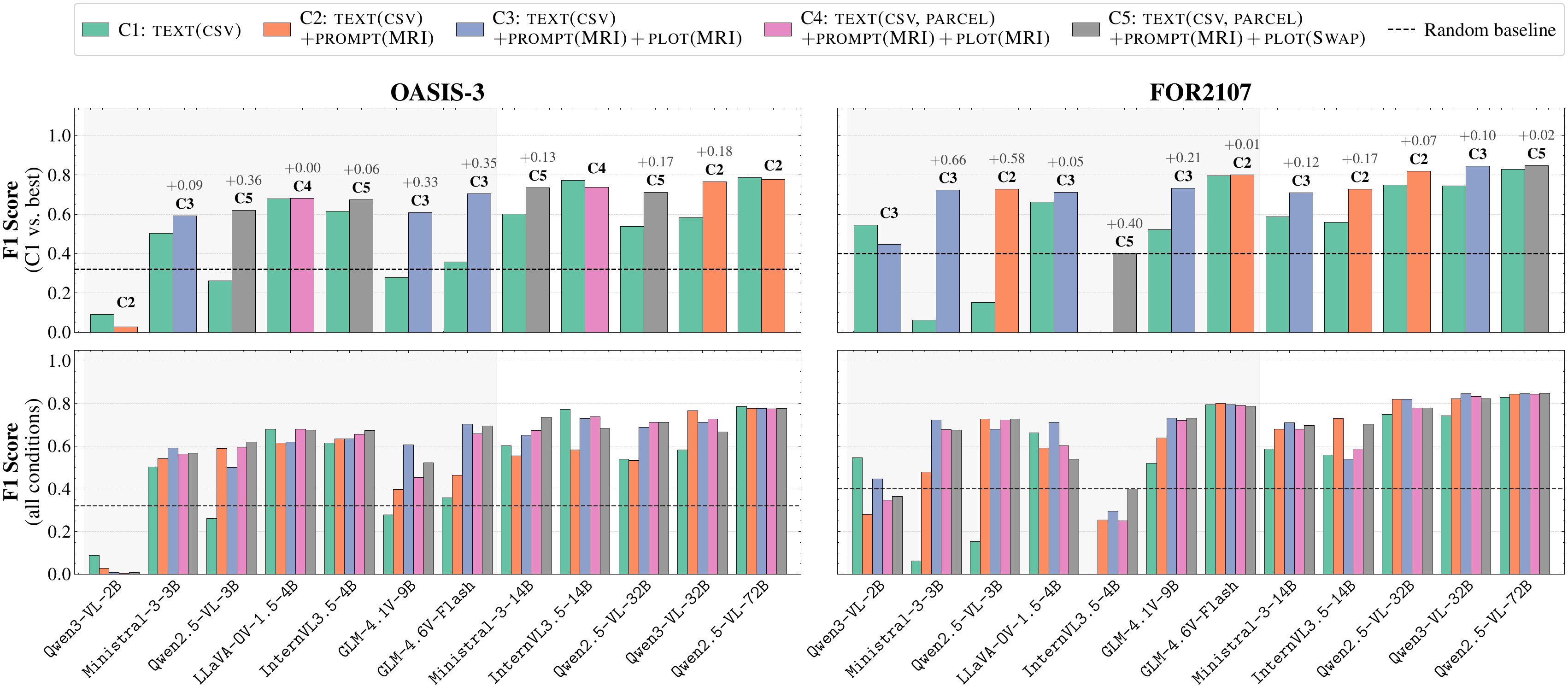}
\caption{F1 scores on both datasets for the 12 tested VLMs. \textbf{Top row}: text-only baseline (C1, teal-green) versus the best augmented setup per model (best of C2--C5). \textbf{Bottom row}: full breakdown across all five input conditions. Models are sorted from smallest to largest (parameter count).}
    \label{fig:f1}
\end{figure*}

\paragraph{Models}\label{sec:models}
We test 12 open-weight vision-language models\footnote{We also conduct ablation experiments on the medically specialized VLM \texttt{medgemma} \cite{sellergren2026medgemmatechnicalreport}, which exhibits failure modes (Appendix~\ref{app:medgemma}).}: \texttt{InternVL3.5-\allowbreak{}4B} and \texttt{14B} \cite{wang2025internvl3}, \texttt{GLM-\allowbreak{}4.1V-\allowbreak{}9B} \cite{hong2025glm}, \texttt{GLM-\allowbreak{}4.6V-\allowbreak{}Flash} \cite{zeng2025glm}, \texttt{LLaVA-\allowbreak{}OV-\allowbreak{}1.5-\allowbreak{}4B} \cite{li2025llavaonevision}, \texttt{Ministral-\allowbreak{}3-\allowbreak{}3B} and \texttt{14B} \cite{liu2026ministral}, \texttt{Qwen2.5-\allowbreak{}VL-\allowbreak{}3B}, \texttt{32B} and \texttt{72B} \cite{xu2025qwen25vltechnicalreport}, \texttt{Qwen3-\allowbreak{}VL-\allowbreak{}2B} and \texttt{32B} \cite{bai2025qwen3}. For all models, we use the Hugging Face Transformers~\cite{wolf-etal-2020-transformers} implementation\footnote{For the full model names, see Table~\ref{tab:hf_models} in the Appendix.}, setting \texttt{do\_sample=False} to ensure deterministic, reproducible outputs across runs. To contextualize model performance, we also include a stratified random baseline on both datasets (derivation details in Appendix~\ref{app:random_baseline}).

\paragraph{Input Conditions}\label{sec:conditions}
To evaluate how models respond to neuroimaging context that is not established as reliable stand-alone diagnostic evidence, we define five conditions that vary the MRI-related framing and supplied input components, summarized in Table~\ref{tab:conditions} (Appendix~\ref{app:prompt} provides the exact prompts):
\begin{itemize}[leftmargin=*]
    \item \textsc{text(csv)} consists of serialized clinical and demographic variables from the participant record.
    \item \textsc{prompt(mri)} adds an MRI-availability preamble that mentions brain-parcellation volumes and visualization of brain regions without providing either component.
    \item \textsc{text(parcel)} includes a textual representation of the brain parcellation data.
    \item \textsc{plot(MRI)} adds the rendered MRI plot as visual input.
    \item \textsc{plot(swap)} replaces the brain plot with an out-of-distribution image (a dog photograph or a stylized non-clinical brain rendering; see Appendix~\ref{app:ood_images}). Because C5 results are not stratified by image variant, we use this condition only as a broad control for image presence and do not compare the two variants.
\end{itemize}

\subsection{Experimental Results}\label{subsec:results}
Figure~\ref{fig:f1} reports F1 across all models and conditions. Under the text-only baseline (C1), several smaller models fail to exceed the random baseline, most strikingly on \textsc{FOR2107}, where \texttt{Ministral-\allowbreak{}3-\allowbreak{}3B} (0.064) and \texttt{Qwen2.5-\allowbreak{}VL-\allowbreak{}3B} (0.153) fall far below it.

This pattern changes once MRI-related context is added: Figure~\ref{fig:f1} shows that smaller models gain the most, with \texttt{Qwen2.5-VL-3B} and \texttt{Ministral-3-3B} improving by up to $+0.58$ and $+0.66$ F1 on \textsc{FOR2107}, while larger counterparts change little. The swap-image condition (C5) shows that replacing the MRI plot with an unrelated image does not consistently reduce performance. Qualitative inspection suggests that adding neuroimaging context changes the generated justifications even when classification performance is unchanged: larger models produce MRI-referencing justifications under multimodal input despite stable F1.

\begin{figure*}[tp]
    \centering
    \begin{subfigure}{0.49\textwidth}
        \centering
        \includegraphics[width=\linewidth]{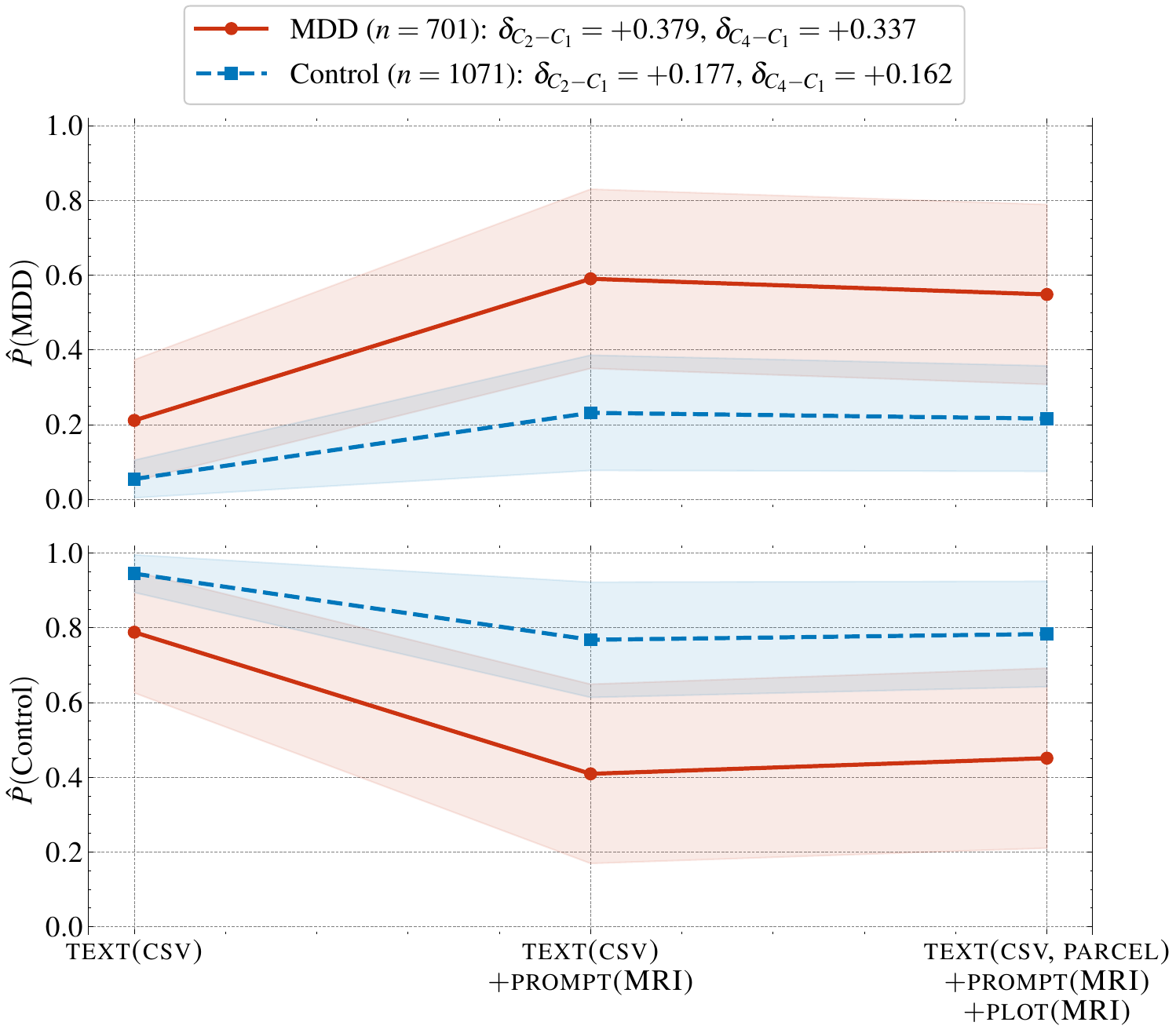}
        \caption{\texttt{Qwen2.5-VL-3B}}
        \label{fig:first}
    \end{subfigure}
    \hfill
    \begin{subfigure}{0.49\textwidth}
        \centering
        \includegraphics[width=\linewidth]{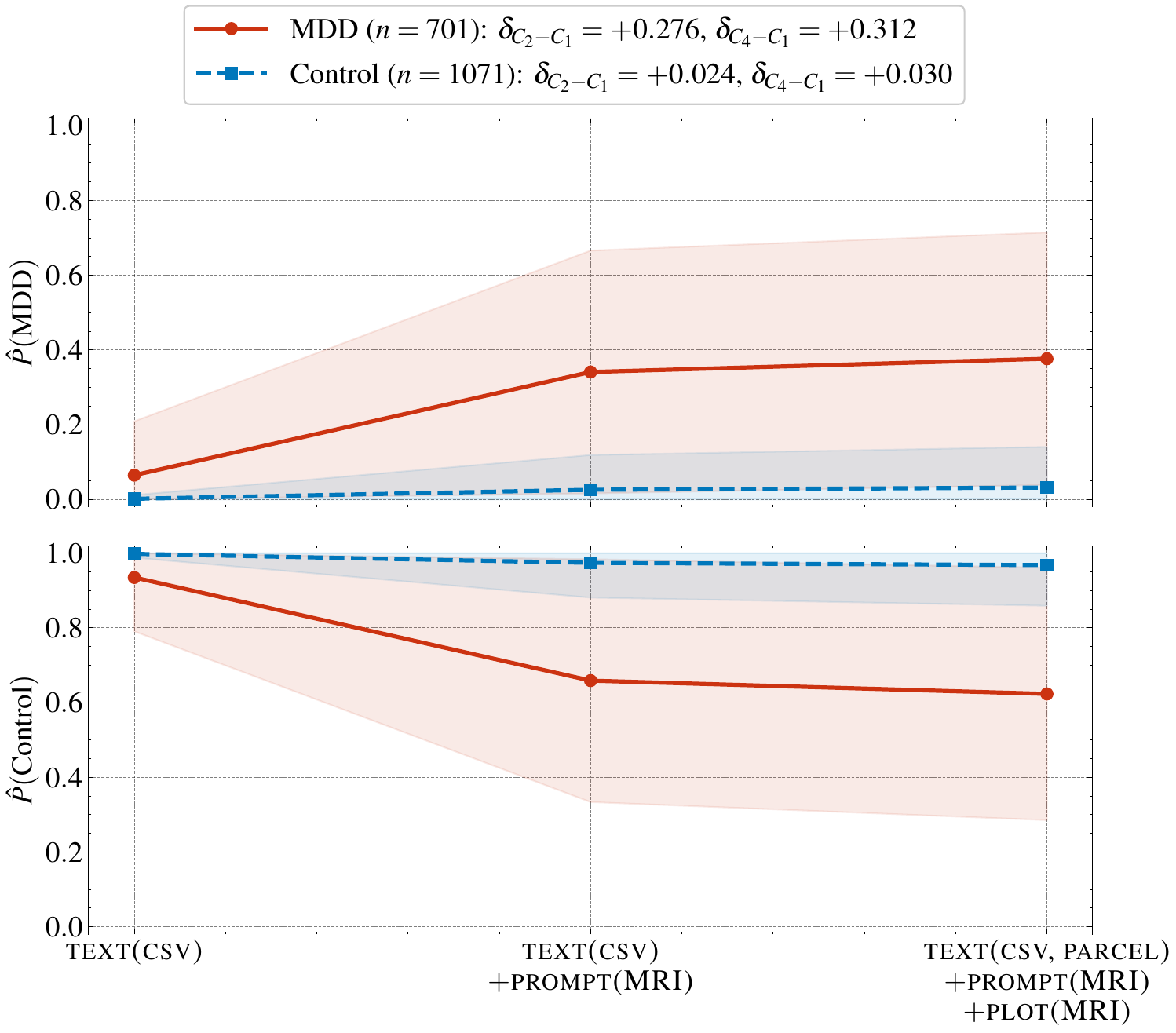}
        \caption{\texttt{Ministral-3-3B}}
        \label{fig:second}
    \end{subfigure}
    \caption{\textsc{FOR2107} per-group mean $\hat{P}(\text{MDD})$ (top) and $\hat{P}(\text{Control})$ (bottom) across the three input conditions. Both diagnostic groups exhibit an upward shift in $\hat{P}(\text{MDD})$; the corresponding decrease in $\hat{P}(\text{Control})$ is implied by two-class normalization. Shaded bands show one sample standard deviation around each group mean and are clipped to the $[0,1]$ probability range.}
    \label{fig:dual}
\end{figure*}

\section{Confidence Estimation}\label{sec:confidence}
Since the smaller models exhibit large F1 increases, we further examine the label probabilities they assign. For each participant, we compute a confidence score from the candidate label-token probabilities after the fixed JSON prefix, renormalized across $\{\text{MDD}, \text{Control}\}$ (formal definition in Appendix~\ref{app:confidence}). This formulation follows label-token probability extraction approaches~\cite{pmlr-v139-zhao21c, geng-etal-2024-survey, xiong2024can}. From these scores we measure \emph{calibration}~\citep{xiong2024can}: the degree to which the model's confidence agrees with its actual classification accuracy, captured by Expected Calibration Error (ECE)~\cite{Pakdaman_Naeini_Cooper_Hauskrecht_2015, pmlr-v70-guo17a} and the Brier score~\cite{Brier1950}.

We report this analysis on \texttt{Qwen2.5-VL-3B} and \texttt{Ministral-3-3B}, the two smaller models with the largest C1$\to$C2 gains on \textsc{FOR2107} and therefore the clearest cases for examining where the gains occur. We extend the analysis to \texttt{InternVL3.5-4B} and \texttt{GLM-4.1V-9B} in Appendix~\ref{app:conf-more}, where we find the same pattern. 

\begin{table}[htbp]
    \small
    \centering
    \begin{tabular}{ccccc}
    \toprule
    \multirow{2}{*}{\textbf{Cond.}}
    & \multicolumn{2}{c}{{\texttt{Qwen2.5-VL-3B}}}
    & \multicolumn{2}{c}{{\texttt{Ministral-3-3B}}} \\
    \cmidrule(lr){2-3} \cmidrule(lr){4-5}
    & \textbf{ECE\,$\downarrow$}
    & \textbf{Brier\,$\downarrow$}
    & \textbf{ECE\,$\downarrow$}
    & \textbf{Brier\,$\downarrow$} \\
    \midrule
    C1 & $0.376$          & $0.370$          & $0.361$          & $0.354$ \\
    C2 & $0.164$          & $\mathbf{0.178}$ & $0.199$          & $0.219$ \\
    C4 & $\mathbf{0.155}$ & $0.184$          & $\mathbf{0.183}$ & $\mathbf{0.207}$ \\
    \bottomrule
    \end{tabular}%
    \caption{Expected calibration error (ECE; 15 equal-width bins) and Brier score by condition. Lower ECE and Brier indicate better calibration. \textbf{Bold} indicates the best score per model.}
    \label{tab:calibration}
\end{table}

Table~\ref{tab:calibration} reports ECE and Brier. For both models, C2 and C4 are more accurate and have lower ECE and Brier scores than C1. However, the shared upward shift toward MDD across both diagnostic groups indicates that this improvement can coexist with a label-directional prompt bias. Most of the ECE improvement occurs under the MRI prompt alone (C2), before any imaging content is provided: the C1$\to$C2 transition accounts for the majority of the total C1$\to$C4 ECE reduction for both models. This pattern is consistent with a substantial prompt-framing contribution but does not identify the underlying mechanism.

Figure~\ref{fig:dual} is also consistent with the ECE and Brier results: for both models, $\hat{P}(\text{MDD})$ increases under the two plotted preamble-bearing conditions, C2 and C4, in both the MDD and Control groups. Because confidence is normalized across two labels, the corresponding decrease in $\hat{P}(\text{Control})$ follows by construction. The increase in the Control group is inconsistent with a shift driven solely by improved diagnostic discrimination and is consistent with a prompt-framing contribution. We examine this preamble-associated pattern in detail in Section~\ref{sec:preamble}.

\section{Preliminary Expert Case Study of Reasoning Traces}\label{sec:multimodal_inference}
To examine whether the large F1 gains under multimodal context are accompanied by better reasoning-trace quality, we conducted a preliminary human expert case study of reasoning traces generated by \texttt{Qwen2.5-VL-3B}, the model exhibiting the largest C1$\to$C2 preamble-only shift.

\begin{table}[tp]
    \centering
    \small
    \resizebox{\linewidth}{!}{%
    \begin{tabular}{clccc}
        \toprule
        \textbf{Grp.} & \textbf{Cond., Class}
        & \textbf{Faith.}$\,\uparrow$
        & \textbf{Clin.\ Acc.}$\,\uparrow$ & \textbf{Reasoning}$\,\uparrow$ \\
        \midrule
        0 & C1, Control & $2.17$ & $2.25$ & $2.46$ \\
        1 & C1, MDD     & $\underline{2.08}$ & $\underline{2.17}$ & $\underline{2.17}$ \\
        \midrule
        2 & C4, Control & $2.79$ & $\mathbf{2.63}$ & $\mathbf{2.96}$ \\
        3 & C4, MDD     & $\mathbf{2.83}$ & $2.46$ & $2.83$ \\
        \bottomrule
    \end{tabular}}
    \caption{Mean ratings by a clinical psychologist of reasoning traces generated by \texttt{Qwen2.5-VL-3B}, using a 1--4 scale with 0.5-point increments ($n = 12$ traces per group, \textsc{FOR2107}). \textbf{Bold} indicates the best score and \underline{underlining} indicates the worst score in each metric column. See Table~\ref{tab:conditions} for a description of the notation.}
    \label{tab:human_eval}
\end{table}

\paragraph{Case study protocol}
We sampled 12 predictions for each class (MDD vs.\ Control) from the \textsc{FOR2107} dataset under two conditions: the baseline \textsc{text(csv)}, which includes only the serialized clinical record, and the full multimodal condition \textsc{text(csv, parcel)} $+$ \textsc{prompt(mri)} $+$ \textsc{plot(MRI)}, which additionally incorporates brain parcellation text and a \texttt{nilearn}-generated MRI visualization, yielding 24 reasoning traces per condition (48 total). A clinical psychologist evaluated each trace against three criteria (Faithfulness, Clinical Accuracy \& Safety, Diagnostic Reasoning) using a 1--4 rating scale permitting 0.5-point increments. The full scoring rubric appears in Appendix~\ref{app:human_eval}.

\begin{table*}[tp]
    \centering
    \small
    \resizebox{\linewidth}{!}{%
    \begin{tabular}{clcccc}
    \toprule
    \multirow{2}{*}{\textbf{Cond.}}
      & \multirow{2}{*}{\textbf{Input}}
      & \multicolumn{2}{c}{\textsc{FOR2107}}
      & \multicolumn{2}{c}{\textsc{OASIS-3}} \\
    \cmidrule(lr){3-4} \cmidrule(lr){5-6}
    & & \texttt{Qwen2.5-VL-3B} & \texttt{Ministral-3-3B}
      & \texttt{Qwen2.5-VL-3B} & \texttt{Ministral-3-3B} \\
    \midrule
    C1
      & \textsc{text(csv)}
      & 0.153 & 0.064 & 0.262 & 0.504 \\
    C2
      & \textsc{text(csv)} $+$ \textsc{prompt(mri)}
      & \textbf{0.728} & \textbf{0.480} & \textbf{0.589} & 0.541 \\
    C2\rlap{$_{\dagger}$}
      & \textsc{text(csv)} $+$ \textsc{prompt(fmri)}
      & 0.702 & 0.361 & 0.379 & \textbf{0.569} \\
    C2\rlap{$_{\ddagger}$}
      & \textsc{text(csv)} $+$ \textsc{prompt(weather)}
      & 0.056 & 0.031 & 0.148 & 0.425 \\
    \bottomrule
    \end{tabular}}
    \caption{F1 under two preamble ablations---a factually false fMRI mention and a semantically irrelevant weather preamble---evaluated with \texttt{Qwen2.5-VL-3B} and \texttt{Ministral-3-3B} across both cohorts. \textbf{Bold} indicates the best score for each model and cohort.}
    \label{tab:preamble_ablation}
\end{table*}

\paragraph{Results}
Table~\ref{tab:human_eval} reports results for each criterion across the four groups. Every group has mean scores below $3.0$ on all three criteria. Faithfulness scores indicate recurring unverified clinical details, including when multimodal context is available. Traces under the full multimodal condition (Groups 2 and 3) score higher on all three criteria than traces under the baseline condition (Groups 0 and 1). Group 3 (C4, MDD) reaches the highest faithfulness score of $2.83$, while Group 2 (C4, Control) reaches the highest clinical accuracy and diagnostic reasoning scores, at $2.63$ and $2.96$, respectively. These higher ratings do not show that the model uses the supplied evidence faithfully.

\section{Preamble-Associated Prediction Shifts}
\label{sec:preamble}
While modality collapse is well documented~\cite{sim-etal-2025-vlms}, to our knowledge, prior work has not yet investigated which textual cues are associated with a collapse-like pattern in a clinical setting. Section~\ref{sec:confidence} shows that most of the C1$\to$C4 ECE reduction occurs during the C1$\to$C2 transition, when the preamble is added without imaging content. We now ask which types of phrases are associated with this shift and whether the effect is specific to the exact preamble wording used in our pipeline. We first characterize this association using \texttt{Qwen2.5-VL-3B} on \textsc{FOR2107}, then use an ablation to test whether the qualitative contrast recurs with \texttt{Ministral-3-3B} and on \textsc{OASIS-3}. For more formal details, we refer to the Appendix.~\ref{app:preamble}.

\begin{figure}[htbp]
    \centering
    \includegraphics[width=\linewidth]{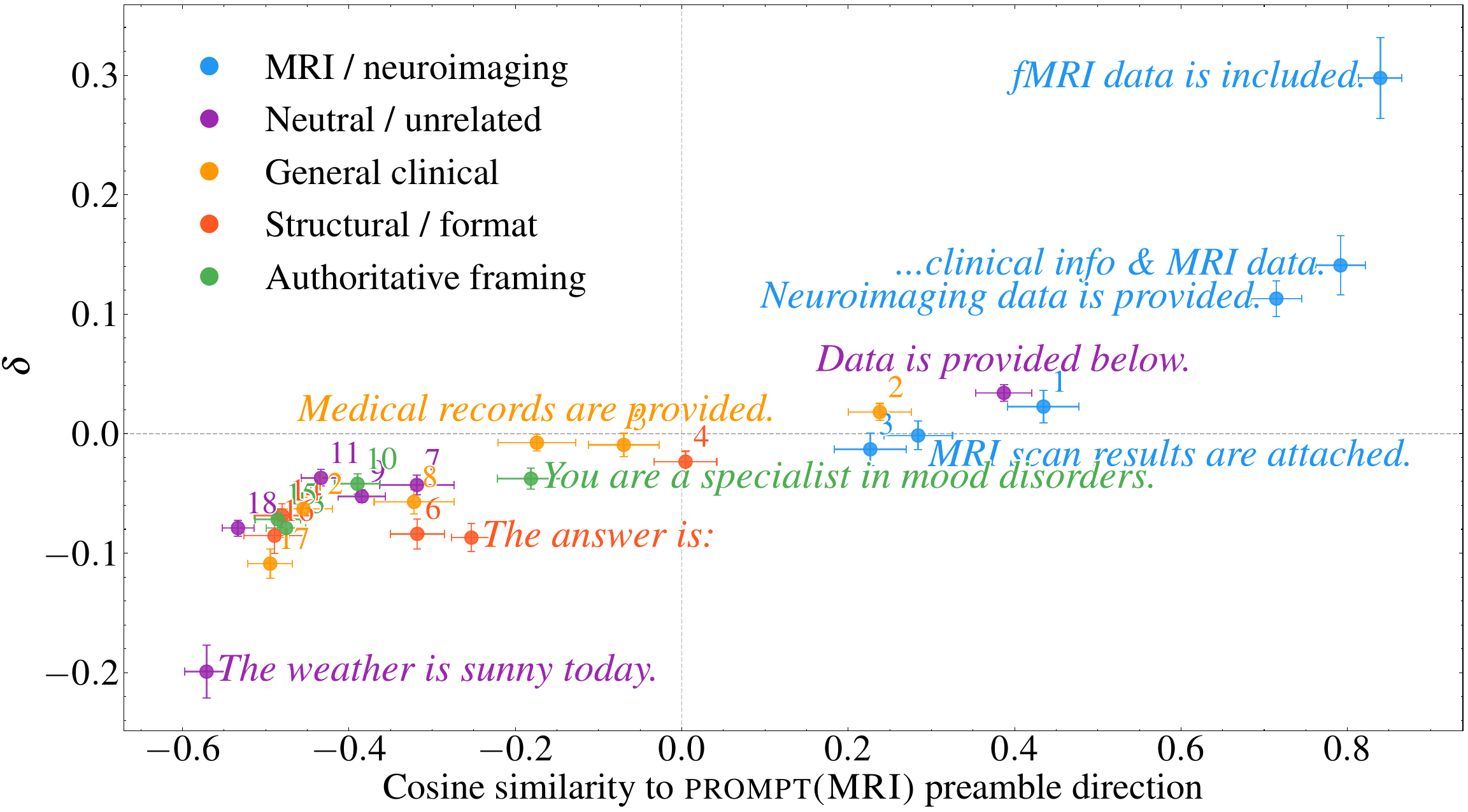}
    \caption{Shift in mean $\hat{P}(\text{MDD})$ relative to the \textsc{text(csv)} baseline (vertical) versus cosine similarity to the MRI-preamble direction (horizontal), for candidate phrases across semantic categories, evaluated with \texttt{Qwen2.5-VL-3B} on 15 participants from the \textsc{FOR2107} MDD group. Error bars show one standard deviation across participants. Phrases in the top-right quadrant are associated with both a preamble-aligned hidden-state shift and increased mean $\hat{P}(\text{MDD})$ without providing imaging data.}
    \label{fig:phrase_probe}
\end{figure}

\paragraph{Phrase probe}
We evaluate 27 candidate preamble phrases spanning five semantic categories: \textit{MRI/neuroimaging, general clinical, authoritative framing, neutral, structural/format}. Selected phrases are written out directly in Figure~\ref{fig:phrase_probe}, while Table~\ref{tab:preamble_phrases} lists the phrases and scores corresponding to the 19 numbered points. For each probe, we replace the original \textsc{prompt(mri)} with the candidate phrase, evaluate the \textsc{FOR2107} MDD group with \texttt{Qwen2.5-VL-3B}, and measure how much the phrase shifts $\hat{P}(\text{MDD})$ relative to the \textsc{text(csv)} baseline. We additionally compute the cosine similarity of the hidden-state shift associated with each phrase to that of the original MRI preamble. Figure~\ref{fig:phrase_probe} plots both quantities jointly.

The phrase \emph{fMRI data is included} is the point farthest toward the top right in Figure~\ref{fig:phrase_probe}. Although several clinical-assessment and format/neutral phrases fall at or below zero on one or both axes, the neutral phrase \emph{Data is provided below} is a notable positive exception. Within this exploratory probe, announcement-style claims that a result is available are most strongly associated with the measured shift, with MRI availability providing the strongest example.

\paragraph{False-modality ablation}
To test whether the shift occurs with a surface-level phrase alone and extends beyond the model and dataset used in the phrase probe, we compare two conditions: a false statement that fMRI data are available, with no imaging supplied, and a semantically irrelevant weather preamble used as a non-clinical negative control. We evaluate both on \texttt{Qwen2.5-VL-3B} and \texttt{Ministral-3-3B} across \textsc{FOR2107} and \textsc{OASIS-3}. Table~\ref{tab:preamble_ablation} shows that, on \textsc{FOR2107}, the fMRI preamble yields F1 scores of 0.702 and 0.361 for \texttt{Qwen2.5-VL-3B} and \texttt{Ministral-3-3B}, respectively, whereas the weather preamble yields 0.056 and 0.031. Although its magnitude varies by model, the fMRI-over-weather difference is large in this cohort and also appears on \textsc{OASIS-3}, where the weather condition underperforms the fMRI preamble for both models. Appendix~\ref{app:preamble_extended} reports the same comparison for \texttt{InternVL3.5-4B} and \texttt{GLM-4.1V-9B} across both cohorts, extending the qualitative pattern to those additional model families.

\section{Preference Learning Intervention} \label{sec:dpo}
In its justifications, \texttt{Qwen2.5-\allowbreak{}VL-\allowbreak{}3B} frequently referred to MRI-derived or brain-parcellation features, and the preliminary expert case study also identified unverified clinical details in the reviewed traces. To suppress the MRI-referencing language, we use Direct Preference Optimization (DPO)~\cite{rafailov2023direct} through its multimodal extension, Mixed Preference Optimization (MPO)~\cite{wang2024enhancing}.

\paragraph{Dataset construction}
We construct a preference dataset from \textsc{OASIS-3} using the outputs of the 12 models described in Section~\ref{sec:models}. Source responses are drawn from three input modes: \textsc{text(csv)}, \textsc{text(csv)} $+$ \textsc{prompt(mri)}, \textsc{text(csv, parcel)} $+$ \textsc{prompt(mri)} $+$ \textsc{plot(MRI)}. For every subject, we define the \emph{chosen} response as the \textsc{text(csv)} mode output of any model that predicted the correct label, and the \emph{rejected} response as any output, from any mode, whose text explicitly mentions MRI findings, brain parcellation, or related neuroimaging language. Since rejected responses are defined by MRI-referencing language rather than independently verified fabrication, the objective targets that surface cue by design. The final dataset comprises 17,202 preference pairs balanced across the two target classes: \emph{Cognitive Normal} and \emph{Cognitive Decline} ($\approx$50\% each).

\begin{table}[htbp]
    \centering
    \small
    \begin{tabular}{ccc}
    \toprule
    \textbf{Cond.} & \textbf{Before MPO}$\,\uparrow$ & \textbf{After MPO}$\,\uparrow$ \\
    \midrule
    C1 & 0.153 & 0.459 \\
    C2 & \textbf{0.728} & \textbf{0.496} \\
    C4 & 0.723 & 0.477 \\
    \bottomrule
    \end{tabular}
    \caption{F1 of \texttt{Qwen2.5-VL-3B} on \textsc{FOR2107} before and after MPO. Random baseline F1\,=\,0.40. \textbf{Bold} indicates the best score. See Table~\ref{tab:conditions} for a description of the notation.}
    \label{tab:dpo_results}
\end{table}

\paragraph{Evaluation and results}
We fine-tune \texttt{Qwen2.5-\allowbreak{}VL-\allowbreak{}3B} using MPO and report the training details in Appendix~\ref{app:preference}. We then evaluate the trained model on \textsc{FOR2107}, a cohort distinct from \textsc{OASIS-3}, which we used to construct the preference set. Table~\ref{tab:dpo_results} compares pre- and post-alignment F1 on \textsc{FOR2107}. After alignment, text-only performance in C1 increases substantially ($0.153 \rightarrow 0.459$, $+0.306$), while performance declines in C2 and C4; in the full multimodal setting, it falls from $0.723$ to $0.477$ ($-0.246$). In this evaluation, the aligned model produces no outputs that reference MRI findings or brain parcellation, and all three evaluated conditions remain below the strongest unaligned augmented score of 0.728. We consider two explanations for this combination of results:
\begin{itemize}[topsep=2pt, partopsep=0pt, parsep=0pt, itemsep=2pt, leftmargin=*]
    \item \textbf{Entanglement:} The strong multimodal performance before alignment may be linked to the preamble-associated behavior examined in our contrastive analysis. Under this explanation, MPO may suppress this behavior along with the targeted MRI-referencing language.
    \item \textbf{Surface suppression:} MPO may have learned the MRI-suppression objective independently; the F1 drop may result from losing a useful textual signal and does not establish suppression of a deeper inference mechanism.
\end{itemize}
Appendix~\ref{app:preference} discusses both explanations in more detail, although the present experiment cannot distinguish between them.

\section{Conclusion}
We present a clinical case of a modality-collapse-like pattern in two cohorts for which prior work does not establish the included neuroimaging inputs as reliable stand-alone diagnostic evidence for the present tasks. We exclude obvious classification shortcuts in consultation with domain experts and use preamble-only and swap-image conditions as controls. Under these conditions, smaller VLMs exhibit substantial F1 gains when MRI-related context is introduced. For the analyzed smaller models, most of the calibration improvement occurs when the textual preamble is added, before any image is supplied. We characterize this shift using a phrase-level probe and show with a false-modality ablation that the qualitative contrast recurs in additional model families. A preliminary expert case study finds that faithfulness remains low in every condition examined. In our single-model intervention, preference alignment suppresses MRI-referencing behavior while augmented-condition performance declines; whether this is related to deeper modality dependence remains unresolved. Our findings call for evaluation protocols that study the causal relationship between input content and model outputs before VLMs are deployed in clinical settings.

\section*{Limitations}
Our evaluation is limited to open-weight models with up to 72 billion parameters and two binary-classification cohorts. Generalization to closed systems, larger models, and multiclass or richer clinical tasks remains an open question. For reproducibility, all results use deterministic (greedy) decoding, and we do not examine whether the effect persists under sampling-based decoding.

The preamble-associated gain varies across domains: on \textsc{OASIS-3}, which was not used to develop the preamble, adding the preamble degrades performance for several models that are already competitive under text-only input, consistent with negative transfer. The direction and magnitude of the preamble-associated effect depend on the model and clinical domain, and the gain does not transfer consistently across settings.

Our main-text confidence analysis covers the two models with the largest gains; the appendix reports results for two additional models. Claims about how much of the calibration change occurs when the preamble is added should be read in that context. We emphasize, however, that the broader image-content finding is supported by swap-image results evaluated across all twelve models (Section~\ref{subsec:results}), while cross-family consistency is supported by the false-modality ablation (Section~\ref{sec:preamble}).

Due to the scarcity of qualified annotators for participant-level \textsc{FOR2107} traces and resource constraints, the preliminary expert case study uses a single clinical psychologist. We therefore treat the case study as qualitative and its absolute scores as descriptive; multi-annotator validation remains future work.

The residual-stream preamble direction in Appendix~\ref{app:preamble} is a correlational feature identified in a single model at a specific layer selected by a fixed onset criterion. We do not verify that the same geometry holds across model families. Establishing causal responsibility for the C1$\to$C2 shift would require intervention experiments, which are left for future work.

Our single-model MPO experiment shows suppression of MRI-referencing behavior after preference alignment, without establishing the factual accuracy of the resulting traces. Our dataset also cannot separate the two accounts of the post-alignment performance drop in Section~\ref{sec:dpo}, since MRI-mention status and correctness are correlated by construction. Finally, our probe centers on a single MRI-framing preamble, and the broader space of triggering phrases and the role of tabular serialization remain to be characterized.

\section*{Ethics Statement}
This study uses two clinical datasets governed by strict data-use agreements. \textsc{FOR2107} data access was granted under the consortium's institutional review protocol, which requires formal application and prohibits redistribution to third parties. \textsc{OASIS-3} is available through a controlled-access application process administered by the Washington University Knight Alzheimer Disease Research Center, subject to a data-use agreement that similarly prohibits redistribution. No individually identifiable participant information is presented in this work. Predictive metrics are reported in aggregate over the evaluated cohorts; the expert case study and residual-stream probe use the subsets described in their respective sections.

The study of reasoning traces was conducted by a qualified clinical psychologist who reviewed only model-generated outputs. No patient-facing clinical decisions were made on the basis of model predictions, and no participant data were disclosed outside the authorized research environments governed by the applicable data-use agreements.

We emphasize that the VLM outputs analyzed in this study are not suitable for clinical deployment. Several of the smaller evaluated models show augmented-condition gains that are not fully explained by added neuroimaging evidence under our controls. In the preliminary expert case study, the reviewed model produces justifications containing unverified clinical details. The tested MPO intervention suppresses MRI-referencing language, while performance declines in the augmented conditions. This gap between surface classification performance and evidence-grounded inference motivates the present evaluation. Before any patient-facing use of VLMs in clinical psychiatry or neurology is considered, future applications must undergo rigorous prospective validation by domain experts. This validation must include evaluation protocols that explicitly probe the causal relationship between input evidence and model outputs.

\section*{Acknowledgments}
This work is funded by the LOEWE Center DYNAMIC as part of the Hessian program for promoting cutting-edge research, LOEWE, under grant number LOEWE1/16/519/03/09.001(0009)/98. We are grateful to our UKPLab colleagues at TU Darmstadt for technical discussions at the AI4Psych, SIG Clinical/Psych, SIG Multimodal, SIG Foundation Models, and SIG LLM Understanding and Interpretability meetings. We appreciate the clinical discussions with colleagues at the universities of Marburg and Frankfurt. We especially thank Elena Maher and Dr. Nico Bast of the MIND CHILD--Emmy-Noether Research Group at Goethe University Frankfurt for their help in the clinical annotation of model-generated responses.

\begingroup
\righthyphenmin=2
\hyphenation{Rei-mann Mi-cha-e-lis Cars-ten Bark-hau}
\bibliography{custom, acl}

@inproceedings{ismithdeen-etal-2025-promptception,
    title = "Promptception: How Sensitive Are Large Multimodal Models to Prompts?",
    author = "Ismithdeen, Mohamed Insaf  and
      Khattak, Muhammad Uzair  and
      Khan, Salman",
    editor = "Christodoulopoulos, Christos  and
      Chakraborty, Tanmoy  and
      Rose, Carolyn  and
      Peng, Violet",
    booktitle = "Findings of the Association for Computational Linguistics: EMNLP 2025",
    month = nov,
    year = "2025",
    address = "Suzhou, China",
    publisher = "Association for Computational Linguistics",
    url = "https://aclanthology.org/2025.findings-emnlp.1302/",
    doi = "10.18653/v1/2025.findings-emnlp.1302",
    pages = "23950--23985",
    ISBN = "979-8-89176-335-7",
}

@inproceedings{ceballos-arroyo-etal-2024-open,
    title = "Open (Clinical) {LLM}s are Sensitive to Instruction Phrasings",
    author = "Ceballos-Arroyo, Alberto Mario  and
      Munnangi, Monica  and
      Sun, Jiuding  and
      Zhang, Karen  and
      McInerney, Jered  and
      Wallace, Byron C.  and
      Amir, Silvio",
    editor = "Demner-Fushman, Dina  and
      Ananiadou, Sophia  and
      Miwa, Makoto  and
      Roberts, Kirk  and
      Tsujii, Junichi",
    booktitle = "Proceedings of the 23rd Workshop on Biomedical Natural Language Processing",
    month = aug,
    year = "2024",
    address = "Bangkok, Thailand",
    publisher = "Association for Computational Linguistics",
    url = "https://aclanthology.org/2024.bionlp-1.5/",
    doi = "10.18653/v1/2024.bionlp-1.5",
    pages = "50--71",
}

@inproceedings{lu-etal-2024-prompts,
    title = "How are Prompts Different in Terms of Sensitivity?",
    author = "Lu, Sheng  and
      Schuff, Hendrik  and
      Gurevych, Iryna",
    editor = "Duh, Kevin  and
      Gomez, Helena  and
      Bethard, Steven",
    booktitle = "Proceedings of the 2024 Conference of the North American Chapter of the Association for Computational Linguistics: Human Language Technologies (Volume 1: Long Papers)",
    month = jun,
    year = "2024",
    address = "Mexico City, Mexico",
    publisher = "Association for Computational Linguistics",
    url = "https://aclanthology.org/2024.naacl-long.325/",
    doi = "10.18653/v1/2024.naacl-long.325",
    pages = "5833--5856",
}

@inproceedings{ishibashi-etal-2023-evaluating,
    title = "Evaluating the Robustness of Discrete Prompts",
    author = "Ishibashi, Yoichi  and
      Bollegala, Danushka  and
      Sudoh, Katsuhito  and
      Nakamura, Satoshi",
    editor = "Vlachos, Andreas  and
      Augenstein, Isabelle",
    booktitle = "Proceedings of the 17th Conference of the European Chapter of the Association for Computational Linguistics",
    month = may,
    year = "2023",
    address = "Dubrovnik, Croatia",
    publisher = "Association for Computational Linguistics",
    url = "https://aclanthology.org/2023.eacl-main.174/",
    doi = "10.18653/v1/2023.eacl-main.174",
    pages = "2373--2384",
}

@inproceedings{wolf-etal-2020-transformers,
    title = "Transformers: State-of-the-Art Natural Language Processing",
    author = "Wolf, Thomas  and
      Debut, Lysandre  and
      Sanh, Victor  and
      Chaumond, Julien  and
      Delangue, Clement  and
      Moi, Anthony  and
      Cistac, Pierric  and
      Rault, Tim  and
      Louf, Remi  and
      Funtowicz, Morgan  and
      Davison, Joe  and
      Shleifer, Sam  and
      von Platen, Patrick  and
      Ma, Clara  and
      Jernite, Yacine  and
      Plu, Julien  and
      Xu, Canwen  and
      Le Scao, Teven  and
      Gugger, Sylvain  and
      Drame, Mariama  and
      Lhoest, Quentin  and
      Rush, Alexander",
    editor = "Liu, Qun  and
      Schlangen, David",
    booktitle = "Proceedings of the 2020 Conference on Empirical Methods in Natural Language Processing: System Demonstrations",
    month = oct,
    year = "2020",
    address = "Online",
    publisher = "Association for Computational Linguistics",
    url = "https://aclanthology.org/2020.emnlp-demos.6/",
    doi = "10.18653/v1/2020.emnlp-demos.6",
    pages = "38--45",
}

@inproceedings{koo-etal-2024-benchmarking,
    title = "Benchmarking Cognitive Biases in Large Language Models as Evaluators",
    author = "Koo, Ryan  and
      Lee, Minhwa  and
      Raheja, Vipul  and
      Park, Jong Inn  and
      Kim, Zae Myung  and
      Kang, Dongyeop",
    editor = "Ku, Lun-Wei  and
      Martins, Andre  and
      Srikumar, Vivek",
    booktitle = "Findings of the Association for Computational Linguistics: ACL 2024",
    month = aug,
    year = "2024",
    address = "Bangkok, Thailand",
    publisher = "Association for Computational Linguistics",
    url = "https://aclanthology.org/2024.findings-acl.29/",
    doi = "10.18653/v1/2024.findings-acl.29",
    pages = "517--545",
}

@inproceedings{yoshida-etal-2025-visual,
    title = "Visual Priming Effect on Large-scale Vision Language Models",
    author = "Yoshida, Daiki  and
      Sakajo, Haruki  and
      Hayashi, Kazuki  and
      Sakai, Yusuke  and
      Kamigaito, Hidetaka  and
      Hayashi, Katsuhiko  and
      Watanabe, Taro",
    editor = "Angelova, Galia  and
      Kunilovskaya, Maria  and
      Escribe, Marie  and
      Mitkov, Ruslan",
    booktitle = "Proceedings of the 15th International Conference on Recent Advances in Natural Language Processing - Natural Language Processing in the Generative AI Era",
    month = sep,
    year = "2025",
    address = "Varna, Bulgaria",
    publisher = "INCOMA Ltd., Shoumen, Bulgaria",
    url = "https://aclanthology.org/2025.ranlp-1.160/",
    pages = "1385--1395",
}

@inproceedings{howard-etal-2025-uncovering,
    title = "Uncovering Bias in Large Vision-Language Models at Scale with Counterfactuals",
    author = "Howard, Phillip  and
      Fraser, Kathleen C.  and
      Bhiwandiwalla, Anahita  and
      Kiritchenko, Svetlana",
    editor = "Chiruzzo, Luis  and
      Ritter, Alan  and
      Wang, Lu",
    booktitle = "Proceedings of the 2025 Conference of the Nations of the Americas Chapter of the Association for Computational Linguistics: Human Language Technologies (Volume 1: Long Papers)",
    month = apr,
    year = "2025",
    address = "Albuquerque, New Mexico",
    publisher = "Association for Computational Linguistics",
    url = "https://aclanthology.org/2025.naacl-long.305/",
    doi = "10.18653/v1/2025.naacl-long.305",
    pages = "5946--5991",
    ISBN = "979-8-89176-189-6",
}

@inproceedings{zhong-etal-2024-investigating,
    title = "Investigating and Mitigating the Multimodal Hallucination Snowballing in Large Vision-Language Models",
    author = "Zhong, Weihong  and
      Feng, Xiaocheng  and
      Zhao, Liang  and
      Li, Qiming  and
      Huang, Lei  and
      Gu, Yuxuan  and
      Ma, Weitao  and
      Xu, Yuan  and
      Qin, Bing",
    editor = "Ku, Lun-Wei  and
      Martins, Andre  and
      Srikumar, Vivek",
    booktitle = "Proceedings of the 62nd Annual Meeting of the Association for Computational Linguistics (Volume 1: Long Papers)",
    month = aug,
    year = "2024",
    address = "Bangkok, Thailand",
    publisher = "Association for Computational Linguistics",
    url = "https://aclanthology.org/2024.acl-long.648/",
    doi = "10.18653/v1/2024.acl-long.648",
    pages = "11991--12011",
}

@inproceedings{parcalabescu-frank-2023-mm,
    title = "{MM}-{SHAP}: A Performance-agnostic Metric for Measuring Multimodal Contributions in Vision and Language Models {\&} Tasks",
    author = "Parcalabescu, Letitia  and
      Frank, Anette",
    booktitle = "Proceedings of the 61st Annual Meeting of the Association for Computational Linguistics (Volume 1: Long Papers)",
    month = jul,
    year = "2023",
    address = "Toronto, Canada",
    publisher = "Association for Computational Linguistics",
    url = "https://aclanthology.org/2023.acl-long.223",
    doi = "10.18653/v1/2023.acl-long.223",
    pages = "4032--4059",
}

@inproceedings{zhu-etal-2022-diagnosing,
    title = "Diagnosing Vision-and-Language Navigation: What Really Matters",
    author = "Zhu, Wanrong  and
      Qi, Yuankai  and
      Narayana, Pradyumna  and
      Sone, Kazoo  and
      Basu, Sugato  and
      Wang, Xin  and
      Wu, Qi  and
      Eckstein, Miguel  and
      Wang, William Yang",
    editor = "Carpuat, Marine  and
      de Marneffe, Marie-Catherine  and
      Meza Ruiz, Ivan Vladimir",
    booktitle = "Proceedings of the 2022 Conference of the North American Chapter of the Association for Computational Linguistics: Human Language Technologies",
    month = jul,
    year = "2022",
    address = "Seattle, United States",
    publisher = "Association for Computational Linguistics",
    url = "https://aclanthology.org/2022.naacl-main.438/",
    doi = "10.18653/v1/2022.naacl-main.438",
    pages = "5981--5993",
}

@inproceedings{sim-etal-2025-vlms,
    title = "Can {VLM}s Actually See and Read? A Survey on Modality Collapse in Vision-Language Models",
    author = "Sim, Mong Yuan  and
      Zhang, Wei Emma  and
      Dai, Xiang  and
      Fang, Biaoyan",
    editor = "Che, Wanxiang  and
      Nabende, Joyce  and
      Shutova, Ekaterina  and
      Pilehvar, Mohammad Taher",
    booktitle = "Findings of the Association for Computational Linguistics: ACL 2025",
    month = jul,
    year = "2025",
    address = "Vienna, Austria",
    publisher = "Association for Computational Linguistics",
    url = "https://aclanthology.org/2025.findings-acl.1256/",
    doi = "10.18653/v1/2025.findings-acl.1256",
    pages = "24452--24470",
    ISBN = "979-8-89176-256-5",
}

@inproceedings{geng-etal-2024-survey,
    title = "A Survey of Confidence Estimation and Calibration in Large Language Models",
    author = "Geng, Jiahui  and
      Cai, Fengyu  and
      Wang, Yuxia  and
      Koeppl, Heinz  and
      Nakov, Preslav  and
      Gurevych, Iryna",
    editor = "Duh, Kevin  and
      Gomez, Helena  and
      Bethard, Steven",
    booktitle = "Proceedings of the 2024 Conference of the North American Chapter of the Association for Computational Linguistics: Human Language Technologies (Volume 1: Long Papers)",
    month = jun,
    year = "2024",
    address = "Mexico City, Mexico",
    publisher = "Association for Computational Linguistics",
    url = "https://aclanthology.org/2024.naacl-long.366/",
    doi = "10.18653/v1/2024.naacl-long.366",
    pages = "6577--6595",
}

@inproceedings{zhang-etal-2026-images,
    title = "Do Images Speak Louder than Words? Investigating the Effect of Textual Misinformation in {VLM}s",
    author = "Zhang, Chi  and
      Ding, Wenxuan  and
      Liu, Jiale  and
      Wu, Mingrui  and
      Wu, Qingyun  and
      Mooney, Ray",
    editor = "Demberg, Vera  and
      Inui, Kentaro  and
      Marquez, Llu{\'i}s",
    booktitle = "Proceedings of the 19th Conference of the {E}uropean Chapter of the {A}ssociation for {C}omputational {L}inguistics (Volume 1: Long Papers)",
    month = mar,
    year = "2026",
    address = "Rabat, Morocco",
    publisher = "Association for Computational Linguistics",
    url = "https://aclanthology.org/2026.eacl-long.323/",
    doi = "10.18653/v1/2026.eacl-long.323",
    pages = "6872--6895",
    ISBN = "979-8-89176-380-7",
}

@inproceedings{balloccu-etal-2024-leak,
    title = "Leak, Cheat, Repeat: Data Contamination and Evaluation Malpractices in Closed-Source {LLM}s",
    author = "Balloccu, Simone  and
      Schmidtov{\'a}, Patr{\'i}cia  and
      Lango, Mateusz  and
      Dusek, Ondrej",
    editor = "Graham, Yvette  and
      Purver, Matthew",
    booktitle = "Proceedings of the 18th Conference of the European Chapter of the Association for Computational Linguistics (Volume 1: Long Papers)",
    month = mar,
    year = "2024",
    address = "St. Julian{'}s, Malta",
    publisher = "Association for Computational Linguistics",
    url = "https://aclanthology.org/2024.eacl-long.5/",
    doi = "10.18653/v1/2024.eacl-long.5",
    pages = "67--93",
}

@article{bai2025qwen3,
  title={Qwen3-vl technical report},
  author={Bai, Shuai and Cai, Yuxuan and Chen, Ruizhe and Chen, Keqin and Chen, Xionghui and Cheng, Zesen and Deng, Lianghao and Ding, Wei and Gao, Chang and Ge, Chunjiang and others},
  journal={arXiv preprint arXiv:2511.21631},
  year={2025}
}

@article{wang2025internvl3,
  title={Internvl3. 5: Advancing open-source multimodal models in versatility, reasoning, and efficiency},
  author={Wang, Weiyun and Gao, Zhangwei and Gu, Lixin and Pu, Hengjun and Cui, Long and Wei, Xingguang and Liu, Zhaoyang and Jing, Linglin and Ye, Shenglong and Shao, Jie and others},
  journal={arXiv preprint arXiv:2508.18265},
  year={2025}
}

@article{hong2025glm,
  title={Glm-4.5 v and glm-4.1 v-thinking: Towards versatile multimodal reasoning with scalable reinforcement learning},
  author={Hong, Wenyi and Yu, Wenmeng and Gu, Xiaotao and Wang, Guo and Gan, Guobing and Tang, Haomiao and Cheng, Jiale and Qi, Ji and Ji, Junhui and Pan, Lihang and others},
  journal={arXiv preprint arXiv:2507.01006},
  year={2025}
}

@article{zeng2025glm,
  title={Glm-4.5: Agentic, reasoning, and coding (arc) foundation models},
  author={Zeng, Aohan and Lv, Xin and Zheng, Qinkai and Hou, Zhenyu and Chen, Bin and Xie, Chengxing and Wang, Cunxiang and Yin, Da and Zeng, Hao and Zhang, Jiajie and others},
  journal={arXiv preprint arXiv:2508.06471},
  year={2025}
}

@article{li2025llavaonevision,
  title={{LL}a{VA}-OneVision: Easy Visual Task Transfer},
  author={Bo Li and Yuanhan Zhang and Dong Guo and Renrui Zhang and Feng Li and Hao Zhang and Kaichen Zhang and Peiyuan Zhang and Yanwei Li and Ziwei Liu and Chunyuan Li},
  journal={Transactions on Machine Learning Research},
  issn={2835-8856},
  year={2025},
  url={https://openreview.net/forum?id=zKv8qULV6n},
}

@article{liu2026ministral,
  title={Ministral 3},
  author={Liu, Alexander H and Khandelwal, Kartik and Subramanian, Sandeep and Jouault, Victor and Rastogi, Abhinav and Sad{\'e}, Adrien and Jeffares, Alan and Jiang, Albert and Cahill, Alexandre and Gavaudan, Alexandre and others},
  journal={arXiv preprint arXiv:2601.08584},
  year={2026}
}

@misc{xu2025qwen25vltechnicalreport,
      title={Qwen2.5-VL Technical Report}, 
      author={Yiheng Xu and Peng Wang and Hang Zhang and Pengfei Wang and Shuai Bai and Shijie Wang and Junyang Lin and Tianbao Xie and Yuanzhi Zhu and Zhibo Yang and Wei Ding and Xi Zhang and Jianqiang Wan and Jun Tang and Haiyang Xu and Jiabo Ye and Keqin Chen and Xuejing Liu and Jialin Wang and Wenbin Ge and Kai Dang and Zesen Cheng and Mingkun Yang and Sibo Song and Zhaohai Li and Humen Zhong and Zheren Fu},
      year={2025},
      eprint={2502.13923},
      archivePrefix={arXiv},
      primaryClass={cs.CV},
      url={https://arxiv.org/abs/2502.13923}, 
}

@ARTICLE{kircher2019-db,
  title    = "Neurobiology of the major psychoses: a translational perspective on brain structure and function---the {FOR2107} consortium",
  author   = "Kircher, Tilo and W{\"o}hr, Markus and Nenadic, Igor and Schwarting, Rainer and Schratt, Gerhard and Alferink, Judith and Culmsee, Carsten and Garn, Holger and Hahn, Tim and M{\"u}ller-Myhsok, Bertram and Dempfle, Astrid and Hahmann, Maik and Jansen, Andreas and Pfefferle, Petra and Renz, Harald and Rietschel, Marcella and Witt, Stephanie H and N{\"o}then, Markus and Krug, Axel and Dannlowski, Udo",
  journal  = "European Archives of Psychiatry and Clinical Neuroscience",
  volume   =  269,
  number   =  8,
  pages    = "949--962",
  month    =  dec,
  year     =  2019
}

@article{LaMontagne2019,
	author = {LaMontagne, Pamela J. and Benzinger, Tammie LS. and Morris, John C. and Keefe, Sarah and Hornbeck, Russ and Xiong, Chengjie and Grant, Elizabeth and Hassenstab, Jason and Moulder, Krista and Vlassenko, Andrei G. and Raichle, Marcus E. and Cruchaga, Carlos and Marcus, Daniel},
	title = {OASIS-3: Longitudinal Neuroimaging, Clinical, and Cognitive Dataset for Normal Aging and Alzheimer Disease},
	elocation-id = {2019.12.13.19014902},
	year = {2019},
	doi = {10.1101/2019.12.13.19014902},
	publisher = {Cold Spring Harbor Laboratory Press},
	URL = {https://www.medrxiv.org/content/early/2019/12/15/2019.12.13.19014902},
	journal = {medRxiv}
}

@inproceedings{hegselmann2023tabllm,
  title={Tabllm: Few-shot classification of tabular data with large language models},
  author={Hegselmann, Stefan and Buendia, Alejandro and Lang, Hunter and Agrawal, Monica and Jiang, Xiaoyi and Sontag, David},
  booktitle={International Conference on Artificial Intelligence and Statistics},
  pages={5549--5581},
  year={2023},
  organization={PMLR}
}

@inproceedings{jones2022capturing,
  title={Capturing Failures of Large Language Models via Human Cognitive Biases},
  author={Erik Jones and Jacob Steinhardt},
  booktitle={Advances in Neural Information Processing Systems},
  editor={Alice H. Oh and Alekh Agarwal and Danielle Belgrave and Kyunghyun Cho},
  year={2022},
  url={https://openreview.net/forum?id=fcO9Cgn-X-R}
}

@misc{nilearn,
  author       = {Nilearn contributors, Nilearn},
  title        = {nilearn},
  month        = dec,
  year         = 2014,
  publisher    = {Zenodo},
  version      = {0.13.1},
  doi          = {10.5281/zenodo.18600337},
  url          = {https://doi.org/10.5281/zenodo.18600337},
}

@ARTICLE{havard-oxford,
  AUTHOR={Rushmore, R. Jarrett  and Sunderland, Kyle  and Carrington, Holly  and Chen, Justine  and Halle, Michael  and Lasso, Andras  and Papadimitriou, G.  and Prunier, N.  and Rizzoni, Elizabeth  and Vessey, Brynn  and Wilson-Braun, Peter  and Rathi, Yogesh  and Kubicki, Marek  and Bouix, Sylvain  and Yeterian, Edward  and Makris, Nikos },
  TITLE={Anatomically curated segmentation of human subcortical structures in high resolution magnetic resonance imaging: An open science approach},
  JOURNAL={Frontiers in Neuroanatomy},
  VOLUME={Volume 16 - 2022},
  YEAR={2022},
  URL={https://www.frontiersin.org/journals/neuroanatomy/articles/10.3389/fnana.2022.894606},
  DOI={10.3389/fnana.2022.894606},
  ISSN={1662-5129},
}

@article{Gulati__Sebe_Lepri_Oliver_2025, 
  title={Beauty and the Bias: Exploring the Impact of Attractiveness on Multimodal Large Language Models}, 
  volume={8}, 
  url={https://ojs.aaai.org/index.php/AIES/article/view/36619}, 
  DOI={10.1609/aies.v8i2.36619}, 
  number={2}, 
  journal={Proceedings of the AAAI/ACM Conference on AI, Ethics, and Society}, 
  author={Gulati, Aditya and D’Incà, Moreno and Sebe, Nicu and Lepri, Bruno and Oliver, Nuria}, 
  year={2025}, 
  month={Oct.}, 
  pages={1154-1168} 
}

@inproceedings{shu2025when,
    title={When Semantics Mislead Vision: Mitigating Large Multimodal Models Hallucinations in Scene Text Spotting and Understanding},
    author={Yan Shu and Hangui Lin and Yexin Liu and Yan Zhang and Gangyan Zeng and Yan Li and Yu ZHOU and Ser-Nam Lim and Harry Yang and Nicu Sebe},
    booktitle={The Thirty-ninth Annual Conference on Neural Information Processing Systems},
    year={2025},
    url={https://openreview.net/forum?id=BbWrp6O8Lm}
}

@article{Meyer1971FacilitationIR,
  title={Facilitation in recognizing pairs of words: evidence of a dependence between retrieval operations.},
  author={David E. Meyer and Roger W. Schvaneveldt},
  journal={Journal of experimental psychology},
  year={1971},
  volume={90 2},
  pages={
          227-34
        },
  url={https://api.semanticscholar.org/CorpusID:36672941}
}

@inproceedings{hu2022lora,
    title={Lo{RA}: Low-Rank Adaptation of Large Language Models},
    author={Edward J Hu and yelong shen and Phillip Wallis and Zeyuan Allen-Zhu and Yuanzhi Li and Shean Wang and Lu Wang and Weizhu Chen},
    booktitle={International Conference on Learning Representations},
    year={2022},
    url={https://openreview.net/forum?id=nZeVKeeFYf9}
}

@inproceedings{rafailov2023direct,
    title={Direct Preference Optimization: Your Language Model is Secretly a Reward Model},
    author={Rafael Rafailov and Archit Sharma and Eric Mitchell and Christopher D Manning and Stefano Ermon and Chelsea Finn},
    booktitle={Thirty-seventh Conference on Neural Information Processing Systems},
    year={2023},
    url={https://openreview.net/forum?id=HPuSIXJaa9}
}

@article{wang2024enhancing,
  title={Enhancing the reasoning ability of multimodal large language models via mixed preference optimization},
  author={Wang, Weiyun and Chen, Zhe and Wang, Wenhai and Cao, Yue and Liu, Yangzhou and Gao, Zhangwei and Zhu, Jinguo and Zhu, Xizhou and Lu, Lewei and Qiao, Yu and others},
  journal={arXiv preprint arXiv:2411.10442},
  year={2024}
}

@Article{moor2023foundation,
    author={Moor, Michael
    and Banerjee, Oishi
    and Abad, Zahra Shakeri Hossein
    and Krumholz, Harlan M.
    and Leskovec, Jure
    and Topol, Eric J.
    and Rajpurkar, Pranav},
    title={Foundation models for generalist medical artificial intelligence},
    journal={Nature},
    year={2023},
    month={Apr},
    day={01},
    volume={616},
    number={7956},
    pages={259-265},
    issn={1476-4687},
    doi={10.1038/s41586-023-05881-4},
    url={https://doi.org/10.1038/s41586-023-05881-4}
}

@Article{Singhal2023,
    author={Singhal, Karan
    and Azizi, Shekoofeh
    and Tu, Tao
    and Mahdavi, S. Sara
    and Wei, Jason
    and Chung, Hyung Won
    and Scales, Nathan
    and Tanwani, Ajay
    and Cole-Lewis, Heather
    and Pfohl, Stephen
    and Payne, Perry
    and Seneviratne, Martin
    and Gamble, Paul
    and Kelly, Chris
    and Babiker, Abubakr
    and Sch{\"a}rli, Nathanael
    and Chowdhery, Aakanksha
    and Mansfield, Philip
    and Demner-Fushman, Dina
    and Ag{\"u}era y Arcas, Blaise
    and Webster, Dale
    and Corrado, Greg S.
    and Matias, Yossi
    and Chou, Katherine
    and Gottweis, Juraj
    and Tomasev, Nenad
    and Liu, Yun
    and Rajkomar, Alvin
    and Barral, Joelle
    and Semturs, Christopher
    and Karthikesalingam, Alan
    and Natarajan, Vivek},
    title={Large language models encode clinical knowledge},
    journal={Nature},
    year={2023},
    month={Aug},
    day={01},
    volume={620},
    number={7972},
    pages={172-180},
    issn={1476-4687},
    doi={10.1038/s41586-023-06291-2},
    url={https://doi.org/10.1038/s41586-023-06291-2}
}

@inproceedings{li2023llavamed,
    title={{LL}a{VA}-Med: Training a Large Language-and-Vision Assistant for Biomedicine in One Day},
    author={Chunyuan Li and Cliff Wong and Sheng Zhang and Naoto Usuyama and Haotian Liu and Jianwei Yang and Tristan Naumann and Hoifung Poon and Jianfeng Gao},
    booktitle={Thirty-seventh Conference on Neural Information Processing Systems Datasets and Benchmarks Track},
    year={2023},
    url={https://openreview.net/forum?id=GSuP99u2kR}
}

@ARTICLE{winter2022,
  title    = "Quantifying Deviations of Brain Structure and Function in Major
              Depressive Disorder Across Neuroimaging Modalities",
  author   = "Winter, Nils R and Leenings, Ramona and Ernsting, Jan and Sarink,
              Kelvin and Fisch, Lukas and Emden, Daniel and Blanke, Julian and
              Goltermann, Janik and Opel, Nils and Barkhau, Carlotta and
              Meinert, Susanne and Dohm, Katharina and Repple, Jonathan and
              Mauritz, Marco and Gruber, Marius and Leehr, Elisabeth J and
              Grotegerd, Dominik and Redlich, Ronny and Jansen, Andreas and
              Nenadic, Igor and N{\"o}then, Markus M and Forstner, Andreas and
              Rietschel, Marcella and Gro{\ss}, Joachim and Bauer, Jochen and
              Heindel, Walter and Andlauer, Till and Eickhoff, Simon B and
              Kircher, Tilo and Dannlowski, Udo and Hahn, Tim",
  journal  = "JAMA Psychiatry",
  volume   =  79,
  number   =  9,
  pages    = "879--888",
  month    =  sep,
  year     =  2022,
  address  = "United States",
  language = "en"
}

@ARTICLE{winter2024,
  title    = "A Systematic Evaluation of Machine {Learning-Based} Biomarkers
              for Major Depressive Disorder",
  author   = "Winter, Nils R and Blanke, Julian and Leenings, Ramona and
              Ernsting, Jan and Fisch, Lukas and Sarink, Kelvin and Barkhau,
              Carlotta and Emden, Daniel and Thiel, Katharina and
              Flinkenfl{\"u}gel, Kira and Winter, Alexandra and Goltermann,
              Janik and Meinert, Susanne and Dohm, Katharina and Repple,
              Jonathan and Gruber, Marius and Leehr, Elisabeth J and Opel, Nils
              and Grotegerd, Dominik and Redlich, Ronny and Nitsch, Robert and
              Bauer, Jochen and Heindel, Walter and Gross, Joachim and Risse,
              Benjamin and Andlauer, Till F M and Forstner, Andreas J and
              N{\"o}then, Markus M and Rietschel, Marcella and Hofmann, Stefan
              G and Pfarr, Julia-Katharina and Teutenberg, Lea and Usemann,
              Paula and Thomas-Odenthal, Florian and Wroblewski, Adrian and
              Brosch, Katharina and Stein, Frederike and Jansen, Andreas and
              Jamalabadi, Hamidreza and Alexander, Nina and Straube, Benjamin
              and Nenadic, Igor and Kircher, Tilo and Dannlowski, Udo and Hahn,
              Tim",
  journal  = "JAMA Psychiatry",
  volume   =  81,
  number   =  4,
  pages    = "386--395",
  month    =  apr,
  year     =  2024,
  address  = "United States",
  language = "en"
}

@InProceedings{pmlr-v162-javaloy22a,
  title = 	 {Mitigating Modality Collapse in Multimodal {VAE}s via Impartial Optimization},
  author =       {Javaloy, Adrian and Meghdadi, Maryam and Valera, Isabel},
  booktitle = 	 {Proceedings of the 39th International Conference on Machine Learning},
  pages = 	 {9938--9964},
  year = 	 {2022},
  editor = 	 {Chaudhuri, Kamalika and Jegelka, Stefanie and Song, Le and Szepesvari, Csaba and Niu, Gang and Sabato, Sivan},
  volume = 	 {162},
  series = 	 {Proceedings of Machine Learning Research},
  month = 	 {17--23 Jul},
  publisher =    {PMLR},
}

@article{vu2025roleplaying,
  title={Roleplaying with Structure: Synthetic Therapist-Client Conversation Generation from Questionnaires},
  author={Vu, Doan Nam Long and Tan, Rui and Moench, Lena and Francke, Svenja Jule and Woiwod, Daniel and Thomas-Odenthal, Florian and Stroth, Sanna and Kircher, Tilo and Hermann, Christiane and Dannlowski, Udo and others},
  journal={arXiv preprint arXiv:2510.25384},
  year={2025}
}

@inproceedings{kapoor2025computer,
    title={Computer Graphics from a Neuroscientist's perspective},
    author={Shreya Kapoor and Bernhard Egger},
    booktitle={Second Workshop on Representational Alignment at ICLR 2025},
    year={2025},
    url={https://openreview.net/forum?id=IOzjgYvxw1}
}

@article{Galteau2025,
    author = {Galteau, Marie E. and Broadwater, Margaret and Chen, Yi and Desrosiers-Gregoire, Gabriel and Gil, Rita and Kaesser, Johannes and Kim, Eugene and Kıryağdı, Pervin and Lambers, Henriette and Liu, Yanyan Y. and López-Gil, Xavier and MacNicol, Eilidh and Mohebkhodaei, Parastoo and De Oliveira, Ricardo X.N. and Pereira, Carolina A. and Reimann, Henning M. and Rivera-Olvera, Alejandro and Selingue, Erwan and Sirmpilatze, Nikoloz and Strobelt, Sandra and Sumiyoshi, Akira and Tham, Channelle and Tudela, Raul and Vrooman, Roël M. and Wank, Isabel and Zhang, Yongzhi and van Engelenburg, Wessel A. and Baudewig, Jürgen and Boretius, Susann and Cash, Diana and Chakravarty, M. Mallar and Chuang, Kai-Hsiang and Ciobanu, Luisa and Devenyi, Gabriel A. and Faber, Cornelius and Hess, Andreas and Homberg, Judith R. and Jelescu, Ileana O. and Justicia, Carles and Kawashima, Ryuta and Niendorf, Thoralf and Scheenen, Tom W.J. and Shemesh, Noam and Soria, Guadalupe and Todd, Nick and Wachsmuth, Lydia and Yu, Xin and Zhang, Baogui B. and Shih, Yen-Yu Ian and Lee, Sung-Ho and Grandjean, Joanes},
    title = {Activation mapping in multi-center retrospective rat sensory-evoked functional MRI datasets using a unified pipeline},
    journal = {Imaging Neuroscience},
    volume = {3},
    pages = {IMAG.a.157},
    year = {2025},
    month = {09},
    issn = {2837-6056},
    doi = {10.1162/IMAG.a.157},
    url = {https://doi.org/10.1162/IMAG.a.157},
    eprint = {https://direct.mit.edu/imag/article-pdf/doi/10.1162/IMAG.a.157/2550619/imag.a.157.pdf},
}

@InProceedings{pmlr-v139-zhao21c,
  title = 	 {Calibrate Before Use: Improving Few-shot Performance of Language Models},
  author =       {Zhao, Zihao and Wallace, Eric and Feng, Shi and Klein, Dan and Singh, Sameer},
  booktitle = 	 {Proceedings of the 38th International Conference on Machine Learning},
  pages = 	 {12697--12706},
  year = 	 {2021},
  editor = 	 {Meila, Marina and Zhang, Tong},
  volume = 	 {139},
  series = 	 {Proceedings of Machine Learning Research},
  month = 	 {18--24 Jul},
  publisher =    {PMLR},
  url = 	 {https://proceedings.mlr.press/v139/zhao21c.html},
}

@InProceedings{pmlr-v70-guo17a,
  title = 	 {On Calibration of Modern Neural Networks},
  author =       {Chuan Guo and Geoff Pleiss and Yu Sun and Kilian Q. Weinberger},
  booktitle = 	 {Proceedings of the 34th International Conference on Machine Learning},
  pages = 	 {1321--1330},
  year = 	 {2017},
  editor = 	 {Precup, Doina and Teh, Yee Whye},
  volume = 	 {70},
  series = 	 {Proceedings of Machine Learning Research},
  month = 	 {06--11 Aug},
  publisher =    {PMLR},
  url = 	 {https://proceedings.mlr.press/v70/guo17a.html},
}

@article {Brier1950,
      author = "Glenn W.  Brier",
      title = "Verification of Forecasts Expressed in Terms of Probability",
      journal = "Monthly Weather Review",
      year = "1950",
      publisher = "American Meteorological Society",
      address = "Boston MA, USA",
      volume = "78",
      number = "1",
      doi = "10.1175/1520-0493(1950)078<0001:VOFEIT>2.0.CO;2",
      pages=      "1 - 3",
      url = "https://journals.ametsoc.org/view/journals/mwre/78/1/1520-0493_1950_078_0001_vofeit_2_0_co_2.xml"
}

@article{Pakdaman_Naeini_Cooper_Hauskrecht_2015,
    title={Obtaining Well Calibrated Probabilities Using Bayesian Binning},
    volume={29},
    url={https://ojs.aaai.org/index.php/AAAI/article/view/9602},
    DOI={10.1609/aaai.v29i1.9602},
   number={1},
   journal={Proceedings of the AAAI Conference on Artificial Intelligence},
   author={Pakdaman Naeini, Mahdi and Cooper, Gregory and Hauskrecht, Milos},
   year={2015},
   month={Feb.} 
}

@misc{sellergren2026medgemmatechnicalreport,
      title={MedGemma Technical Report}, 
      author={Andrew Sellergren and Sahar Kazemzadeh and Tiam Jaroensri and Atilla Kiraly and Madeleine Traverse and Timo Kohlberger and Shawn Xu and Fayaz Jamil and Cían Hughes and Charles Lau and Justin Chen and Fereshteh Mahvar and Liron Yatziv and Tiffany Chen and Bram Sterling and Stefanie Anna Baby and Susanna Maria Baby and Jeremy Lai and Samuel Schmidgall and Lu Yang and Kejia Chen and Per Bjornsson and Shashir Reddy and Ryan Brush and Kenneth Philbrick and Mercy Asiedu and Ines Mezerreg and Howard Hu and Howard Yang and Richa Tiwari and Sunny Jansen and Preeti Singh and Yun Liu and Shekoofeh Azizi and Aishwarya Kamath and Johan Ferret and Shreya Pathak and Nino Vieillard and Ramona Merhej and Sarah Perrin and Tatiana Matejovicova and Alexandre Ramé and Morgane Riviere and Louis Rouillard and Thomas Mesnard and Geoffrey Cideron and Jean-bastien Grill and Sabela Ramos and Edouard Yvinec and Michelle Casbon and Elena Buchatskaya and Jean-Baptiste Alayrac and Dmitry Lepikhin and Vlad Feinberg and Sebastian Borgeaud and Alek Andreev and Cassidy Hardin and Robert Dadashi and Léonard Hussenot and Armand Joulin and Olivier Bachem and Yossi Matias and Katherine Chou and Avinatan Hassidim and Kavi Goel and Clement Farabet and Joelle Barral and Tris Warkentin and Jonathon Shlens and David Fleet and Victor Cotruta and Omar Sanseviero and Gus Martins and Phoebe Kirk and Anand Rao and Shravya Shetty and David F. Steiner and Can Kirmizibayrak and Rory Pilgrim and Daniel Golden and Lin Yang},
      year={2026},
      eprint={2507.05201},
      archivePrefix={arXiv},
      primaryClass={cs.AI},
      url={https://arxiv.org/abs/2507.05201}, 
}

@article{10.1162/IMAG.a.157,
    author = {Galteau, Marie E. and Broadwater, Margaret and Chen, Yi and Desrosiers-Gregoire, Gabriel and Gil, Rita and Kaesser, Johannes and Kim, Eugene and Kıryağdı, Pervin and Lambers, Henriette and Liu, Yanyan Y. and López-Gil, Xavier and MacNicol, Eilidh and Mohebkhodaei, Parastoo and De Oliveira, Ricardo X.N. and Pereira, Carolina A. and Reimann, Henning M. and Rivera-Olvera, Alejandro and Selingue, Erwan and Sirmpilatze, Nikoloz and Strobelt, Sandra and Sumiyoshi, Akira and Tham, Channelle and Tudela, Raul and Vrooman, Roël M. and Wank, Isabel and Zhang, Yongzhi and van Engelenburg, Wessel A. and Baudewig, Jürgen and Boretius, Susann and Cash, Diana and Chakravarty, M. Mallar and Chuang, Kai-Hsiang and Ciobanu, Luisa and Devenyi, Gabriel A. and Faber, Cornelius and Hess, Andreas and Homberg, Judith R. and Jelescu, Ileana O. and Justicia, Carles and Kawashima, Ryuta and Niendorf, Thoralf and Scheenen, Tom W.J. and Shemesh, Noam and Soria, Guadalupe and Todd, Nick and Wachsmuth, Lydia and Yu, Xin and Zhang, Baogui B. and Shih, Yen-Yu Ian and Lee, Sung-Ho and Grandjean, Joanes},
    title = {Activation mapping in multi-center retrospective rat sensory-evoked functional MRI datasets using a unified pipeline},
    journal = {Imaging Neuroscience},
    volume = {3},
    pages = {IMAG.a.157},
    year = {2025},
    month = {09},
    issn = {2837-6056},
    doi = {10.1162/IMAG.a.157},
    url = {https://doi.org/10.1162/IMAG.a.157},
    eprint = {https://direct.mit.edu/imag/article-pdf/doi/10.1162/IMAG.a.157/2550619/imag.a.157.pdf},
}

@Article{geirhos-etal-2020-shortcut,
    author={Geirhos, Robert
    and Jacobsen, J{\"o}rn-Henrik
    and Michaelis, Claudio
    and Zemel, Richard
    and Brendel, Wieland
    and Bethge, Matthias
    and Wichmann, Felix A.},
    title={Shortcut learning in deep neural networks},
    journal={Nature Machine Intelligence},
    year={2020},
    month={Nov},
    day={01},
    volume={2},
    number={11},
    pages={665-673},
    issn={2522-5839},
    doi={10.1038/s42256-020-00257-z},
    url={https://doi.org/10.1038/s42256-020-00257-z}
}

@Article{degrave-etal-2021-shortcut,
    author={DeGrave, Alex J.
    and Janizek, Joseph D.
    and Lee, Su-In},
    title={AI for radiographic COVID-19 detection selects shortcuts over signal},
    journal={Nature Machine Intelligence},
    year={2021},
    month={Jul},
    day={01},
    volume={3},
    number={7},
    pages={610-619},
    issn={2522-5839},
    doi={10.1038/s42256-021-00338-7},
    url={https://doi.org/10.1038/s42256-021-00338-7}
}

@inproceedings{xiong2024can,
    title={Can {LLM}s Express Their Uncertainty? An Empirical Evaluation of Confidence Elicitation in {LLM}s},
    author={Miao Xiong and Zhiyuan Hu and Xinyang Lu and YIFEI LI and Jie Fu and Junxian He and Bryan Hooi},
    booktitle={The Twelfth International Conference on Learning Representations},
    year={2024},
    url={https://openreview.net/forum?id=gjeQKFxFpZ}
}

@ARTICLE{Lombardi2020-oj,
  title    = "Structural magnetic resonance imaging for the early diagnosis of
              dementia due to Alzheimer's disease in people with mild cognitive
              impairment",
  author   = "Lombardi, Gemma and Crescioli, Giada and Cavedo, Enrica and
              Lucenteforte, Ersilia and Casazza, Giovanni and Bellatorre,
              Alessandro-Giacco and Lista, Chiara and Costantino, Giorgio and
              Frisoni, Giovanni and Virgili, Gianni and Filippini, Graziella",
  journal  = "Cochrane Database Syst Rev",
  volume   =  3,
  number   =  3,
  pages    = "CD009628",
  month    =  mar,
  year     =  2020,
  address  = "England",
  language = "en"
}

@ARTICLE{Ansart2020-mk,
  title    = "Predicting the progression of mild cognitive impairment using
              machine learning: A systematic, quantitative and critical review",
  author   = "Ansart, Manon and Epelbaum, St{\'e}phane and Bassignana, Giulia
              and B{\^o}ne, Alexandre and Bottani, Simona and Cattai, Tiziana
              and Couronn{\'e}, Rapha{\"e}l and Faouzi, Johann and Koval, Igor
              and Louis, Maxime and Thibeau-Sutre, Elina and Wen, Junhao and
              Wild, Adam and Burgos, Ninon and Dormont, Didier and Colliot,
              Olivier and Durrleman, Stanley",
  journal  = "Med Image Anal",
  volume   =  67,
  pages    = "101848",
  month    =  oct,
  year     =  2020,
  address  = "Netherlands",
  language = "en"
}
\endgroup
\appendix
\section{Prompt}\label{app:prompt}

Figures~\ref{fig:for2107_textcsv} and~\ref{fig:for2107_textcsv_promptmri} show the \textsc{FOR2107} prompts; Figures~\ref{fig:oasis_textcsv} and~\ref{fig:oasis_textcsv_promptmri} show the \textsc{OASIS-3} prompts. Labels are extracted for every condition and model with a 100\% parse rate using the fixed JSON prefix \texttt{\{"category": "}. Table~\ref{tab:prefix_token_ids} lists the corresponding prefix token IDs.

\begin{figure}[htbp]
    \centering
    \fbox{%
      \colorbox{beigebackground}{%
        \begin{minipage}{0.93\linewidth}
          \sffamily\small\setlength{\parindent}{0pt}%
    You are given patient clinical information

    Classify the patient into one of the following categories:

    - Major Depressive Disorder

    - Control (no disorder detected)

    Return your answer as a JSON object with two fields:

    - "category": the chosen category (exactly one of the two above)

    - "explanation": a short reasoning for the choice
    \end{minipage}%
      }%
    }
    \caption{\textsc{text(csv)} prompt for \textsc{FOR2107}.}
    \label{fig:for2107_textcsv}
\end{figure}

\begin{figure}[htbp]
    \centering
    \fbox{%
      \colorbox{beigebackground}{%
        \begin{minipage}{0.93\linewidth}
          \sffamily\small\setlength{\parindent}{0pt}%
    You are given patient clinical information \textbf{and their MRI data (brain parcellation volume, visualization of brain regions)}

    Classify the patient into one of the following categories:

    - Major Depressive Disorder

    - Control (no disorder detected)

    Return your answer as a JSON object with two fields:

    - "category": the chosen category (exactly one of the two above)

    - "explanation": a short reasoning for the choice
    \end{minipage}%
      }%
    }
    \caption{\textsc{text(csv)} $+$ \textsc{prompt(mri)} prompt for \textsc{FOR2107}. \textbf{Bold text} denotes the MRI-availability phrase added by \textsc{prompt(mri)} in C2.}
    \label{fig:for2107_textcsv_promptmri}
\end{figure}

\begin{figure}[htbp]
    \centering
    \fbox{%
      \colorbox{beigebackground}{%
        \begin{minipage}{0.93\linewidth}
          \sffamily\small\setlength{\parindent}{0pt}%
    You are given patient clinical information

    Classify the patient into one of the following categories:

    - Cognitive Normal

    - Cognitive Decline

    Return your answer as a JSON object with two fields:

    - "category": the chosen category (exactly one of the two above)

    - "explanation": a short reasoning for the choice
    \end{minipage}%
      }%
    }
    \caption{\textsc{text(csv)} prompt for \textsc{OASIS-3}.}
    \label{fig:oasis_textcsv}
\end{figure}

\begin{figure}[htbp]
    \centering
    \fbox{%
      \colorbox{beigebackground}{%
        \begin{minipage}{0.93\linewidth}
          \sffamily\small\setlength{\parindent}{0pt}%
    You are given patient clinical information \textbf{and their MRI data (brain parcellation volume, visualization of brain regions)}

    Classify the patient into one of the following categories:

    - Cognitive Normal

    - Cognitive Decline

    Return your answer as a JSON object with two fields:

    - "category": the chosen category (exactly one of the two above)

    - "explanation": a short reasoning for the choice
    \end{minipage}%
      }%
    }
    \caption{\textsc{text(csv)} $+$ \textsc{prompt(mri)} prompt for \textsc{OASIS-3}. \textbf{Bold text} denotes the MRI-availability phrase added by \textsc{prompt(mri)} in C2.}
    \label{fig:oasis_textcsv_promptmri}
\end{figure}

\begin{table*}[tp]
  \centering
  \small
  \setlength{\tabcolsep}{4pt}
  \begin{tabular}{l*{10}{r}}
    \toprule
    \textbf{Model} & \multicolumn{10}{c}{\textbf{Prefix token IDs}} \\
    \midrule
    \texttt{GLM-4.1V-9B} & 73022 & 2236 & 198 & 515 & 220 & 330 & 5471 & 788 & 330 &  \\
    \texttt{GLM-4.6V-Flash} & 73022 & 2236 & 198 & 515 & 220 & 330 & 5471 & 788 & 330 &  \\
    \addlinespace[2pt]
    \texttt{LLaVA-OV-1.5-4B} & 73594 & 2236 & 198 & 515 & 220 & 330 & 5471 & 788 & 330 &  \\
    \addlinespace[2pt]
    \texttt{Ministral-3-3B} & 1975 & 8353 & 1010 & 2030 & 1032 & 1429 & 17278 & 2811 & 1429 &  \\
    \texttt{Ministral-3-14B} & 1975 & 8353 & 1010 & 2030 & 1032 & 1429 & 17278 & 2811 & 1429 &  \\
    \addlinespace[2pt]
    \texttt{Qwen2.5-VL-3B} & 73594 & 2236 & 198 & 515 & 220 & 330 & 5471 & 788 & 330 &  \\
    \texttt{Qwen2.5-VL-32B} & 73594 & 2236 & 198 & 515 & 220 & 330 & 5471 & 788 & 330 &  \\
    \texttt{Qwen2.5-VL-72B} & 73594 & 2236 & 198 & 515 & 220 & 330 & 5471 & 788 & 330 &  \\
    \texttt{Qwen3-VL-2B} & 73594 & 2236 & 198 & 515 & 220 & 330 & 5471 & 788 & 330 &  \\
    \texttt{Qwen3-VL-32B} & 73594 & 2236 & 198 & 515 & 220 & 330 & 5471 & 788 & 330 &  \\
    \texttt{InternVL3.5-4B} & 73594 & 2236 & 198 & 515 & 220 & 330 & 5471 & 788 & 330 &  \\
    \texttt{InternVL3.5-14B} & 73594 & 2236 & 198 & 515 & 220 & 330 & 5471 & 788 & 330 &  \\
    \addlinespace[2pt]
    \texttt{medgemma-4b-it} & 2717 & 3723 & 107 & 236782 & 107 & 138 & 236775 & 10618 & 1083 &  \\
    \texttt{medgemma-27b-it} & 2717 & 3723 & 107 & 236782 & 107 & 138 & 236775 & 10618 & 1083 &  \\
    \bottomrule
  \end{tabular}
  \caption{Token IDs of the JSON prefix for all evaluated models in the paper.}
  \label{tab:prefix_token_ids}
\end{table*}

\begin{table*}[tp]
    \centering
    \begin{tabular}{ll}
        \toprule
        \textbf{Abbreviation} & \textbf{Checkpoint (Link to HF)} \\
        \midrule
        \texttt{GLM-4.1V-9B}      & \href{https://huggingface.co/zai-org/GLM-4.1V-9B-Thinking}{\texttt{zai-org/GLM-4.1V-9B-Thinking}} \\
        \texttt{GLM-4.6V-Flash}   & \href{https://huggingface.co/zai-org/GLM-4.6V-Flash}{\texttt{zai-org/GLM-4.6V-Flash}} \\
        \texttt{LLaVA-OV-1.5-4B}  & \href{https://huggingface.co/lmms-lab/LLaVA-OneVision-1.5-4B-Instruct}{\texttt{lmms-lab/LLaVA-OneVision-1.5-4B-Instruct}} \\
        \texttt{Ministral-3-3B}   & \href{https://huggingface.co/mistralai/Ministral-3-3B-Instruct-2512-BF16}{\texttt{mistralai/Ministral-3-3B-Instruct-2512-BF16}} \\
        \texttt{Ministral-3-14B}  & \href{https://huggingface.co/mistralai/Ministral-3-14B-Instruct-2512-BF16}{\texttt{mistralai/Ministral-3-14B-Instruct-2512-BF16}} \\
        \texttt{Qwen2.5-VL-3B}    & \href{https://huggingface.co/Qwen/Qwen2.5-VL-3B-Instruct}{\texttt{Qwen/Qwen2.5-VL-3B-Instruct}} \\
        \texttt{Qwen2.5-VL-32B}   & \href{https://huggingface.co/Qwen/Qwen2.5-VL-32B-Instruct}{\texttt{Qwen/Qwen2.5-VL-32B-Instruct}} \\
        \texttt{Qwen2.5-VL-72B}   & \href{https://huggingface.co/Qwen/Qwen2.5-VL-72B-Instruct}{\texttt{Qwen/Qwen2.5-VL-72B-Instruct}} \\
        \texttt{Qwen3-VL-2B}      & \href{https://huggingface.co/Qwen/Qwen3-VL-2B-Instruct}{\texttt{Qwen/Qwen3-VL-2B-Instruct}} \\
        \texttt{Qwen3-VL-32B}     & \href{https://huggingface.co/Qwen/Qwen3-VL-32B-Instruct}{\texttt{Qwen/Qwen3-VL-32B-Instruct}} \\
        \texttt{InternVL3.5-4B}   & \href{https://huggingface.co/OpenGVLab/InternVL3_5-4B}{\texttt{OpenGVLab/InternVL3\_5-4B}} \\
        \texttt{InternVL3.5-14B}  & \href{https://huggingface.co/OpenGVLab/InternVL3_5-14B}{\texttt{OpenGVLab/InternVL3\_5-14B}} \\
        \midrule
        \multicolumn{2}{l}{\textit{Domain-specialized ablation}} \\
        \texttt{medgemma-4b-it}  & \href{https://huggingface.co/google/medgemma-4b-it}{\texttt{google/medgemma-4b-it}} \\
        \texttt{medgemma-27b-it}  & \href{https://huggingface.co/google/medgemma-27b-it}{\texttt{google/medgemma-27b-it}} \\
        \bottomrule
    \end{tabular}
    \caption{Model abbreviations and hyperlinked checkpoint paths.}
    \label{tab:hf_models}
\end{table*}

\section{Parcellation in Detail}
For parcellation, we use the Harvard-Oxford probabilistic atlas~\cite{havard-oxford}, which delineates the cerebral cortex into 48 anatomically defined regions of interest (ROIs, e.g., Insular Cortex, Temporal Pole). The atlas was constructed by spatially normalizing manually labeled T1-weighted scans of healthy adults into MNI space and computing voxelwise label probabilities across subjects. We apply the cortical component of this atlas to extract per-region volumetric measurements, which are then serialized into descriptive text following the same approach as the tabular data.

\paragraph{MRI visualization}
For each scan, we render sagittal, coronal, and axial slices annotated with crosshairs at MNI coordinates $(x, y, z)$, hemisphere labels (L/R), voxel contour overlays, and a zero-anchored colormap to enhance tissue contrast.

\section{Models in Detail}
Table~\ref{tab:hf_models} shows the abbreviations and checkpoints of the models we used in this work.

\section{\textsc{FOR2107} Variables}~\label{app:for2107-variables}
\begin{figure*}[tp]
    \centering
    \begin{subfigure}{0.49\textwidth}
        \centering
        \includegraphics[width=\linewidth]{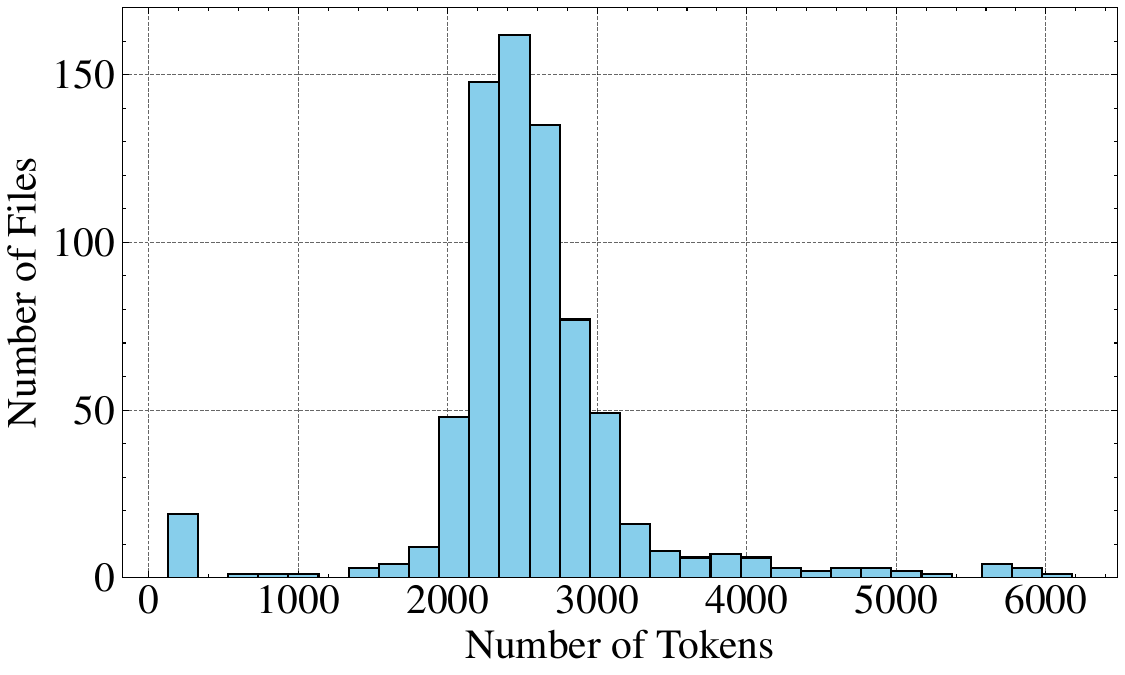}
        \caption{Major Depressive Disorder group.}
        \label{fig:mdd}
    \end{subfigure}
    \hfill
    \begin{subfigure}{0.49\textwidth}
        \centering
        \includegraphics[width=\linewidth]{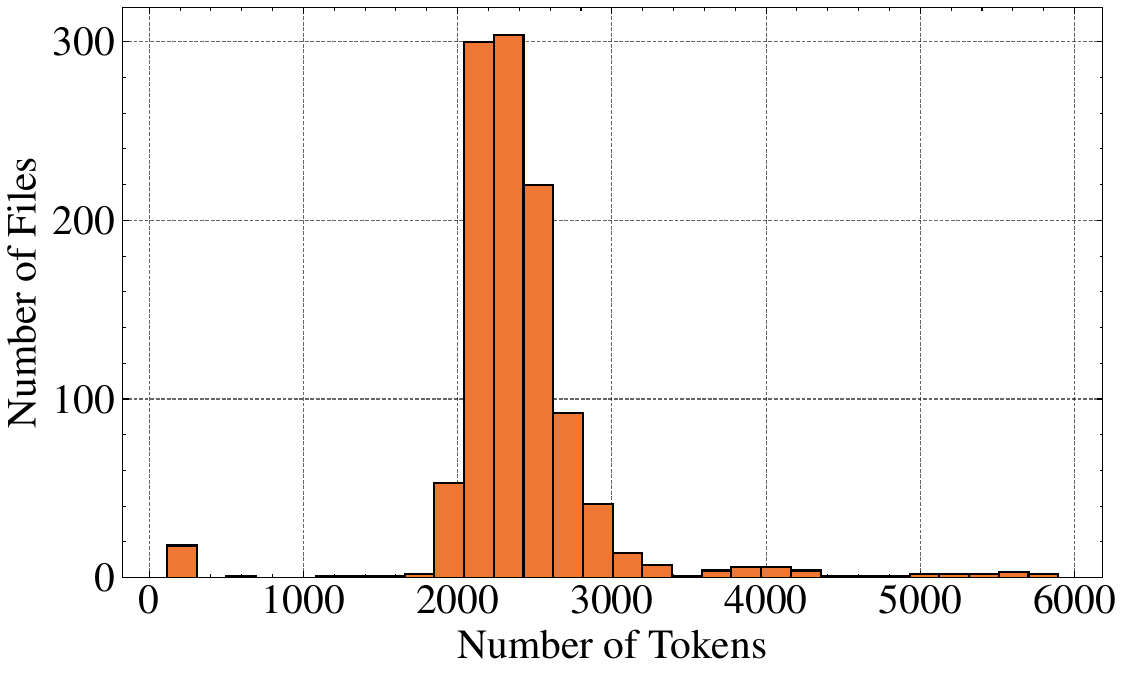}
        \caption{Control group.}
        \label{fig:control}
    \end{subfigure}
    \caption{Distribution of input-token counts in the \textsc{FOR2107} dataset.}
    \label{fig:for2107_token}
\end{figure*}

Tables~\ref{tab:demographics},~\ref{tab:rs_pss},~\ref{tab:fsozu},~\ref{tab:leq_1},~\ref{tab:leq_2},~\ref{tab:leq_3},~\ref{tab:sociodemo}, and~\ref{tab:fzt_shaps_stait} list the variables we serialize into text. Figure~\ref{fig:for2107_token} shows the distribution of input-token counts in \textsc{FOR2107}, split by the Major Depressive Disorder and Control groups. We use \texttt{tiktoken}\footnote{\url{https://github.com/openai/tiktoken}} for tokenization.

\begin{figure*}[tp]
    \centering
    \begin{subfigure}{0.49\textwidth}
        \centering
        \includegraphics[width=\linewidth]{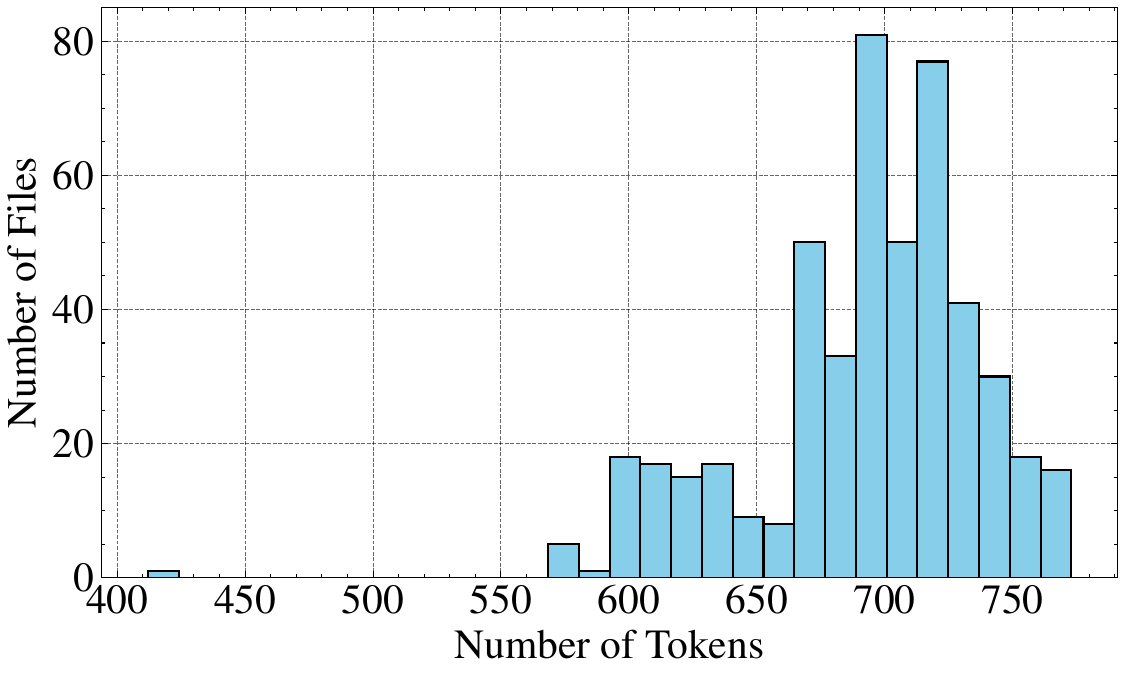}
        \caption{Cognitive Decline group.}
        \label{fig:cd}
    \end{subfigure}
    \hfill
    \begin{subfigure}{0.49\textwidth}
        \centering
        \includegraphics[width=\linewidth]{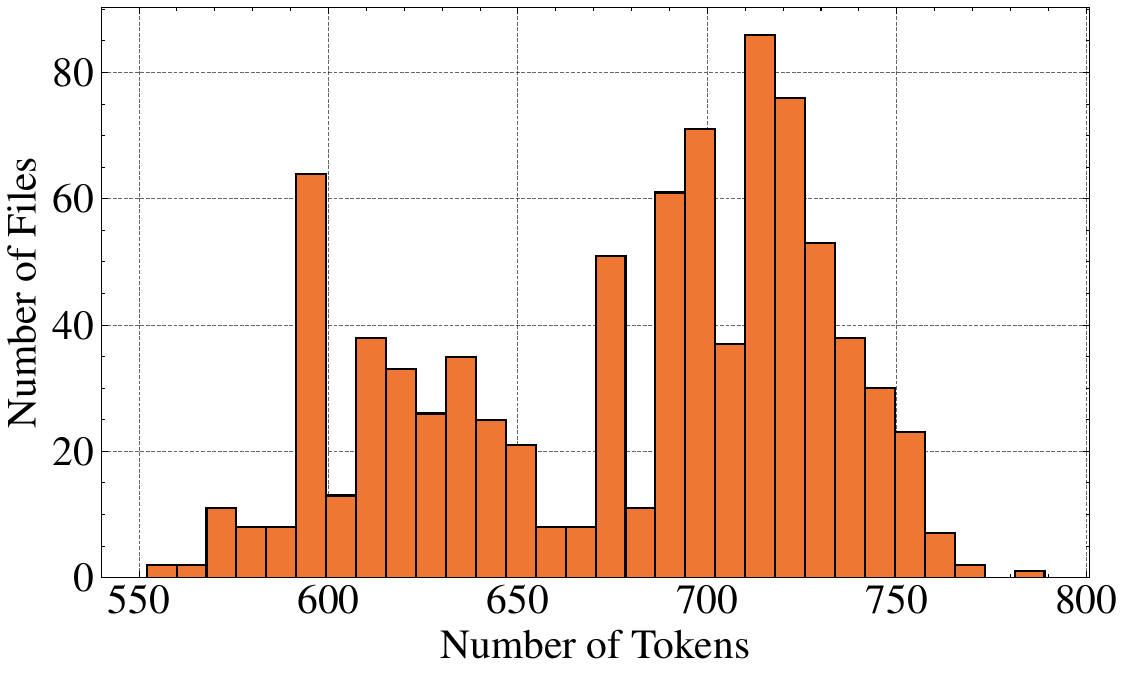}
        \caption{Cognitive Normal group.}
        \label{fig:cn}
    \end{subfigure}
    \caption{Distribution of input-token counts in the \textsc{OASIS-3} dataset.}
    \label{fig:oasis_token}
\end{figure*}

\section{\textsc{OASIS-3} Variables}~\label{app:oasis-3-variables}
Tables~\ref{tab:oasis_demographics},~\ref{tab:oasis_informant},~\ref{tab:oasis_lifestyle_sensory}, and~\ref{tab:oasis_medical} list the variables we serialize into text. Figure~\ref{fig:oasis_token} shows the distribution of input-token counts in \textsc{OASIS-3}, split by the Cognitive Normal and Cognitive Decline groups. We also use \texttt{tiktoken} for tokenization.

\section{Random Baseline}~\label{app:random_baseline}
To contextualize model performance, we report a random baseline corresponding to a stratified random classifier that predicts the positive class with a probability equal to its prevalence in the cohort. Our model tables report binary F1 with the positive class defined as MDD for \textsc{FOR2107} and Cognitive Decline for \textsc{OASIS-3}. For a stratified random classifier, both precision and recall equal the positive class prevalence $p_{+}$, so the expected F1 reduces to $p_{+}$ itself:
\[
F1_{\mathrm{random}} = \frac{2 \cdot p_{+} \cdot p_{+}}{p_{+} + p_{+}} = p_{+}.
\]
On \textsc{FOR2107} this gives $F1_{\mathrm{random}} = 701/1772 = 0.40$, and on \textsc{OASIS-3} it gives $F1_{\mathrm{random}} = 487/1336 = 0.36$.

\section{Out-of-Distribution Images for C5}\label{app:ood_images}
\begin{figure*}[tp]
    \centering
    \begin{subfigure}[t]{0.48\linewidth}
        \centering
        \includegraphics[width=\linewidth]{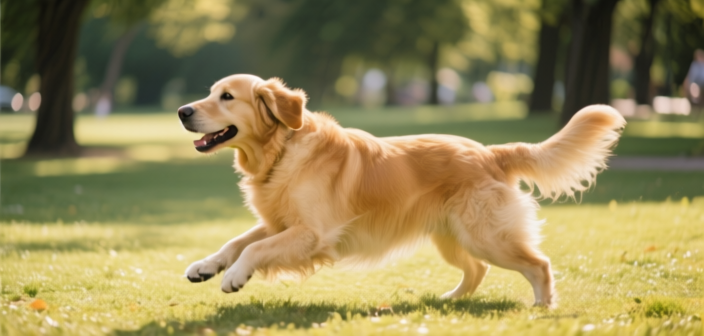}
        \caption{Natural photograph (dog).}
        \label{fig:ood_dog}
    \end{subfigure}
    \hfill
    \begin{subfigure}[t]{0.48\linewidth}
        \centering
        \includegraphics[width=\linewidth]{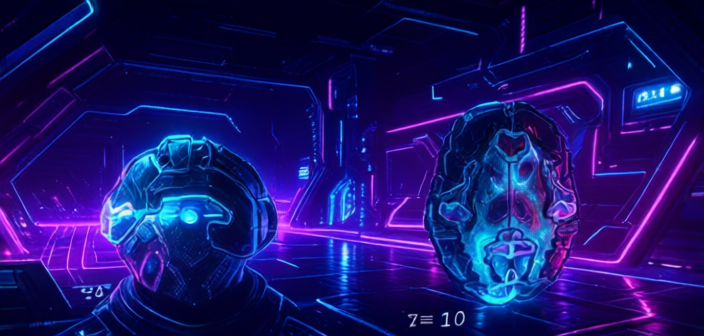}
        \caption{Stylized non-clinical brain rendering.}
        \label{fig:ood_brain}
    \end{subfigure}
    \caption{Out-of-distribution images used in condition C5. Both replace the subject-specific MRI plot while the rest of the input remains unchanged.}
    \label{fig:ood_images}
\end{figure*}

Condition C5 replaces the subject-specific MRI plot with an out-of-distribution image while retaining all other input components. Two image variants are used across the experiments, shown in Figure~\ref{fig:ood_images}: an unrelated natural photograph (a dog) and a stylized non-clinical brain rendering. Neither image carries diagnostic information about the participant. Because the reported C5 results are not stratified by image variant, we interpret C5 only as a broad control for image presence and do not attribute differences to either variant.

\section{Full Results on \textsc{FOR2107} and \textsc{OASIS-3}}
The following sections report complete results for \textsc{FOR2107} and \textsc{OASIS-3}, beginning with approximate confidence intervals for condition-wise F1 differences.

\subsection{Approximate Confidence Intervals for F1 Differences}\label{app:f1_differences}
Tables~\ref{tab:f1_diff_for2107} and~\ref{tab:f1_diff_oasis} report arithmetic differences between the F1 scores in Tables~\ref{tab:for2107_full} and~\ref{tab:oasis3_full}, together with approximate 95\% confidence intervals. We estimate the intervals from the reported aggregate metrics and cohort class totals using 10{,}000 class-stratified bootstrap resamples (random seed 0). We show the four smaller models examined in the confidence analysis, for which the preamble-associated changes are most visible.

The C2 $-$ C1 difference is positive for all four models in both cohorts, ranging from $+0.119$ to $+0.575$ on \textsc{FOR2107} and from $+0.020$ to $+0.327$ on \textsc{OASIS-3}. The approximate intervals exclude zero for all four \textsc{FOR2107} models and for \texttt{Qwen2.5-VL-3B} and \texttt{GLM-4.1V-9B} on \textsc{OASIS-3}. Its magnitude is smaller and less uniform on \textsc{OASIS-3}, consistent with the cross-cohort differences discussed in the Limitations.

Adding parcellation text and the subject-specific MRI on top of the preamble, C4 $-$ C2, changes F1 in both directions on \textsc{FOR2107}. The largest increase is $+0.197$ for \texttt{Ministral-3-3B}, while \texttt{Qwen2.5-VL-3B} and \texttt{InternVL3.5-4B} change by less than $0.01$. On \textsc{OASIS-3}, all four differences are positive but remain smaller than $0.06$, and all four approximate intervals include zero.

The C5 $-$ C4 difference also varies across models. It ranges from $-0.002$ to $+0.147$ on \textsc{FOR2107} and from $+0.005$ to $+0.067$ on \textsc{OASIS-3}. Only the intervals for \texttt{InternVL3.5-4B} on \textsc{FOR2107} and \texttt{GLM-4.1V-9B} on \textsc{OASIS-3} exclude zero, so replacing the subject-specific image does not produce a uniform drop in F1.

\begin{table*}[tp]
    \centering
    \small
    \begin{tabular}{lccc}
    \toprule
    Model & $\Delta$F1 (C2 $-$ C1) & $\Delta$F1 (C4 $-$ C2) & $\Delta$F1 (C5 $-$ C4) \\
    \midrule
    \texttt{Qwen2.5-VL-3B}   & +0.575 [+0.531, +0.620] & $-$0.005 [$-$0.039, +0.030] & +0.005 [$-$0.024, +0.035] \\
    \texttt{Ministral-3-3B}  & +0.416 [+0.370, +0.461] & +0.197 [+0.148, +0.247] & $-$0.002 [$-$0.045, +0.040] \\
    \texttt{InternVL3.5-4B}  & +0.255 [+0.215, +0.295] & $-$0.003 [$-$0.059, +0.053] & +0.147 [+0.090, +0.205] \\
    \texttt{GLM-4.1V-9B}     & +0.119 [+0.068, +0.170] & +0.080 [+0.035, +0.125] & +0.012 [$-$0.030, +0.052] \\
    \bottomrule
    \end{tabular}
    \caption{Condition-wise F1 differences with approximate 95\% confidence intervals for \textsc{FOR2107}, calculated from Table~\ref{tab:for2107_full}. C2 $-$ C1 compares the MRI preamble with the text-only baseline, C4 $-$ C2 adds parcellation text and the subject-specific MRI, and C5 $-$ C4 replaces that MRI with an out-of-distribution image.}
    \label{tab:f1_diff_for2107}
    \end{table*}

    \begin{table*}[tp]
    \centering
    \small
    \begin{tabular}{lccc}
    \toprule
    Model & $\Delta$F1 (C2 $-$ C1) & $\Delta$F1 (C4 $-$ C2) & $\Delta$F1 (C5 $-$ C4) \\
    \midrule
    \texttt{Qwen2.5-VL-3B}   & +0.327 [+0.262, +0.391] & +0.006 [$-$0.053, +0.066] & +0.024 [$-$0.033, +0.081] \\
    \texttt{Ministral-3-3B}  & +0.037 [$-$0.010, +0.087] & +0.022 [$-$0.019, +0.061] & +0.005 [$-$0.041, +0.052] \\
    \texttt{InternVL3.5-4B}  & +0.020 [$-$0.008, +0.047] & +0.019 [$-$0.014, +0.053] & +0.020 [$-$0.017, +0.055] \\
    \texttt{GLM-4.1V-9B}     & +0.118 [+0.051, +0.184] & +0.059 [$-$0.008, +0.125] & +0.067 [+0.002, +0.132] \\
    \bottomrule
    \end{tabular}
    \caption{Condition-wise F1 differences with approximate 95\% confidence intervals for \textsc{OASIS-3}, calculated from Table~\ref{tab:oasis3_full}. The three comparisons follow the definitions in Table~\ref{tab:f1_diff_for2107}.}
    \label{tab:f1_diff_oasis}
\end{table*}

\subsection{\textsc{FOR2107} -- Per-Condition Results}
Table~\ref{tab:for2107_full} reports F1, Precision, Recall, and Accuracy for all twelve models across C1--C5. The main-text pattern is clearest for several small models that fail to exceed the random baseline under C1 (\texttt{InternVL3.5-4B} $0.000$, \texttt{Ministral-3-3B} $0.064$, \texttt{Qwen2.5-VL-3B} $0.153$): adding the MRI preamble (C2) produces large gains for these models, while the remaining models show a mix of increases and decreases. Adding the subject-specific MRI (C3) and parcellation features (C4) yields mixed changes, and the out-of-distribution swap (C5) does not consistently reduce performance relative to C4 (e.g.\ \texttt{Qwen2.5-VL-72B} scores 0.845 under C4 and 0.848 under C5). This pattern suggests sensitivity to preamble framing; added image content provides no clear additional benefit.

\begin{table*}[tp]
\centering
\scriptsize
\setlength{\tabcolsep}{3.5pt}
\renewcommand{\arraystretch}{1.1}
\begin{minipage}[t]{0.495\textwidth}
\centering
\begin{tabular}[t]{lcccc}
\toprule
\textbf{Model} & \textbf{F1} & \textbf{P} & \textbf{R} & \textbf{ACC} \\
\midrule
\multicolumn{5}{l}{\textit{C1:}~\textsc{text(csv)}} \\
\midrule
\texttt{InternVL3.5-4B}   & 0.000          & 0.000 & 0.000 & 0.604 \\
\texttt{InternVL3.5-14B}  & 0.559          & 0.946 & 0.397 & 0.752 \\
\texttt{GLM-4.1V-9B}      & 0.521          & 0.958 & 0.358 & 0.740 \\
\texttt{GLM-4.6V-Flash}   & 0.795          & 0.896 & 0.715 & 0.854 \\
\texttt{LLaVA-OV-1.5-4B}  & 0.663          & 0.954 & 0.508 & 0.796 \\
\texttt{Ministral-3-3B}   & 0.064          & 1.000 & 0.033 & 0.617 \\
\texttt{Ministral-3-14B}  & 0.588          & 0.983 & 0.419 & 0.767 \\
\texttt{Qwen2.5-VL-3B}    & 0.153          & 1.000 & 0.083 & 0.637 \\
\texttt{Qwen2.5-VL-32B}   & 0.749          & 0.900 & 0.642 & 0.830 \\
\texttt{Qwen2.5-VL-72B}   & \textbf{0.828} & 0.853 & 0.805 & 0.868 \\
\texttt{Qwen3-VL-2B}      & 0.546          & 0.869 & 0.398 & 0.738 \\
\texttt{Qwen3-VL-32B}     & 0.744          & 0.928 & 0.621 & 0.831 \\
\midrule
\multicolumn{5}{l}{\textit{C2:}~\textsc{text(csv)} $+$ \textsc{prompt(mri)}} \\
\midrule
\texttt{InternVL3.5-4B}   & 0.255          & 0.963 & 0.147 & 0.660 \\
\texttt{InternVL3.5-14B}  & 0.729          & 0.884 & 0.621 & 0.818 \\
\texttt{GLM-4.1V-9B}      & 0.640          & 0.915 & 0.492 & 0.781 \\
\texttt{GLM-4.6V-Flash}   & 0.802          & 0.847 & 0.760 & 0.851 \\
\texttt{LLaVA-OV-1.5-4B}  & 0.592          & 0.980 & 0.424 & 0.769 \\
\texttt{Ministral-3-3B}   & 0.480          & 0.720 & 0.359 & 0.691 \\
\texttt{Ministral-3-14B}  & 0.680          & 0.961 & 0.526 & 0.804 \\
\texttt{Qwen2.5-VL-3B}    & 0.728          & 0.876 & 0.623 & 0.816 \\
\texttt{Qwen2.5-VL-32B}   & 0.820          & 0.897 & 0.756 & 0.869 \\
\texttt{Qwen2.5-VL-72B}   & \textbf{0.844} & 0.804 & 0.889 & 0.870 \\
\texttt{Qwen3-VL-2B}      & 0.280          & 0.966 & 0.164 & 0.667 \\
\texttt{Qwen3-VL-32B}     & 0.823          & 0.900 & 0.757 & 0.871 \\
\midrule
\multicolumn{5}{l}{\textit{C3:}~\textsc{text(csv)} $+$ \textsc{prompt(mri)} $+$ \textsc{plot(MRI)}} \\
\midrule
\texttt{InternVL3.5-4B}   & 0.296          & 0.992 & 0.174 & 0.673 \\
\texttt{InternVL3.5-14B}  & 0.540          & 0.981 & 0.372 & 0.749 \\
\texttt{GLM-4.1V-9B}      & 0.733          & 0.887 & 0.625 & 0.820 \\
\texttt{GLM-4.6V-Flash}   & 0.794          & 0.875 & 0.728 & 0.851 \\
\texttt{LLaVA-OV-1.5-4B}  & 0.713          & 0.937 & 0.575 & 0.817 \\
\texttt{Ministral-3-3B}   & 0.722          & 0.851 & 0.628 & 0.809 \\
\texttt{Ministral-3-14B}  & 0.710          & 0.948 & 0.568 & 0.817 \\
\texttt{Qwen2.5-VL-3B}    & 0.680          & 0.878 & 0.555 & 0.793 \\
\texttt{Qwen2.5-VL-32B}   & 0.820          & 0.840 & 0.802 & 0.861 \\
\texttt{Qwen2.5-VL-72B}   & \textbf{0.846} & 0.795 & 0.903 & 0.870 \\
\texttt{Qwen3-VL-2B}      & 0.447          & 0.908 & 0.297 & 0.710 \\
\texttt{Qwen3-VL-32B}     & \textbf{0.846} & 0.827 & 0.866 & 0.875 \\
\bottomrule
\end{tabular}
\end{minipage}
\hfill
\begin{minipage}[t]{0.495\textwidth}
\centering
\begin{tabular}[t]{lcccc}
\toprule
\textbf{Model} & \textbf{F1} & \textbf{P} & \textbf{R} & \textbf{ACC} \\
\midrule
\multicolumn{5}{l}{\textit{C4:}~\textsc{text(csv, parcel)} $+$ \textsc{prompt(mri)} $+$ \textsc{plot(MRI)}} \\
\midrule
\texttt{InternVL3.5-4B}   & 0.252          & 0.990 & 0.144 & 0.661 \\
\texttt{InternVL3.5-14B}  & 0.587          & 0.913 & 0.432 & 0.759 \\
\texttt{GLM-4.1V-9B}      & 0.720          & 0.916 & 0.593 & 0.818 \\
\texttt{GLM-4.6V-Flash}   & 0.791          & 0.848 & 0.740 & 0.845 \\
\texttt{LLaVA-OV-1.5-4B}  & 0.603          & 0.975 & 0.437 & 0.773 \\
\texttt{Ministral-3-3B}   & 0.677          & 0.903 & 0.542 & 0.796 \\
\texttt{Ministral-3-14B}  & 0.680          & 0.966 & 0.525 & 0.805 \\
\texttt{Qwen2.5-VL-3B}    & 0.723          & 0.632 & 0.845 & 0.744 \\
\texttt{Qwen2.5-VL-32B}   & 0.779          & 0.896 & 0.689 & 0.845 \\
\texttt{Qwen2.5-VL-72B}   & \textbf{0.845} & 0.798 & 0.897 & 0.870 \\
\texttt{Qwen3-VL-2B}      & 0.347          & 0.974 & 0.211 & 0.686 \\
\texttt{Qwen3-VL-32B}     & 0.835          & 0.865 & 0.806 & 0.874 \\
\midrule
\multicolumn{5}{l}{\textit{C5:}~\textsc{text(csv, parcel)} $+$ \textsc{prompt(mri)} $+$ \textsc{plot(swap)}} \\
\midrule
\texttt{InternVL3.5-4B}   & 0.399          & 0.972 & 0.251 & 0.701 \\
\texttt{InternVL3.5-14B}  & 0.703          & 0.902 & 0.576 & 0.808 \\
\texttt{GLM-4.1V-9B}      & 0.732          & 0.888 & 0.622 & 0.819 \\
\texttt{GLM-4.6V-Flash}   & 0.788          & 0.864 & 0.725 & 0.846 \\
\texttt{LLaVA-OV-1.5-4B}  & 0.540          & 0.992 & 0.371 & 0.750 \\
\texttt{Ministral-3-3B}   & 0.675          & 0.759 & 0.608 & 0.769 \\
\texttt{Ministral-3-14B}  & 0.697          & 0.955 & 0.549 & 0.812 \\
\texttt{Qwen2.5-VL-3B}    & 0.728          & 0.666 & 0.803 & 0.763 \\
\texttt{Qwen2.5-VL-32B}   & 0.780          & 0.902 & 0.686 & 0.847 \\
\texttt{Qwen2.5-VL-72B}   & \textbf{0.848} & 0.838 & 0.859 & 0.879 \\
\texttt{Qwen3-VL-2B}      & 0.365          & 0.935 & 0.227 & 0.688 \\
\texttt{Qwen3-VL-32B}     & 0.822          & 0.893 & 0.762 & 0.870 \\
\bottomrule
\end{tabular}
\end{minipage}

\caption{Full results on \textsc{FOR2107} across all five input conditions (C1--C5) and all twelve models. Metrics are F1, Precision (P), Recall (R), and Accuracy (ACC). Best F1 per condition is shown in \textbf{bold}. Left panel: text-only and preamble/image conditions (C1--C3). Right panel: parcellation-augmented and out-of-distribution swap conditions (C4--C5).}
\label{tab:for2107_full}
\end{table*}

\subsection{\textsc{OASIS-3} -- Per-Condition Results}
Table~\ref{tab:oasis3_full} reports F1, Precision, Recall, and Accuracy for all twelve models across C1--C5 on \textsc{OASIS-3}, a cohort in a different clinical domain (cognitive decline vs.\ depression). The results are more mixed than on \textsc{FOR2107}. Several models that are already competitive under C1 (e.g.\ \texttt{InternVL3.5-14B}, $0.773$) perform worse when the MRI preamble is added. Unlike on \textsc{FOR2107}, the subject-specific MRI (C3) tends to help more than the preamble alone, and parcellation features (C4) again add little. The out-of-distribution swap (C5) does not consistently reduce performance relative to C4 (e.g.\ \texttt{Qwen2.5-VL-72B} scores 0.775 under C4 and 0.778 under C5). This C5--C4 comparison, in which the C4 image is replaced, does not establish that all \textsc{OASIS-3} gains arise from the preamble, especially because C3 often exceeds C2. The \textsc{OASIS-3} results therefore show heterogeneous responses to prompt framing and the supplied image.

\begin{table*}[tp]
\centering
\scriptsize
\setlength{\tabcolsep}{3.5pt}
\renewcommand{\arraystretch}{1.1}
\begin{minipage}[t]{0.495\textwidth}
\centering
\begin{tabular}[t]{lcccc}
\toprule
\textbf{Model} & \textbf{F1} & \textbf{P} & \textbf{R} & \textbf{ACC} \\
\midrule
\multicolumn{5}{l}{\textit{C1:}~\textsc{text(csv)}} \\
\midrule
\texttt{InternVL3.5-4B}   & 0.616          & 0.470 & 0.893 & 0.594 \\
\texttt{InternVL3.5-14B}  & 0.773          & 0.865 & 0.698 & 0.850 \\
\texttt{GLM-4.1V-9B}      & 0.278          & 0.909 & 0.164 & 0.689 \\
\texttt{GLM-4.6V-Flash}   & 0.359          & 0.901 & 0.224 & 0.708 \\
\texttt{LLaVA-OV-1.5-4B}  & 0.679          & 0.591 & 0.799 & 0.725 \\
\texttt{Ministral-3-3B}   & 0.504          & 0.692 & 0.396 & 0.716 \\
\texttt{Ministral-3-14B}  & 0.602          & 0.878 & 0.458 & 0.779 \\
\texttt{Qwen2.5-VL-3B}    & 0.262          & 0.961 & 0.152 & 0.689 \\
\texttt{Qwen2.5-VL-32B}   & 0.539          & 0.374 & 0.963 & 0.400 \\
\texttt{Qwen2.5-VL-72B}   & \textbf{0.786} & 0.927 & 0.682 & 0.865 \\
\texttt{Qwen3-VL-2B}      & 0.090          & 0.920 & 0.047 & 0.651 \\
\texttt{Qwen3-VL-32B}     & 0.582          & 0.945 & 0.421 & 0.780 \\
\midrule
\multicolumn{5}{l}{\textit{C2:}~\textsc{text(csv)} $+$ \textsc{prompt(mri)}} \\
\midrule
\texttt{InternVL3.5-4B}   & 0.636          & 0.509 & 0.848 & 0.646 \\
\texttt{InternVL3.5-14B}  & 0.584          & 0.421 & 0.953 & 0.505 \\
\texttt{GLM-4.1V-9B}      & 0.396          & 0.696 & 0.277 & 0.692 \\
\texttt{GLM-4.6V-Flash}   & 0.465          & 0.835 & 0.322 & 0.730 \\
\texttt{LLaVA-OV-1.5-4B}  & 0.615          & 0.464 & 0.912 & 0.584 \\
\texttt{Ministral-3-3B}   & 0.541          & 0.425 & 0.745 & 0.540 \\
\texttt{Ministral-3-14B}  & 0.555          & 0.387 & 0.982 & 0.426 \\
\texttt{Qwen2.5-VL-3B}    & 0.589          & 0.910 & 0.435 & 0.778 \\
\texttt{Qwen2.5-VL-32B}   & 0.534          & 0.365 & 1.000 & 0.365 \\
\texttt{Qwen2.5-VL-72B}   & \textbf{0.778} & 0.902 & 0.684 & 0.858 \\
\texttt{Qwen3-VL-2B}      & 0.028          & 1.000 & 0.014 & 0.641 \\
\texttt{Qwen3-VL-32B}     & 0.767          & 0.887 & 0.676 & 0.850 \\
\midrule
\multicolumn{5}{l}{\textit{C3:}~\textsc{text(csv)} $+$ \textsc{prompt(mri)} $+$ \textsc{plot(MRI)}} \\
\midrule
\texttt{InternVL3.5-4B}   & 0.634          & 0.502 & 0.860 & 0.638 \\
\texttt{InternVL3.5-14B}  & 0.731          & 0.726 & 0.735 & 0.802 \\
\texttt{GLM-4.1V-9B}      & 0.608          & 0.827 & 0.480 & 0.774 \\
\texttt{GLM-4.6V-Flash}   & 0.705          & 0.915 & 0.573 & 0.825 \\
\texttt{LLaVA-OV-1.5-4B}  & 0.620          & 0.468 & 0.920 & 0.589 \\
\texttt{Ministral-3-3B}   & 0.592          & 0.498 & 0.729 & 0.633 \\
\texttt{Ministral-3-14B}  & 0.652          & 0.522 & 0.867 & 0.662 \\
\texttt{Qwen2.5-VL-3B}    & 0.501          & 0.885 & 0.349 & 0.746 \\
\texttt{Qwen2.5-VL-32B}   & 0.688          & 0.625 & 0.766 & 0.747 \\
\texttt{Qwen2.5-VL-72B}   & \textbf{0.776} & 0.828 & 0.731 & 0.847 \\
\texttt{Qwen3-VL-2B}      & 0.008          & 1.000 & 0.004 & 0.637 \\
\texttt{Qwen3-VL-32B}     & 0.713          & 0.641 & 0.803 & 0.764 \\
\bottomrule
\end{tabular}
\end{minipage}
\hfill
\begin{minipage}[t]{0.495\textwidth}
\centering
\begin{tabular}[t]{lcccc}
\toprule
\textbf{Model} & \textbf{F1} & \textbf{P} & \textbf{R} & \textbf{ACC} \\
\midrule
\multicolumn{5}{l}{\textit{C4:}~\textsc{text(csv, parcel)} $+$ \textsc{prompt(mri)} $+$ \textsc{plot(MRI)}} \\
\midrule
\texttt{InternVL3.5-4B}   & 0.655          & 0.563 & 0.784 & 0.699 \\
\texttt{InternVL3.5-14B}  & 0.738          & 0.783 & 0.698 & 0.820 \\
\texttt{GLM-4.1V-9B}      & 0.455          & 0.867 & 0.308 & 0.731 \\
\texttt{GLM-4.6V-Flash}   & 0.659          & 0.836 & 0.544 & 0.795 \\
\texttt{LLaVA-OV-1.5-4B}  & 0.680          & 0.570 & 0.844 & 0.711 \\
\texttt{Ministral-3-3B}   & 0.563          & 0.521 & 0.612 & 0.653 \\
\texttt{Ministral-3-14B}  & 0.675          & 0.670 & 0.680 & 0.761 \\
\texttt{Qwen2.5-VL-3B}    & 0.595          & 0.799 & 0.474 & 0.765 \\
\texttt{Qwen2.5-VL-32B}   & 0.712          & 0.889 & 0.593 & 0.825 \\
\texttt{Qwen2.5-VL-72B}   & \textbf{0.775} & 0.878 & 0.694 & 0.853 \\
\texttt{Qwen3-VL-2B}      & 0.004          & 1.000 & 0.002 & 0.636 \\
\texttt{Qwen3-VL-32B}     & 0.727          & 0.860 & 0.630 & 0.828 \\
\midrule
\multicolumn{5}{l}{\textit{C5:}~\textsc{text(csv, parcel)} $+$ \textsc{prompt(mri)} $+$ \textsc{plot(swap)}} \\
\midrule
\texttt{InternVL3.5-4B}   & 0.675          & 0.593 & 0.782 & 0.725 \\
\texttt{InternVL3.5-14B}  & 0.683          & 0.658 & 0.710 & 0.760 \\
\texttt{GLM-4.1V-9B}      & 0.522          & 0.899 & 0.368 & 0.754 \\
\texttt{GLM-4.6V-Flash}   & 0.696          & 0.813 & 0.608 & 0.806 \\
\texttt{LLaVA-OV-1.5-4B}  & 0.677          & 0.579 & 0.815 & 0.716 \\
\texttt{Ministral-3-3B}   & 0.568          & 0.556 & 0.581 & 0.678 \\
\texttt{Ministral-3-14B}  & 0.736          & 0.781 & 0.696 & 0.818 \\
\texttt{Qwen2.5-VL-3B}    & 0.619          & 0.833 & 0.493 & 0.779 \\
\texttt{Qwen2.5-VL-32B}   & 0.713          & 0.870 & 0.604 & 0.823 \\
\texttt{Qwen2.5-VL-72B}   & \textbf{0.778} & 0.902 & 0.684 & 0.858 \\
\texttt{Qwen3-VL-2B}      & 0.008          & 1.000 & 0.004 & 0.637 \\
\texttt{Qwen3-VL-32B}     & 0.667          & 0.860 & 0.544 & 0.802 \\
\bottomrule
\end{tabular}
\end{minipage}

\caption{Full results on \textsc{OASIS-3} across all five input conditions (C1--C5) and all twelve models. Metrics are F1, Precision (P), Recall (R), and Accuracy (ACC). Best F1 per condition is shown in \textbf{bold}. Left panel: text-only and preamble/image conditions (C1--C3). Right panel: parcellation-augmented and out-of-distribution swap conditions (C4--C5). Note that \texttt{Qwen3-VL-2B} produces near-degenerate outputs across all conditions on this dataset, and \texttt{Qwen2.5-VL-32B} under C1--C2 exhibits near-constant positive predictions (ACC~$\approx$~0.40, R~$\approx$~1.00), consistent with a near-degenerate prediction strategy that has little discriminative ability.}
\label{tab:oasis3_full}
\end{table*}

\paragraph{Ablation with \texttt{medgemma} models}\label{app:medgemma}
\texttt{medgemma} \cite{sellergren2026medgemmatechnicalreport} is Google's medically specialized open-weight VLM, with a SigLIP image encoder pretrained on de-identified medical imagery (including radiology) and an LLM component trained on medical text and clinical records. We include it to examine whether the same preamble-associated pattern appears in a medically pretrained VLM.

The two \texttt{medgemma} model sizes show contrasting performance patterns across the two datasets. On \textsc{FOR2107}, \texttt{medgemma-27b-it} is stronger across all conditions, with F1 rising from $0.467$ to $0.582$ when the MRI preamble is added with no imaging data; \texttt{medgemma-4b-it} produces near-degenerate predictions, with near-zero recall across C3--C5. On \textsc{OASIS-3}, the roles invert: \texttt{medgemma-4b-it} maintains strong text-only performance (C1 F1 $= 0.739$), whereas \texttt{medgemma-27b-it} collapses to a near-degenerate negative-prediction strategy under C1 and C2 (F1 $\approx 0.07$, P $= 1.000$, R $\approx 0.04$). For both \textsc{OASIS-3} models, the \emph{byte-identical} outputs under C1 and C2 suggest that predictions are more sensitive to the presence and form of the image input than to prompt framing. Neither model size matches the strongest text-only performance of the general-purpose VLMs in Table~\ref{tab:oasis3_full}. The cross-dataset role reversal between the 4B and 27B checkpoints indicates that medical-domain pretraining did not yield consistently strong classification performance across both cohorts.

\begin{table*}[tp]
    \centering
    \small
    \begin{tabular}{cccccccccc}
    \toprule
    \textbf{Dataset} & \textbf{Cond.}
      & \multicolumn{4}{c}{\texttt{medgemma-4b-it}}
      & \multicolumn{4}{c}{\texttt{medgemma-27b-it}} \\
    \cmidrule(lr){3-6} \cmidrule(lr){7-10}
     & & F1 & P & R & ACC & F1 & P & R & ACC \\
    \midrule
    \multirow{5}{*}{\textsc{FOR2107}}
      & C1                       & 0.130 & 0.891 & 0.070 & 0.629 & \textbf{0.467} & 0.982 & 0.307 & 0.724 \\
      & C2                       & 0.100 & 0.902 & 0.053 & 0.623 & \textbf{0.582} & 0.914 & 0.427 & 0.757 \\
      & C3                       & 0.003 & 1.000 & 0.001 & 0.605 & \textbf{0.625} & 0.925 & 0.472 & 0.776 \\
      & C4                       & 0.014 & 1.000 & 0.007 & 0.607 & \textbf{0.386} & 0.966 & 0.241 & 0.696 \\
      & C5                       & 0.003 & 1.000 & 0.001 & 0.605 & \textbf{0.322} & 0.951 & 0.194 & 0.677 \\
    \midrule
    \multirow{5}{*}{\textsc{OASIS-3}}
      & C1                       & \textbf{0.739} & 0.876 & 0.639 & 0.835 & 0.071 & 1.000 & 0.037 & 0.649 \\
      & C2                       & \textbf{0.739} & 0.876 & 0.639 & 0.835 & 0.071 & 1.000 & 0.037 & 0.649 \\
      & C3                       & 0.413 & 0.774 & 0.281 & 0.708 & \textbf{0.606} & 0.503 & 0.760 & 0.639 \\
      & C4                       & \textbf{0.591} & 0.445 & 0.877 & 0.556 & 0.443 & 0.740 & 0.316 & 0.710 \\
      & C5                       & \textbf{0.576} & 0.450 & 0.801 & 0.570 & 0.400 & 0.798 & 0.267 & 0.708 \\
    \bottomrule
    \end{tabular}
    \caption{Ablation of \texttt{medgemma} models on \textsc{FOR2107} and \textsc{OASIS-3}: F1, Precision (P), Recall (R), and Accuracy (ACC) across the five input conditions. \textbf{Bold} marks the higher F1 between the two model sizes per row. See Table~\ref{tab:conditions} for a description of the notation.}
    \label{tab:medgemma_appendix}
\end{table*}

\subsection{Expert Consultation}\label{app:experts}
We excluded obvious classification shortcuts in consultation with domain experts in clinical psychology. The consulted group comprised postdoctoral researchers and advanced doctoral candidates in clinical psychology or closely related fields; all had research experience involving the relevant clinical populations.

\section{Preliminary Case Study: Scoring Rubric}\label{app:human_eval}
\begin{itemize}[topsep=0pt, partopsep=0pt, parsep=0pt, itemsep=0pt, leftmargin=*]
    \item \textbf{Faithfulness}: whether the output strictly adheres to the input
    context without adding unverified information (1 = severe fabrication,
    4 = perfectly faithful).
    \item \textbf{Clinical Accuracy \& Safety}: whether the conclusions are correct,
    safe, and aligned with clinical standards (1 = completely inaccurate or unsafe,
    4 = perfectly accurate and safe).
    \item \textbf{Diagnostic Reasoning}: whether the reasoning behind the diagnosis
    is clearly and logically explained (1 = poor, 4 = excellent).
\end{itemize}
We use 3.0 as a descriptive reference because scores of 3 and 4 are considered acceptable in our rubric; this cutoff has not been clinically validated.

\section{Confidence Estimation in Detail}\label{app:confidence}
To quantify model confidence during inference on \textsc{FOR2107}, we extract the decoder's next-token distribution after the fixed JSON prefix using greedy decoding with \texttt{output\_scores=True}. Following label-token probability approaches~\cite{pmlr-v139-zhao21c, geng-etal-2024-survey}, we compare the probabilities assigned to the candidate initial label tokens \texttt{Major}, representing Major Depressive Disorder, and \texttt{Control}.

\paragraph{Label-token formulation.}
For each participant $x$ under input condition $c$, we normalize the conditional probabilities of the two candidate label tokens after the fixed JSON prefix to obtain a score for class $\ell \in \{\text{MDD}, \text{Control}\}$:
\begin{equation}
\hat{P}_c(\ell \mid x) =
\frac{P(t^\ell \mid \mathrm{prefix}, x, c)}
{\sum_{\ell'} P(t^{\ell'} \mid \mathrm{prefix}, x, c) + \epsilon}.
\label{eq:norm_prob}
\end{equation}
Here, $\epsilon = 10^{-12}$ only guards against division by zero without affecting the reported value. The predicted class is $\hat{y} = \operatorname{arg\,max}_{\ell}\hat{P}_c(\ell \mid x)$. The score $\hat{P}_c(\ell \mid x)$ is bounded in $[0,1]$ and expresses the model's relative preference between the two labels within a given (participant, condition) pair. All cross-condition comparisons (Eq.~\ref{eq:delta}) are made on this normalized quantity rather than on raw joint probabilities.

\paragraph{Per-participant confidence shift.}
We use $\hat{P}_c(\ell \mid x)$ to compare the shifts associated with prompt framing and subject-specific MRI content. Specifically, we evaluate each participant under three conditions and define the per-participant confidence shift as:
\begin{equation}
    \delta^{\text{cond}}_{\leftarrow\,\text{base}}(x_i)
    \;=\;
    \hat{P}_{\text{cond}}(\ell \mid x_i)
    - \hat{P}_{\text{base}}(\ell \mid x_i),
    \label{eq:delta}
\end{equation}
where $\ell = \text{MDD}$. A positive $\delta$ indicates that the condition increases the model's confidence in the MDD label relative to the baseline.

\subsection{Confidence Estimation for Additional Models}\label{app:conf-more}
\begin{figure*}[tp]
    \centering
    \begin{subfigure}{0.49\textwidth}
        \centering
        \includegraphics[width=\linewidth]{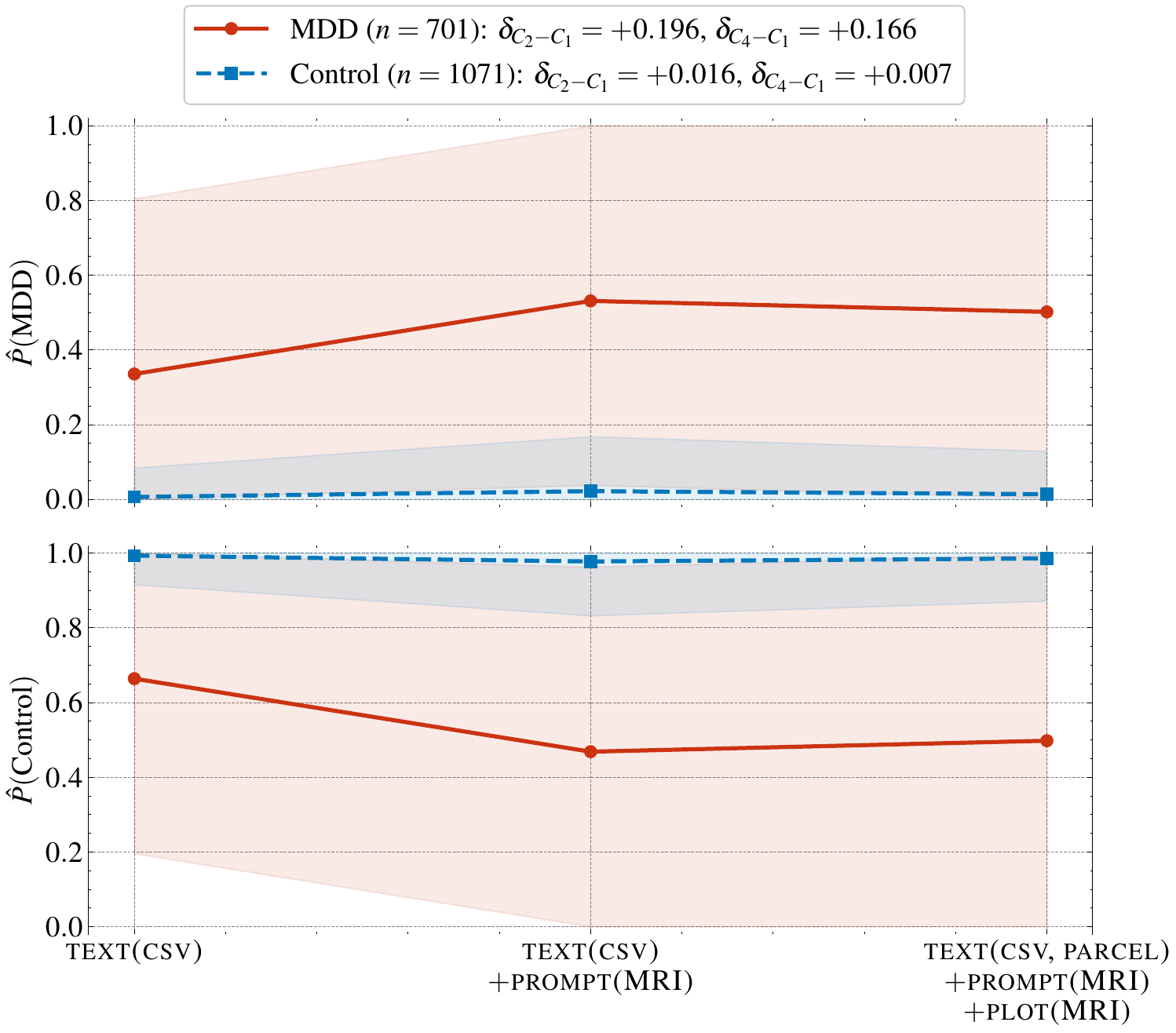}
        \caption{\texttt{GLM-4.1V-9B}}
        \label{fig:first-app}
    \end{subfigure}
    \hfill
    \begin{subfigure}{0.49\textwidth}
        \centering
        \includegraphics[width=\linewidth]{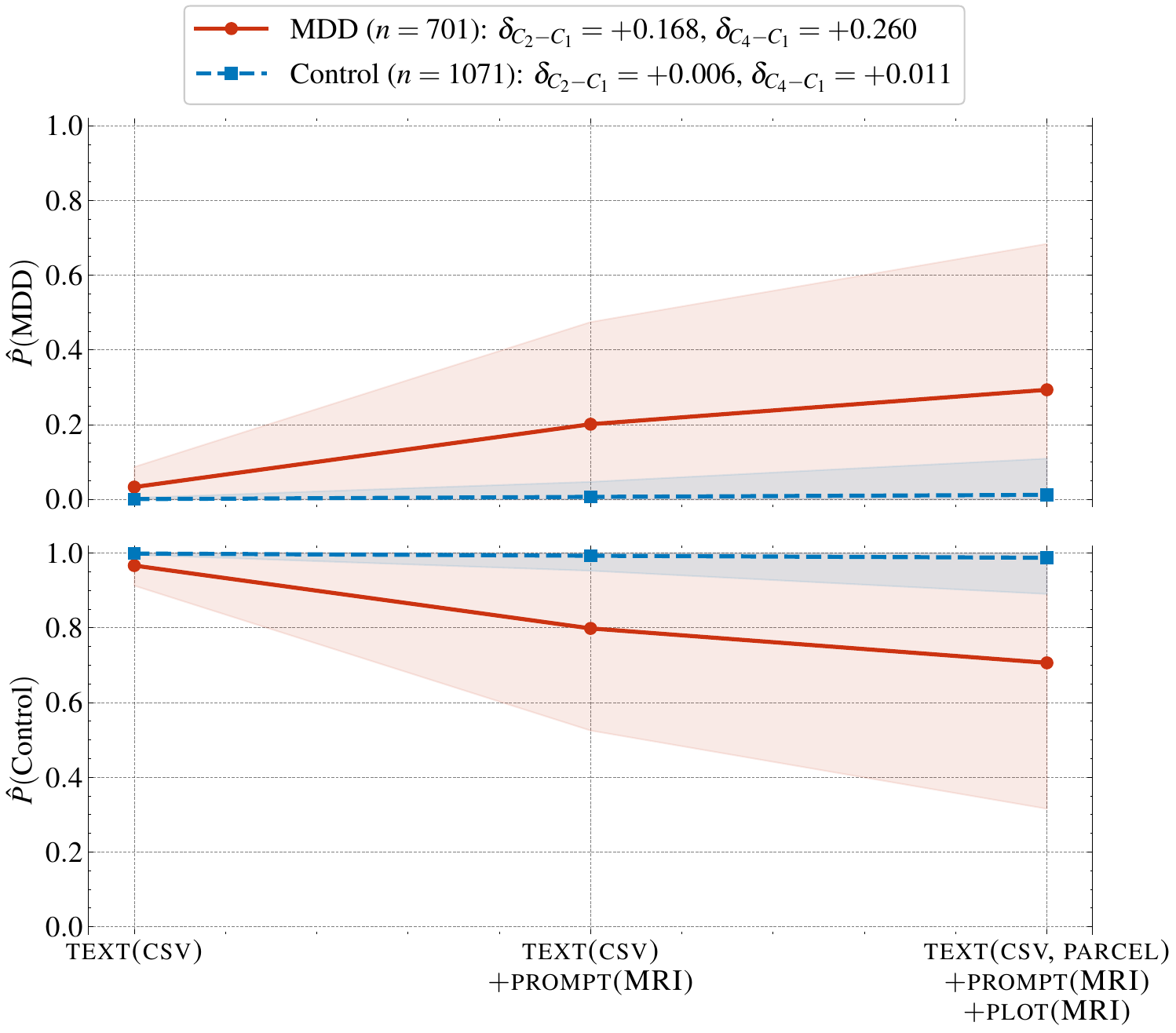}
        \caption{\texttt{InternVL3.5-4B}}
        \label{fig:second-app}
    \end{subfigure}
    \caption{\textsc{FOR2107} per-group mean $\hat{P}(\text{MDD})$ (top) and $\hat{P}(\text{Control})$ (bottom) across the three input conditions. Both diagnostic groups exhibit an upward shift in $\hat{P}(\text{MDD})$, with the complementary decrease in $\hat{P}(\text{Control})$ implied by two-class normalization. Shaded bands show one sample standard deviation around each group mean and are clipped to the $[0,1]$ probability range.}
    \label{fig:dual-2}
\end{figure*}

\begin{table}[htbp]
\centering
\small
\begin{tabular}{lccc}
\toprule
\textbf{Model} & \textbf{Cond.} & \textbf{ECE}$_{15}$ & \textbf{Brier} \\
\midrule
\multirow{3}{*}{\texttt{InternVL3.5-4B}}
  & C1 & 0.382 & 0.371 \\
  & C2 & 0.289 & 0.283 \\
  & C4 & \textbf{0.265} & \textbf{0.264} \\
\midrule
\multirow{3}{*}{\texttt{GLM-4.1V-9B}}
  & C1 & 0.265 & 0.265 \\
  & C2 & \textbf{0.196} & \textbf{0.196} \\
  & C4 & 0.204 & 0.204 \\
\bottomrule
\end{tabular}
\caption{Confidence estimation for \texttt{InternVL3.5-\allowbreak{}4B} and \texttt{GLM-\allowbreak{}4.1V-\allowbreak{}9B} on \textsc{FOR2107}, extending the analysis in Section~\ref{sec:confidence} (Table~\ref{tab:calibration}). Results are computed over all 1,772 participants (MDD and Control combined). \textbf{Bold} denotes the lower score for each model.}
\label{tab:calibration-more}
\end{table}

Table~\ref{tab:calibration-more} reports ECE and Brier under C1, C2, and C4 for \texttt{InternVL3.5-\allowbreak{}4B} and \texttt{GLM-\allowbreak{}4.1V-\allowbreak{}9B} on \textsc{FOR2107}, complementing the two-model analysis in Section~\ref{sec:confidence}. Both models reproduce the central calibration pattern reported there: the C1$\to$C2 transition, in which no imaging content is provided, accounts for the bulk of the calibration improvement. For \texttt{InternVL3.5-\allowbreak{}4B}, $79.5\%$ of the total C1$\to$C4 ECE reduction is realized at C2; for \texttt{GLM-\allowbreak{}4.1V-\allowbreak{}9B} we observe a similar trend, as C4 in fact has marginally worse calibration than C2 (ECE $0.204$ vs.\ $0.196$). Figure~\ref{fig:dual-2} reproduces the pattern observed for the original two models: $\hat{P}(\text{MDD})$ increases under C2 in both the MDD and Control groups, with a mirror-symmetric decrease in $\hat{P}(\text{Control})$. The upward shift in $\hat{P}(\text{MDD})$ also appears in the Control group, where a shift driven solely by genuine diagnostic signal would not be expected. This pattern recurs across the four evaluated models, but the analysis does not establish a shared mechanism.

\section{Preamble-Associated Shift: Formal Detail}
\label{app:preamble}

\subsection{Preamble Direction}
Let $\mathcal{M}$ be a VLM with $L$ layers and hidden dimension $d$. Let $\mathbf{h}^{(l)}_{c}(x) \in \mathbb{R}^{d}$ denote the residual-stream hidden state at layer $l$, extracted at the label-token decoding step for participant record $x$ under input condition $c$. We denote the positive class as $\ell^{+}$ (e.g.\ MDD for \textsc{FOR2107}). For brevity, we refer to the two main-text conditions of interest as $c_{\mathrm{base}}$ (C1: \textsc{text(csv)}) and $c_{\mathrm{cond}}$ (C2: \textsc{text(csv)} $+$ \textsc{prompt(mri)}). Given a set of $N$ participant records $\{x_i\}_{i=1}^{N}$, we define the \textbf{preamble direction} at layer $l^{*}$ as:
\begin{equation}
    \mathbf{d} \;=\;
    \frac{1}{N}\sum_{i=1}^{N} \Big(\mathbf{h}^{(l^{*})}_{c_{\mathrm{cond}}}(x_i) - \mathbf{h}^{(l^{*})}_{c_{\mathrm{base}}}(x_i)\Big),
     \label{eq:preamble_dir}
\end{equation}
with unit-normalized form $\mathbf{u} = \mathbf{d} / \|\mathbf{d}\|_2$. Intuitively, $\mathbf{u}$ captures the average residual-stream difference associated with adding the MRI preamble. The target layer $l^{*}$ is selected per model as the layer immediately preceding the point at which the label first emerges under a logit-lens sweep, a fixed criterion rather than one tuned to maximize the measured effect.

\subsection{Phrase Probe Inventory}
\label{app:preamble_table}

\begin{figure}[htbp]
    \centering
    \includegraphics[width=\linewidth]{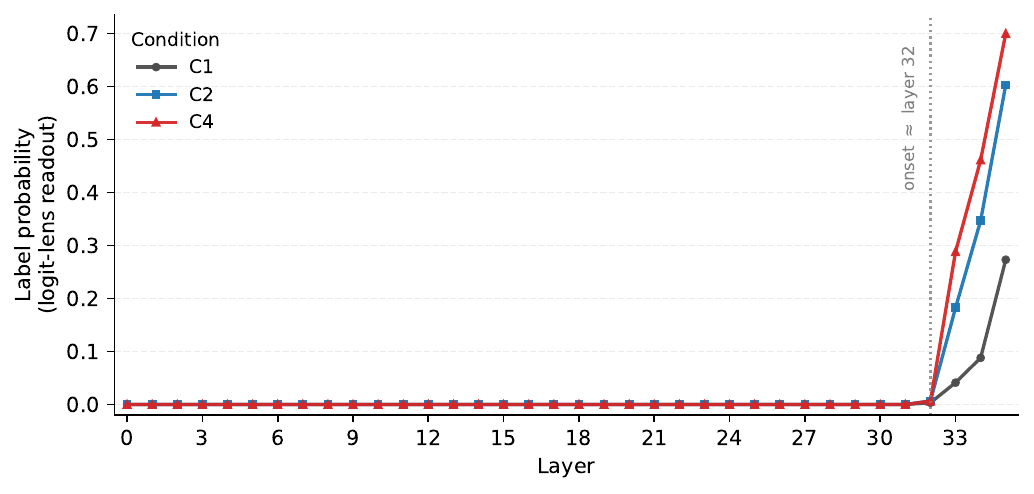}
    \caption{Logit-lens trajectory across all 36 layers of \texttt{Qwen2.5-VL-3B} on \textsc{FOR2107}, under conditions C1, C2, and C4. The label carries negligible weight through the early and middle layers and emerges only in the final layers (onset $\approx$ layer~32). Past this onset, the preamble conditions (C2, C4) sit above the C1 baseline, showing an association between the preamble conditions and higher projected weight toward the target label. We extract activations from the layer immediately preceding the onset.}
    \label{fig:logit_lens}
\end{figure}

Table~\ref{tab:preamble_phrases} lists the phrases and scores for the 19 numbered points in Figure~\ref{fig:phrase_probe}; the figure also writes out selected phrase labels directly.

\begin{table*}[tp]
    \centering\small
    \begin{tabular}{clp{7cm}rr}
    \toprule
    \# & \textbf{Category} & \textbf{Phrase} & $\cos\!\uparrow$ & $\delta\!\uparrow$ \\
    \midrule
    1 & \textcolor[RGB]{33,150,243}{\textbullet} MRI / neuroimaging & \textit{Brain MRI findings are available.} & +0.43 & +0.02 \\
    2 & \textcolor[RGB]{255,152,0}{\textbullet} General clinical & \textit{Diagnostic results are available.} & +0.24 & +0.02 \\
    3 & \textcolor[RGB]{33,150,243}{\textbullet} MRI / neuroimaging & \textit{Brain scans have been performed.} & +0.23 & -0.01 \\
    4 & \textcolor[RGB]{255,87,34}{\textbullet} Structural / format & \textit{Classification result:} & +0.00 & -0.02 \\
    5 & \textcolor[RGB]{255,152,0}{\textbullet} General clinical & \textit{A clinical diagnosis has been established.} & -0.07 & -0.01 \\
    6 & \textcolor[RGB]{255,87,34}{\textbullet} Structural / format & \textit{The answer is:} & -0.25 & -0.09 \\
    7 & \textcolor[RGB]{255,87,34}{\textbullet} Structural / format & \textit{Output: \{"category":} & -0.32 & -0.08 \\
    8 & \textcolor[RGB]{156,39,176}{\textbullet} Neutral / unrelated & \textit{This is a test of the system.} & -0.32 & -0.04 \\
    9 & \textcolor[RGB]{255,152,0}{\textbullet} General clinical & \textit{Clinical evaluation is complete.} & -0.32 & -0.06 \\
    10 & \textcolor[RGB]{156,39,176}{\textbullet} Neutral / unrelated & \textit{Answer the following question carefully.} & -0.38 & -0.05 \\
    11 & \textcolor[RGB]{76,175,80}{\textbullet} Authoritative framing & \textit{As a medical professional, review the following.} & -0.39 & -0.04 \\
    12 & \textcolor[RGB]{156,39,176}{\textbullet} Neutral / unrelated & \textit{Please process the following information.} & -0.43 & -0.04 \\
    13 & \textcolor[RGB]{255,152,0}{\textbullet} General clinical & \textit{The patient has been evaluated by a specialist.} & -0.45 & -0.06 \\
    14 & \textcolor[RGB]{76,175,80}{\textbullet} Authoritative framing & \textit{You have extensive experience in psychiatric diagnosis.} & -0.48 & -0.08 \\
    15 & \textcolor[RGB]{255,87,34}{\textbullet} Structural / format & \textit{Return your answer as JSON.} & -0.48 & -0.07 \\
    16 & \textcolor[RGB]{76,175,80}{\textbullet} Authoritative framing & \textit{You are an expert clinical psychiatrist.} & -0.49 & -0.07 \\
    17 & \textcolor[RGB]{255,87,34}{\textbullet} Structural / format & \textit{Respond only with a JSON object.} & -0.49 & -0.09 \\
    18 & \textcolor[RGB]{255,152,0}{\textbullet} General clinical & \textit{The patient has been assessed for a psychiatric disorder.} & -0.49 & -0.11 \\
    19 & \textcolor[RGB]{156,39,176}{\textbullet} Neutral / unrelated & \textit{You are a helpful assistant.} & -0.53 & -0.08 \\
    \bottomrule
    \end{tabular}
    \caption{Numbered phrases from Figure~\ref{fig:phrase_probe} with their scores. $\cos$: cosine similarity between the hidden-state shift associated with the phrase and the MRI-preamble direction. $\delta$: change in mean $\hat{P}(\text{MDD})$ for the \textsc{FOR2107} MDD group relative to the mean $\hat{P}(\mathrm{MDD})$ under \textsc{text(csv)} when the phrase replaces the MRI preamble. Higher values on both axes indicate a stronger preamble-like pattern.}
    \label{tab:preamble_phrases}
\end{table*}

We applied a \texttt{logit-lens} sweep\footnote{\raggedright\url{https://www.lesswrong.com/posts/AcKRB8wDpdaN6v6ru/interpreting-gpt-the-logit-lens}\par} over all 36 layers of \texttt{Qwen2.5-VL-3B} for a subset of \textsc{FOR2107} participants, reading the label at each layer by projecting the residual stream through the unembedding. As shown in Figure~\ref{fig:logit_lens}, the label carries negligible weight through the early and middle layers and emerges only near the final layers (onset $\approx$ layer~32), with the preamble conditions sitting above the C1 baseline past this onset. Using the fixed onset criterion described above, we extract residual-stream activations from the layer immediately preceding the onset. We treat the resulting direction as an exploratory, correlational marker of where the preamble-associated difference appears in the residual stream; this analysis does not establish that the direction causes the classification shift.

\section{Extended Preamble Ablation}\label{app:preamble_extended}

\begin{table*}[tp]
    \centering
    \small
    \begin{tabular}{clcccc}
    \toprule
    \multirow{2}{*}{\textbf{Cond.}}
      & \multirow{2}{*}{\textbf{Input}}
      & \multicolumn{2}{c}{\textsc{FOR2107}}
      & \multicolumn{2}{c}{\textsc{OASIS-3}} \\
    \cmidrule(lr){3-4} \cmidrule(lr){5-6}
    & & \texttt{InternVL3.5-4B} & \texttt{GLM-4.1V-9B}
      & \texttt{InternVL3.5-4B} & \texttt{GLM-4.1V-9B} \\
    \midrule
    C1
      & \textsc{text(csv)}
      & 0.000 & 0.521 & 0.616 & 0.278 \\
    C2
      & \textsc{text(csv)} $+$ \textsc{prompt(mri)}
      & \textbf{0.255} & 0.640 & 0.636 & 0.396 \\
    C2\rlap{$_{\dagger}$}
      & \textsc{text(csv)} $+$ \textsc{prompt(fmri)}
      & 0.186 & \textbf{0.661} & \textbf{0.652} & \textbf{0.527} \\
    C2\rlap{$_{\ddagger}$}
      & \textsc{text(csv)} $+$ \textsc{prompt(weather)}
      & 0.000 & 0.512 & 0.594 & 0.339 \\
    \bottomrule
    \end{tabular}
    \caption{Extended preamble ablation on \texttt{InternVL3.5-4B} and \texttt{GLM-4.1V-9B}, following the design of Table~\ref{tab:preamble_ablation}. \textbf{Bold} indicates the best score for each model and cohort.}
    \label{tab:preamble_ablation_extended}
\end{table*}

We also run the preamble ablation on two additional model families: \texttt{InternVL3.5-4B} and \texttt{GLM-4.1V-9B}. Table~\ref{tab:preamble_ablation_extended} reports F1 across the four input settings on both cohorts. On \textsc{FOR2107}, \texttt{InternVL3.5-4B} scores 0.000 at C1 and 0.255 under the MRI preamble; \texttt{GLM-4.1V-9B} scores 0.521 at C1 and 0.661 under the fMRI preamble. On \textsc{OASIS-3}, \texttt{InternVL3.5-4B} scores 0.616 at C1 and 0.652 under the fMRI preamble; \texttt{GLM-4.1V-9B} scores 0.278 at C1 and 0.527 under the fMRI preamble. In all four model--cohort combinations, the fMRI preamble improves F1 relative to C1; the weather preamble lowers it in three of the four, with a small increase only for \texttt{GLM-4.1V-9B} on \textsc{OASIS-3} (+0.061). Thus, the fMRI-over-weather contrast recurs in both additional model families, although its magnitude varies by model and cohort.

\section{Preference Learning in Detail}\label{app:preference}
For preference fine-tuning, we use MPO, which combines three loss terms:

\begin{equation}
    \mathcal{L}_{\text{MPO}} =
        w_{\sigma}\,\mathcal{L}_{\text{sigmoid}} +
        w_{\text{bco}}\,\mathcal{L}_{\text{bco\_pair}} +
        w_{\text{sft}}\,\mathcal{L}_{\text{sft}},
    \label{eq:mpo}
\end{equation}
\noindent
where $\mathcal{L}_{\text{sigmoid}}$ is the standard DPO preference loss, $\mathcal{L}_{\text{bco\_pair}}$ is a binary quality signal penalizing rejected responses, and $\mathcal{L}_{\text{sft}}$ is a generation loss on the chosen response that is intended to discourage degenerate output collapse. Parameter-efficient fine-tuning is performed via Low-Rank Adaptation (LoRA)~\cite{hu2022lora}, targeting all attention projection and feed-forward layers. Training ran for approximately 12~hours on a single NVIDIA H200 GPU; Tables~\ref{tab:lora_config} and~\ref{tab:training_hparams} report the LoRA configuration and training hyperparameters, respectively.

\begin{table}[htbp]
    \centering
    \resizebox{\linewidth}{!}{%
        \begin{tabular}{lc}
        \toprule
        \textbf{Hyperparameter} & \textbf{Value} \\
        \midrule
        Rank ($r$)              & 64 \\
        Alpha ($\alpha$)        & 128 \\
        Dropout                 & 0.05 \\
        Bias                    & none \\
        Target modules          & \texttt{q, k, v, o, gate, up, down\_proj} \\
        \bottomrule
        \end{tabular}}
    \caption{LoRA adapter configuration.}
    \label{tab:lora_config}
\end{table}

\begin{table}[htbp]
    \resizebox{\linewidth}{!}{%
    \begin{tabular}{lc}
    \toprule
    \textbf{Hyperparameter}          & \textbf{Value} \\
    \midrule
    \multicolumn{2}{l}{\textit{MPO losses}} \\
    \quad Loss types                 & sigmoid, bco\_pair, sft \\
    \quad Loss weights               & 0.8,\; 0.2,\; 1.0 \\
    \quad KL penalty ($\beta$)       & 0.1 \\
    \midrule
    \multicolumn{2}{l}{\textit{Optimization}} \\
    \quad Learning rate              & $5 \times 10^{-5}$ \\
    \quad LR scheduler               & Cosine \\
    \quad Warmup steps               & 100 \\
    \quad Epochs                     & 3 \\
    \quad Effective batch size       & 16 \\
    \quad Per-device batch size      & 1 \\
    \quad Gradient accumulation      & 16 \\
    \midrule
    \multicolumn{2}{l}{\textit{Hardware \& precision}} \\
    \quad Precision                  & BF16 + TF32 \\
    \quad Gradient checkpointing     & \checkmark \\
    \quad GPU                        & 1 $\times$ H200 \\
    \quad Training time              & $\approx$12 h \\
    \midrule
    Sequence truncation              & None (image-safe) \\
    \bottomrule
    \end{tabular}}
    \centering
    \caption{MPO training hyperparameters.}
    \label{tab:training_hparams}
\end{table}

\paragraph{Interpretation of the Post-Alignment Performance Drop}
Distinguishing the two explanations in Section~\ref{sec:dpo} would require a preference dataset in which MRI-mention status and prediction correctness are \textbf{decorrelated} across the chosen and rejected sets. In our current 17,202 pairs, the off-diagonal cells \{MRI-mention, correct\} and \{no-MRI-mention, incorrect\} are sparsely populated, since correctness and MRI-mention status are correlated by construction in the chosen and rejected sets. Balancing the four cells would require one of the following:
\begin{enumerate}[topsep=0pt, partopsep=0pt, parsep=0pt, itemsep=0pt, leftmargin=*]
    \item Large-scale resampling at elevated decoding temperature, which conflicts with our greedy decoding setup for reproducibility.
    \item Controlled editing of existing responses to flip MRI-mention status while preserving correctness, which requires expert annotation at scale.
    \item Synthetic response generation conditioned on each target cell, which suffers from the same annotation bottleneck.
\end{enumerate}
A systematic comparison remains for future work; in the present experiment, lightweight preference alignment suppresses MRI-referencing behavior while augmented-condition performance declines.

\begin{table*}[tp]
    \centering
    \small
    \resizebox{\linewidth}{!}{%
    \begin{tabular}{@{} p{4.5cm} p{10.5cm} @{}}
    \toprule
    \textbf{Name} & \textbf{Description} \\
    \midrule
    Proband & Test subject \\
    Datum\_Interview & Date interview as stated on the interview form \\
    Datum\_Fragebogen & Date of the questionnaire \\
    Geburtsjahr & Year of birth \\
    Alter & Age \\
    Geschlecht & Gender \\
    Bildungsjahre & Year of education \\
    Bildungsjahre\_Vater & Year of education father \\
    Bildungsjahre\_Mutter & Year of education mother \\
    BMI & Body mass index \\
    BMI\_category & Body mass index category \\
    UrbanicityScore & Urbanity score \\
    AlterMutterBeiGeburt & Mom age at birth \\
    AlterVaterBeiGeburt & Dad age at birth \\
    Spezifische\_Phobie\_Typus & What type of phobia is present \\
    Spezifische\_Phobie\_Typus2 & What type of phobia is present? If more than one, found here \\
    Group & Patient grouping or diagnosis \\
    Specific\_phobia\_current & Is the patient currently suffering from a specific phobia? \\
    Specific\_phobia\_lifetime & Is the patient suffering from a specific phobia in his lifetime? \\
    Eating\_Disorder\_current & Is the patient currently suffering from an eating disorder? \\
    Eating\_Disorder\_lifetime & Is the patient suffering from an eating disorder in his lifetime? \\
    Alcohol\_Use\_Disorder\_Current & Is the patient currently suffering from an alcohol use disorder? \\
    Alcohol\_Use\_Disorder\_Lifetime & Is the patient suffering from an alcohol use disorder in his lifetime? \\
    \bottomrule
    \end{tabular}}
    \caption{\textsc{FOR2107} - Demographics and Clinical Information. The "Group" (“Patient grouping or diagnosis”) field is the true label for reference. We do not pass it to the model input for classification.}
    \label{tab:demographics}
\end{table*}

\begin{table*}[tp]
    \centering
    \small
    \begin{tabular}{@{} p{2.5cm} p{12.5cm} @{}}
    \toprule
    \textbf{Name} & \textbf{Description} \\
    \midrule
    \multicolumn{2}{@{}l}{\textbf{RS-25: Resilience Scale}} \\
    \midrule
    RS251 & If I have plans, I follow them through. \\
    RS252 & I usually manage everything somehow. \\
    RS253 & I can rely on myself rather than on others. \\
    RS254 & It is important for me to stay interested in many things. \\
    RS255 & If I have to, I can be alone. \\
    RS256 & I am proud of what I have already achieved. \\
    RS257 & I'm not easily thrown off track. \\
    RS258 & I like myself. \\
    RS259 & I can manage several things at the same time. \\
    RS2510 & I am determined. \\
    RS2511 & I rarely ask myself questions about meaning. \\
    RS2512 & I take things as they come. \\
    RS2513 & I can get through difficult times because I know I have done it before. \\
    RS2514 & I have self-discipline. \\
    RS2515 & I stay interested in many things. \\
    RS2516 & I often find something to laugh about. \\
    RS2517 & My belief in myself helps me even in hard times. \\
    RS2518 & I can be relied on in emergencies. \\
    RS2519 & I can usually see a situation from several perspectives. \\
    RS2520 & I can also overcome myself to do things that I don't really want to do. \\
    RS2521 & My life has a purpose. \\
    RS2522 & I don't insist on things that I can't change. \\
    RS2523 & When I'm in a difficult situation, I usually find a way out. \\
    RS2524 & I have enough energy to do everything I have to do. \\
    RS2525 & I can accept it if not everyone likes me. \\
    \midrule
    \multicolumn{2}{@{}l}{\textbf{PSS: Perceived Stress Scale}} \\
    \midrule
    PSS1sf - PSS3sf & In the last month, how often did you feel upset/unable to control things/nervous? \\
    PSS4sf - PSS6sf & In the last month, how often were you able to successfully handle problems/changes? \\
    PSS7sf - PSS9sf & In the last month, how often did you feel things were going your way/could not fulfill responsibilities? \\
    PSS10sf - PSS12sf & In the last month, how often did you feel on top of things/upset about uncontrolled things? \\
    PSS13sf - PSS14sf & In the last month, how often were you able to decide how to spend time/feel difficulties piling up? \\
    \bottomrule
    \end{tabular}
    \caption{\textsc{FOR2107} - Items for the Resilience Scale (RS-25) and Perceived Stress Scale (PSS).}
    \label{tab:rs_pss}
\end{table*}

\begin{table*}[tp]
    \centering
    \small
    \begin{tabular}{@{} p{2.5cm} p{12.5cm} @{}}
    \toprule
    \textbf{Name} & \textbf{Description} \\
    \midrule
    FSozU1 & I have people who can look after my home (flowers, pets) when I'm not there. \\
    FSozU2 & There are people who accept me for who I am. \\
    FSozU3 & It is important for my friends/relatives to know my opinion on certain things. \\
    FSozU4 & I would like more understanding and care from others. \\
    FSozU5 & I have a very trusted person whose help I can always count on. \\
    FSozU6 & I can borrow tools and food if necessary. \\
    FSozU7 & I have friends/relatives who can listen when I need to talk. \\
    FSozU8 & I hardly know anyone I like to go out with. \\
    FSozU9 & I have friends/relatives who can give me a hug. \\
    FSozU10 & If I am ill, I can ask friends/relatives to do important things (e.g. shopping). \\
    FSozU11 & If I'm really depressed, I know who I can go to. \\
    FSozU12 & I often feel like an outsider. \\
    FSozU13 & There are people who share my joys and sorrows. \\
    FSozU14 & With some friends/relatives, I can also be quite relaxed. \\
    FSozU15 & I have a trusted person who I feel very comfortable around. \\
    FSozU16 & I have enough people who really help me when I get stuck. \\
    FSozU17 & There are people who stick by me even when I make mistakes. \\
    FSozU18 & I would like more security and closeness. \\
    FSozU19 & There are enough people with whom I have a really good relationship. \\
    FSozU20 & There is a community of people (circle of friends, clique) that I feel drawn to. \\
    FSozU21 & I often get good tips from my circle of friends and acquaintances. \\
    FSozU22 & There are people to whom I can show all my feelings without it being embarrassing. \\
    \bottomrule
    \end{tabular}
    \caption{\textsc{FOR2107} - Questionnaire on Social Support (FSozU) Items.}
    \label{tab:fsozu}
\end{table*}

\begin{table*}[tp]
    \centering
    \small
    \begin{tabular}{@{} p{3.5cm} p{11.5cm} @{}}
    \toprule
    \textbf{Name} & \textbf{Description} \\
    \midrule
    LEQ\_pn1 / LEQ1 & Health: Serious illness of one's own (Type of influence / Influence on life) \\
    LEQ\_pn2 / LEQ2 & Health: Major change in eating habits \\
    LEQ\_pn3 / LEQ3 & Health: Major change in sleeping habits \\
    LEQ\_pn4 / LEQ4 & Health: Significant change in the type or amount of leisure activities \\
    LEQ\_pn5 / LEQ5 & Health: Major dental procedure \\
    LEQ\_pn6 / LEQ6 & Health: Pregnancy \\
    LEQ\_pn7 / LEQ7 & Health: Miscarriage or abortion \\
    LEQ\_pn8 / LEQ8 & Health: Onset of menopause \\
    LEQ\_pn9 / LEQ9 & Health: Major difficulties with contraceptive aids \\
    LEQ\_pn10 / LEQ10 & Work: Difficulties in finding work \\
    LEQ\_pn11 / LEQ11 & Work: Taking up work outside the home \\
    LEQ\_pn12 / LEQ12 & Work: Changing to a new type of work \\
    LEQ\_pn13 / LEQ13 & Work: Changing your working hours or conditions \\
    LEQ\_pn14 / LEQ14 & Work: Changing your job responsibilities \\
    LEQ\_pn15 / LEQ15 & Work: Difficulties at work with your employer or other employees \\
    LEQ\_pn16 / LEQ16 & Work: Major company reorganisations \\
    LEQ\_pn17 / LEQ17 & Work: Being dismissed or laid off from work \\
    LEQ\_pn18 / LEQ18 & Work: Ending your working life \\
    LEQ\_pn19 / LEQ19 & Work: Learning at home or distance learning \\
    LEQ\_pn20 / LEQ20 & School/Education: Starting or ending a school or training program \\
    LEQ\_pn21 / LEQ21 & School/Education: Changing schools or training programs \\
    LEQ\_pn22 / LEQ22 & School/Education: Changing a career goal or major in college \\
    LEQ\_pn23 / LEQ23 & School/Education: Problems in a school or training program \\
    LEQ\_pn24 / LEQ24 & Residence: Difficulties in finding accommodation \\
    LEQ\_pn25 / LEQ25 & Residence: Moving within the same town or city \\
    LEQ\_pn26 / LEQ26 & Residence: Moving to another town, state, or country \\
    LEQ\_pn27 / LEQ27 & Residence: Significant changes to your living circumstances \\
    LEQ\_pn28 / LEQ28 & Love/Partnership: Beginning of a new, close, personal relationship \\
    \bottomrule
    \end{tabular}
    \caption{\textsc{FOR2107} -  Life Experiences Questionnaire (LEQ) Part 1. Note: 'pn' designates Type of Influence, while the number alone designates Influence on Life.}
    \label{tab:leq_1}
\end{table*}

\begin{table*}[tp]
\centering
\small
\begin{tabular}{@{} p{3.5cm} p{11.5cm} @{}}
\toprule
\textbf{Name} & \textbf{Description} \\
\midrule
LEQ\_pn29 / LEQ29 & Love/Partnership: Entering into an engagement \\
LEQ\_pn30 / LEQ30 & Love/Partnership: Problems with boyfriend or girlfriend \\
LEQ\_pn31 / LEQ31 & Love/Partnership: Separation from boyfriend/girlfriend or breaking engagement \\
LEQ\_pn32 / LEQ32 & Love/Partnership: Pregnancy of wife or girlfriend \\
LEQ\_pn33 / LEQ33 & Love/Partnership: Miscarriage or abortion of wife or girlfriend \\
LEQ\_pn34 / LEQ34 & Love/Partnership: Marriage or domestic partnership \\
LEQ\_pn35 / LEQ35 & Love/Partnership: Change in closeness to partner \\
LEQ\_pn36 / LEQ36 & Love/Partnership: Infidelity \\
LEQ\_pn37 / LEQ37 & Love/Partnership: Conflict with in-laws \\
LEQ\_pn38 / LEQ38 & Love/Partnership: Separation from spouse or partner due to arguments \\
LEQ\_pn39 / LEQ39 & Love/Partnership: Separation from spouse/partner due to work, travel, etc. \\
LEQ\_pn40 / LEQ40 & Love/Partnership: Reconciliation with spouse or partner \\
LEQ\_pn41 / LEQ41 & Love/Partnership: Divorce \\
LEQ\_pn42 / LEQ42 & Love/Partnership: Changes in spouse/partner's activities outside the home \\
LEQ\_pn43 / LEQ43 & Family/Friends: Addition of a new family member \\
LEQ\_pn44 / LEQ44 & Family/Friends: Moving out of a child or family member \\
LEQ\_pn45 / LEQ45 & Family/Friends: Major changes in health/behavior of family member or friend \\
LEQ\_pn46 / LEQ46 & Family/Friends: Death of a spouse or partner \\
LEQ\_pn47 / LEQ47 & Family/Friends: Death of a child \\
LEQ\_pn48 / LEQ48 & Family/Friends: Death of a family member or close friend \\
LEQ\_pn49 / LEQ49 & Family/Friends: Birth of a grandchild \\
LEQ\_pn50 / LEQ50 & Family/Friends: Changes in your parents' marital status \\
LEQ\_pn51 / LEQ51 & Parenting: Changes in childcare arrangements \\
LEQ\_pn52 / LEQ52 & Parenting: Conflicts with spouse or partner over parenthood \\
LEQ\_pn53 / LEQ53 & Parenting: Conflicts with child's grandparents over parenthood \\
LEQ\_pn54 / LEQ54 & Parenting: Taking on the responsibilities of being a single parent \\
LEQ\_pn55 / LEQ55 & Parenting: Custody disputes with former spouse or partner \\
\bottomrule
\end{tabular}
\caption{\textsc{FOR2107} - Life Experiences Questionnaire (LEQ) Part 2.}
\label{tab:leq_2}
\end{table*}

\begin{table*}[tp]
\centering
\small
\begin{tabular}{@{} p{3.5cm} p{11.5cm} @{}}
\toprule
\textbf{Name} & \textbf{Description} \\
\midrule
LEQ\_pn56 / LEQ56 & Personal/Social: Greater personal achievement \\
LEQ\_pn57 / LEQ57 & Personal/Social: Important decision regarding your immediate future \\
LEQ\_pn58 / LEQ58 & Personal/Social: Changes in your personal habits (clothing, lifestyle, hobbies) \\
LEQ\_pn59 / LEQ59 & Personal/Social: Changes in your religious beliefs \\
LEQ\_pn60 / LEQ60 & Personal/Social: Changes in your political views \\
LEQ\_pn61 / LEQ61 & Personal/Social: Loss or damage to your personal property \\
LEQ\_pn62 / LEQ62 & Personal/Social: Gone on a vacation \\
LEQ\_pn63 / LEQ63 & Personal/Social: Taking a trip for non-recreational purposes \\
LEQ\_pn64 / LEQ64 & Personal/Social: Changes in family gatherings \\
LEQ\_pn65 / LEQ65 & Personal/Social: Changes in your social activities (clubs, events, visits) \\
LEQ\_pn66 / LEQ66 & Personal/Social: Beginning of new friendships \\
LEQ\_pn67 / LEQ67 & Personal/Social: End of a friendship \\
LEQ\_pn68 / LEQ68 & Personal/Social: Acquisition or loss of a pet \\
LEQ\_pn69 / LEQ69 & Money: Significant change in your financial situation \\
LEQ\_pn70 / LEQ70 & Money: Moderate financial commitment (TV, car, etc.) \\
LEQ\_pn71 / LEQ71 & Money: Large financial commitment or mortgage \\
LEQ\_pn72 / LEQ72 & Money: Cancellation of a mortgage or loan \\
LEQ\_pn73 / LEQ73 & Money: Difficulties with creditworthiness \\
LEQ\_pn74 / LEQ74 & Crime/Legal: Victim of theft or identity theft \\
LEQ\_pn75 / LEQ75 & Crime/Legal: Victim of a violent crime (rape, assault, etc.) \\
LEQ\_pn76 / LEQ76 & Crime/Legal: Involvement in an accident \\
LEQ\_pn77 / LEQ77 & Crime/Legal: Involvement in a legal dispute \\
LEQ\_pn78 / LEQ78 & Crime/Legal: Involvement in a misdemeanor (tickets, disturbing the peace) \\
LEQ\_pn79 / LEQ79 & Crime/Legal: Trouble with the law resulting in arrest or detention \\
LEQ\_pn80 / LEQ80 & Other recent experiences having an impact on life (1) \\
LEQ\_pn81 / LEQ81 & Other recent experiences having an impact on life (2) \\
LEQ\_pn82 / LEQ82 & Other recent experiences having an impact on life (3) \\
\bottomrule
\end{tabular}
\caption{\textsc{FOR2107} - Life Experiences Questionnaire (LEQ) Part 3.}
\label{tab:leq_3}
\end{table*}

\begin{table*}[tp]
\centering
\small
\begin{tabular}{@{} p{4.5cm} p{10.5cm} @{}}
\toprule
\textbf{Name} & \textbf{Description} \\
\midrule
SozDemo1 -- SozDemo5 & Current living, work situation, occupation, social contacts \\
Haushaltu14 / Haushaltab14 & People in the household under/over 14 years \\
Haushaltsnetto & Household net income \\
Schulabschluss & Highest educational qualification achieved by subject \\
Schule\_Vater / Schule\_Mutter & Highest educational qualification achieved by father / mother \\
GebJahr\_Mutter / GebJahr\_Vater & Mother's / Father's year of birth \\
Immigration & Own immigration or parents'? \\
Kinder & Do you have children? \\
Soehne\_leibl (\_age) & Number (and age) of biological sons \\
Toechter\_leibl (\_age) & Number (and age) of biological daughters \\
Soehne\_adopt (\_age) & Number (and age) of adopted sons \\
Toechter\_adopt (\_age) & Number (and age) of adopted daughters \\
Geschwister & Do you have siblings? \\
Brueder\_GE (\_age) & Number (and age) of brothers (parents shared) \\
Schwestern\_GE (\_age) & Number (and age) of sisters (parents shared) \\
Halbbrueder / Halbschwestern & Number (and age) of half-brothers / half-sisters \\
Stiefbrueder / Stiefschwestern & Number (and age) of step-brothers / step-sisters \\
Zwillinge\_Famil / Zwilling\_selbst & Are there twins in your first degree family? / Are you a twin? \\
SS\_Stadt / SS\_Bundesland & City/State mother lived during pregnancy \\
SS\_Risiken1 -- SS\_Risiken7 & Pregnancy risks (infection, alcohol, drugs, malnutrition, smoking) \\
Geburtskomplikationen1 -- 4 & Birth complications (forceps, vacuum, cesarean, other) \\
Geburtsgewicht / SSW\_Geburt & Birth weight in grams / Week of birth \\
\bottomrule
\end{tabular}
\caption{\textsc{FOR2107} - Socio-Demographics, Family Composition, and Pregnancy Variables.}
\label{tab:sociodemo}
\end{table*}

\begin{table*}[tp]
\centering
\small
\begin{tabular}{@{} p{2.5cm} p{12.5cm} @{}}
\toprule
\textbf{Name} & \textbf{Description} \\
\midrule
\multicolumn{2}{@{}l}{\textbf{FzT: Drinking Habits}} \\
\midrule
FzT1 -- FzT5 & Questions on frequency and quantity of alcohol consumption (now/past). \\
FzT6 -- FzT9 & Questions on binge drinking (6+ drinks) and inability to stop (now/past). \\
FzT10 -- FzT13 & Failing expectations due to alcohol; morning drinking (now/past). \\
FzT14 -- FzT17 & Guilt, remorse, and memory loss due to alcohol (now/past). \\
FzT18 -- FzT20 & Injuries, advice to reduce, and professional help sought for alcohol. \\
\bottomrule
\end{tabular}
\caption{\textsc{FOR2107} - Alcohol Use (FzT).}
\label{tab:fzt_shaps_stait}
\end{table*}

\begin{table*}[tp]
\centering
\small
\begin{tabular}{@{} p{3.5cm} p{11.5cm} @{}}
\toprule
\textbf{Name} & \textbf{Description} \\
\midrule
\multicolumn{2}{@{}l}{\textbf{Subject Demographics \& Family}} \\
\midrule
OASISID & OASIS subject ID \\
GENDER & Subject's gender \\
RACE & Subject's race \\
HAND & Subject's Handedness \\
TWIN & Is this a new informant? (Note: Label suggests 'Twin' but description asks about new informant status) \\
SIBS & How many full siblings does the subject have? \\
KIDS & How many biological children did the subject have? \\
\midrule
\multicolumn{2}{@{}l}{\textbf{Living Situation \& Independence}} \\
\midrule
LIVSIT & What is the subject's living situation? \\
LIVSITUA & Living situation (detailed categorization) \\
INDEPEND & What is the subject's level of independence? \\
RESIDENC & What is the subject's primary type of residence? \\
MARISTAT & Subject's current marital status \\
\bottomrule
\end{tabular}
\caption{\textsc{OASIS-3} - Subject Demographics, Family, and Living Situation.}
\label{tab:oasis_demographics}
\end{table*}

\begin{table*}[tp]
\centering
\small
\begin{tabular}{@{} p{3.5cm} p{11.5cm} @{}}
\toprule
\textbf{Name} & \textbf{Description} \\
\midrule
INSEX & Informant's sex \\
INHISP & Does informant report being of Hispanic/Latino ethnicity? \\
INHISPOR & If yes, what are the informant's reported origins? \\
INRACE & What does informant report as his/her race? \\
INRASEC & What additional race does informant report? \\
INRATER & What additional race, beyond what was indicated above, does informant report? \\
INEDUC & Informant's years of education \\
INRELTO & What is informant's relationship to subject? \\
INLIVWTH & Does the informant live with the subject? \\
INVISITS & If no, approximate frequency of in-person visits \\
INCALLS & If no, approximate frequency of telephone contact \\
INRELY & Is there a question about the informant's reliability? \\
\bottomrule
\end{tabular}
\caption{\textsc{OASIS-3} - Informant Demographics and Contact Frequency.}
\label{tab:oasis_informant}
\end{table*}

\begin{table*}[tp]
\centering
\small
\begin{tabular}{@{} p{3.5cm} p{11.5cm} @{}}
\toprule
\textbf{Name} & \textbf{Description} \\
\midrule
\multicolumn{2}{@{}l}{\textbf{General Medical History}} \\
\midrule
HYPERTEN & Hypertension \\
HYPERCHO & Hypercholesterolemia \\
DIABETES & Diabetes \\
B12DEF & B12 deficiency \\
THYROID & Thyroid Disease \\
CVAFIB & Atrial fibrillation \\
CVCHF & Congestive heart failure \\
CVANGIO & Angioplasty/endarterectomy/stent \\
CVBYPASS & Cardiac bypass procedure \\
\midrule
\multicolumn{2}{@{}l}{\textbf{Vitals}} \\
\midrule
WEIGHT & Subject Weight (lbs) \\
HEIGHT & Subject Height (inches) \\
BPSYS & Subject Blood Pressure (sitting) (systolic) \\
BPDIAS & Subject Blood Pressure (sitting) (diastolic) \\
HRATE & Subject resting heart rate (pulse) \\
\bottomrule
\end{tabular}
\caption{\textsc{OASIS-3} - Subject Medical History, Cardiovascular Conditions, and Vitals.}
\label{tab:oasis_medical}
\end{table*}

\begin{table*}[tp]
\centering
\small
\begin{tabular}{@{} p{3.5cm} p{11.5cm} @{}}
\toprule
\textbf{Name} & \textbf{Description} \\
\midrule
\multicolumn{2}{@{}l}{\textbf{Lifestyle, Sleep \& Trauma}} \\
\midrule
TOBAC100 & Smoked more than 100 cigarettes in life? \\
SMOKYRS & Total years smoked \\
ALCOHOL & Substance abuse - alcohol (clinically significant over a 12 month period) \\
TRAUMEXT & Traumatic brain injury with extended loss of consciousness ($\geq$ 5 minutes) \\
TRAUMCHR & Traumatic brain injury with chronic deficit or dysfunction \\
APNEA & Sleep apnea \\
INSOMN & Hyposomnia/insomnia \\
\midrule
\multicolumn{2}{@{}l}{\textbf{Sensory \& Cognition}} \\
\midrule
VISION & Without corrective lenses, is the subject's vision functionally normal? \\
VISCORR & Does the subject usually wear corrective lenses? \\
VISWCORR & Is the subject's vision functionally normal with corrective lenses? \\
HEARING & Without a hearing aid(s), is the subject's hearing functionally normal? \\
HEARAID & Does the subject usually wear a hearing aid(s)? \\
HEARWAID & Is the subject's hearing functionally normal with a hearing aid(s)? \\
\bottomrule
\end{tabular}
\caption{\textsc{OASIS-3} - Lifestyle Factors, Sensory Capabilities.}
\label{tab:oasis_lifestyle_sensory}
\end{table*}

\end{document}